\documentclass[11pt]{article}

\usepackage[margin=1in]{geometry}
\usepackage{times}
\usepackage[numbers]{natbib}
\usepackage{hyperref}
\usepackage{url}
\usepackage{amsmath, amssymb,amsfonts}
\usepackage{graphicx}
\usepackage{booktabs}

\usepackage[utf8]{inputenc} 
\usepackage[T1]{fontenc}    
\usepackage{hyperref}       
\usepackage{url}            
\usepackage{booktabs}       
\usepackage{amsfonts}       
\usepackage{amsmath}
\usepackage{amssymb}
\usepackage{nicefrac}       
\usepackage{microtype}      
\usepackage{xcolor}         
\usepackage{caption}
\usepackage{subcaption}
\usepackage{amsfonts}
\usepackage{url}
\usepackage{amsthm}
\usepackage{mathtools}
\usepackage{xcolor}
\usepackage{stmaryrd}
\usepackage{hyperref}
\usepackage{empheq}
\usepackage{wrapfig}
\usepackage{graphicx}

\newtheorem{remark}{Remark}

\newtheorem{prop}{Proposition}
\newtheorem{lem}{Lemma}
\newtheorem{cor}{Corollary}
\newtheorem{thm}{Theorem}
\newtheorem{ass}{Assumption}

\newcommand{\norm}[1]{\left\|{#1}\right\|}
\newcommand{\tr}[1]{\operatorname{Tr}\left({#1}\right)}
\newcommand{\eqdef}{\coloneqq}
\newcommand{\id}{\operatorname{Id}}

\newcommand{\dotp}[2]{\langle #1, #2 \rangle}

\newcommand{\RR}{\mathbb{R}}
\newcommand{\EE}{\mathbb{E}}
\newcommand{\cov}{\operatorname{Cov}}
\newcommand{\diag}[1]{\operatorname{Diag}\left({#1}\right)}

\definecolor{linkcolor}{RGB}{82, 82, 192}
\hypersetup{
    colorlinks=true,
    linkcolor=linkcolor,
    citecolor=linkcolor
}

\title{From Score Matching to Diffusion: \\ A Fine-Grained Error Analysis in the Gaussian Setting}

\author{
  Samuel Hurault \\
  ENS Paris, PSL, CNRS \\
   \texttt{samuel.hurault@ens.fr} \\
\and
  Matthieu Terris \\
  Univ. Paris-Saclay, Inria, CEA \\
  \texttt{matthieu.terris@inria.fr} \\
\and
  Thomas Moreau \\
  Univ. Paris-Saclay, Inria, CEA \\
  \texttt{thomas.moreau@inria.fr} \\
\and
  Gabriel Peyré \\
  ENS Paris, PSL, CNRS \\
   \texttt{gabriel.peyre@ens.fr} \\
}

\date{}

\begin{document}

\maketitle

\begin{abstract}
Sampling from an unknown distribution, accessible only through discrete samples, is a fundamental problem at the core of generative AI. The current state-of-the-art methods follow a two-step process: first, estimating the score function (the gradient of a smoothed log-distribution) and then applying a diffusion-based sampling algorithm---such as Langevin or diffusion models. The resulting distribution's correctness can be impacted by four major factors: the generalization and optimization errors in score matching, and the discretization and minimal noise amplitude in the diffusion. In this paper, we make the sampling error explicit when using a diffusion sampler in the Gaussian setting. We provide a sharp analysis of the Wasserstein sampling error that arises from these four error sources. This allows us to rigorously track how the anisotropy of the data distribution (encoded by its power spectrum) interacts with key parameters of the end-to-end sampling method, including the number of initial samples, the stepsizes in both score matching and diffusion, and the noise amplitude. Notably, we show that the Wasserstein sampling error can be expressed as a kernel-type norm of the data power spectrum, where the specific kernel depends on the method parameters. This result provides a foundation for further analysis of the tradeoffs involved in optimizing sampling accuracy.
\end{abstract}

\section{Introduction}

Sampling from an unknown distribution based only on a finite dataset is a core challenge in modern generative modeling. Score-based methods address this by estimating the gradient of a smoothed log-density -- known as the score function -- and then using it to guide sample generation via a stochastic differential equation (SDE). Because this score is not available in closed form, \cite{hyvarinen2005estimation} introduced score matching to estimate it directly from data, and later, \cite{vincent2011connection} showed that the optimal score estimator in the minimum mean squared error sense is a Gaussian noise denoising function. This connection laid the foundation for modern denoising score matching, where a neural network is trained to predict clean samples from noise-corrupted inputs using stochastic gradient descent (SGD).
Once trained, the learned score is used to guide sampling through a discretized diffusion process, such as Langevin dynamics or a discretized diffusion stochastic process \cite{song2019generative, song2020score}. This two-stage pipeline has achieved state-of-the-art results in generative tasks across images, audio, and videos \cite{ho2020denoising, ho2022video, dhariwal2021diffusion, karras2024analyzing}.

Despite their strong empirical performance, score-based diffusion models suffer from biases introduced at both stages of the pipeline: training yields an approximate score function, constrained by limited data and optimization errors, while sampling relies on discretized diffusion dynamics that introduce integration errors and noise scheduling truncation.
This paper investigates how these four sources of error interact within a simplified, analytically tractable setting.
Specifically, we focus on Gaussian data and linear scores, which enables exact explicit computation of how each error source affects the resulting sampled distribution.
In this simplified setting, computing the explicit errors already involves cumbersome calculations that uncover complex interactions between the four error sources, data geometry, and algorithmic parameters.
More specifically, our contributions are:

\noindent\textit{Error Analysis in SGD-based Denoising Score Matching:}  
   In Section~\ref{sec:DSM_error}, we conduct a detailed analysis of the errors introduced by stochastic gradient descent (SGD) when approximating the score function using denoising score matching, focusing on the Gaussian data setting. Theorem~\ref{thm:SGD_error} characterizes the covariance structure of the resulting error in the parameter space of a linear score model. Our analysis accounts for both the \emph{generalization error}, arising from the use of a finite dataset of size~$N$, and the \emph{optimization error}, stemming from a positive constant learning rate $\tau$. This explicit derivation of the stationary distribution of SGD for a denoising objective is novel and provides new insights into the magnitude and structure of these two sources of error.

\noindent\textit{Error Analysis in Diffusion-Based Sampling:}  
    In Sections~\ref{sec:langevin} and~\ref{sec:diffusion}, we present a parallel analysis of the sampling error introduced by Langevin and diffusion models. Theorems~\ref{thm:langevin_distance_general} and~\ref{thm:distance_general_diff} provide closed-form asymptotic expressions for the Wasserstein error resulting from (i) the use of an approximate score function (ii) the effects of time discretization with finite stepsize $\gamma$ and (iii) the presence of a nonzero final noise amplitude, modeled as a fixed noise level $\sigma$ for Langevin and a nonzero terminal diffusion time $T-t_K$ for diffusion models.
    
\noindent\textit{Unified Analysis of the Full Pipeline:}  
    Finally, in Corollaries~\ref{cor:langevin_final} and~\ref{cor:diffusion_final}, we combine these two error analyses, so as to analyze how the power spectrum of the initial data distribution influences the final sampling error. Notably, we demonstrate that, for both Langevin and diffusion sampling, the overall error can be expressed as a kernel norm on the power spectrum of the data. We explicitly characterize this kernel in terms of the parameters involved in the training and sampling algorithms.
    
All theoretical error predictions are carefully validated numerically in Figures~\ref{fig:SGD},~\ref{fig:langevin} and~\ref{fig:diffusion} through empirical evaluations across Gaussian data experiments.

\paragraph{Related Work}

\noindent\textit{Error Control in SGD Training.}
%
A comprehensive review of the asymptotic properties of SGD, including the nonconvergent case with a constant stepsize $\tau$, is given in \cite{bottou2018optimization}.
We build on the idea that SGD with constant stepsize $\tau$ converges to a stationary distribution, which was introduced by Mandt et al. \cite{mandt2016variational}.
A refined analysis of this phenomenon was later provided in \cite{dieuleveut2020bridging}.
In the Gaussian setting for score matching, explicit calculations can be carried out, and we leverage this approach in the first part of our contributions, which also integrates the generalization error due to the use of a finite number $N$ of data points.

\noindent\textit{Error Control in Langevin Dynamics.}
Langevin diffusion is a method for sampling from a distribution based on its score, with well-understood convergence properties. Its primary error sources stem from the choice of a positive stepsize $\gamma$ (discretization effects) and early stopping. The resulting biases were extensively studied in \cite{roberts1996exponential, wibisono2018sampling, durmus2019analysis, dalalyan2022bounding, li2021sqrt}.
While these analyses mostly focus on how these errors scale with dimension, we derive its explicit characterization based on the power spectrum of the data distribution.
%
%
Additionally, our analysis accounts for a potential inexact score. In this direction, several works \cite{dalalyan2019user, block2020generative, laumont2022bayesian} derived theoretical upper bounds on the Wasserstein sampling error due to inexact gradient. We adopt a similar perspective but instead of upper bounding the error, we restrict our analysis to linear scores so as to derive an exact expansion of the Wasserstein sampling error. 

\noindent\textit{Error Control in Diffusion Models.}
Recently, the convergence error induced by discretizing diffusion dynamics (with stepsize $\gamma$) and using a nonzero final time $T-t_K$ has been extensively studied. Prior work~\cite{benton2023nearly, chen2023improved, chen2022sampling, li2023towards} has established Wasserstein, KL or TV-convergence upper bounds assuming $L^2$ or $L^\infty$ accurate score estimates. These theoretical works cover both deterministic and stochastic schemes. 
%
In contrast to these works, which focus on bounding convergence, we provide explicit expansions of the error terms, which uncover additional interactions between algorithmic parameters and the underlying data geometry. 
%
Generalization error--stemming from training on a finite dataset of size $N$--has been studied in~\cite{zhu2023sample} for fully connected ReLU networks, with further refinements presented in~\cite{gupta2024improved}, and in~\cite{Biroli2024} in the context of mode collapse. \cite{kadkhodaie2023generalization} empirically verified that the variance of the trained score decreases as the amount of training data increases, which aligns with our theory. Moreover, the influence of the SGD learning rate on the sampling error has been investigated in~\cite{wu2025taking} for one-dimensional data and two-layer neural networks. 
Our work incorporates these error sources more tightly, showing their interplay with the data distribution spectrum in the Gaussian case.

An initial analysis of diffusion in the Gaussian setting was conducted in \cite{pierret2024diffusion}, deriving the explicit solution of the continuous diffusion SDE with the exact score function. We extend this by explicitly incorporating inexact score effects, diffusion discretization error being addressed in \cite{hurault2026geometry}.  Beyond its convenience for explicit calculations, recent empirical results from \cite{wang2024unreasonable, wang2023hidden} suggest that, in high-noise regimes, Gaussian models approximate neural network scores well. \cite{li2024understanding} additionally found that denoising models trained on large datasets tend to favor linearity.

\section{Background}
\label{sec:background}

\paragraph{Denoising Score Matching (DSM) and SGD}
Consider the problem of sampling from an unknown data distribution $p_\mathrm{data}$.
Recent approaches \cite{vincent2011connection, song2019generative, laumont2022bayesian} leverage an approximation of the score $\nabla \log p_\mathrm{data}$ with a denoising neural network \( D_{\sigma} : \mathbb{R}^d \times \RR^n \to \mathbb{R}^d \) with trainable parameters $\theta$.
The denoiser is trained by minimizing the $L^2$ denoising loss $\mathbb{E}_{x \sim p_\mathrm{data}, w \sim \mathcal{N}(0, I_d)} \left[ \frac12 \norm{ x - D_{ \sigma}(x + \sigma w, \theta)}^2 \right]$. According to Tweedie's formula \cite{efron2011tweedie}, the optimal estimator for this loss, called the \textit{MMSE denoiser}, satisfies $D_{\sigma}^\star(x) = x + \sigma^2 \nabla \log p_{\sigma}(x),$ with $p_{\sigma} =  p_\mathrm{data} * \mathcal{N}(0, \sigma^2 \id)$ the distribution of the data $p_\mathrm{data}$, smoothed by a Gaussian of standard deviation $\sigma$. Therefore, a denoiser trained with the $L^2$ loss provides an approximation of the score of the smoothed distribution $p_{\sigma}$. Denoting the approximated score $v_{\sigma}(x, \theta)$, the $L^2$ denoising loss can be rewritten as the denoising score matching loss 
\begin{equation}\label{eq:DSM_loss}
    F(\theta) = \mathbb{E}_{x \sim p, w \sim \mathcal{N}(0, I_d)} \left[ l(x, w, \theta) \right];\quad \text{ with }
    l(x, w, \theta) \eqdef \frac12 \norm{  v_{ \sigma}(x + \sigma w, \theta) + \frac{w}{\sigma}}^2
\end{equation}
To learn the score function, one needs to minimize \eqref{eq:DSM_loss} with respect to $\theta$. Due to its expectation structure, a convenient algorithm is the Stochastic Gradient Descent (SGD), which reads:
\begin{equation} \label{eq:basic_sgd}
\begin{split}
    \text{For $k \geq 0$}, \quad  x_k &\sim p_\mathrm{data}, \quad w_k \sim \mathcal{N}(0, \id), \quad \theta_{k+1} = \theta_k - \tau_k \nabla l(x_k, w_k, \theta_k). 
\end{split}
\end{equation}
When using a decaying stepsize \(\tau_k\) that satisfies \(\sum_k \tau_k = \infty\) and \(\sum_k \tau_k^2 < \infty\), the sequence defined by~\eqref{eq:basic_sgd} converges to the minimizer of $F$~\cite{mandt2016variational}. In contrast, when the learning rate \(\tau\) is constant---a common practice in training denoising neural networks---the sequence \(\theta_k\) typically does not converge but eventually oscillates around a stationary point, leading to a biased estimate that we refer to as the \emph{optimization error}. Additionally, since each SGD update samples \(x_k\) from only a finite dataset of size \(N\) drawn from \(p_{\mathrm{data}}\), this introduces a second source of bias, which we term the \emph{generalization error}. Following the interpretations in~\cite{bach2013non, dieuleveut2020bridging}, we view the SGD iterates as an ergodic Markov chain that converges to a unique stationary distribution dependent on both \(\tau\) and \(N\).

\looseness=-1
\paragraph{Langevin sampling}  Langevin dynamics~\cite{langevin1908theorie} is a stochastic method for sampling from a smooth distribution~\( p \). The continuous-time dynamics follow the stochastic differential equation (SDE):
\begin{equation*} 
    y_0 \in \mathbb{R}^d, \quad d y_t = -\nabla \log p(y_t)\, dt + \sqrt{2}\, d w_t,
\end{equation*}
where \( w_t \) is a standard Wiener process. The drift term guides samples toward high-density regions of~\( p \), while the diffusion term enables exploration. To simulate these dynamics, the Unadjusted Langevin Algorithm (ULA) applies an Euler–Maruyama discretization:
\begin{equation} \label{eq:ULA}
    y_0 \in \mathbb{R}^d, \quad y_{k+1} = y_k + \gamma \nabla \log p(y_k) + \sqrt{2 \gamma}\, w_k,
\end{equation}
where \( w_k \sim \mathcal{N}(0, \mathrm{Id}) \) and \( \gamma > 0 \) is the stepsize. Under assumptions such as strong convexity and Lipschitz continuity of \( -\log p \)~\cite{eberle2016reflection, durmus2017nonasymptotic, dalalyan2017theoretical}, ULA converges to a biased distribution due to discretization. \emph{Plug-and-play} Langevin methods~\cite{song2019generative, laumont2022bayesian} replace the unknown score \( \nabla \log p_{\mathrm{data}} \) with a learned approximation \( v_\sigma(\cdot, \theta) \), trained via denoising score matching~\eqref{eq:DSM_loss} at fixed noise level \( \sigma > 0 \), introducing further bias. The error of Langevin sampling therefore depends on both the discretization stepsize $\gamma$ and the considered noise level $\sigma$.

\paragraph{Diffusion model sampling}
Diffusion models generate samples from the data distribution by reversing a forward stochastic noising process. Several formulations have been proposed, including Variance Preserving (VP)~\cite{song2020score, ho2020denoising}, Variance Exploding (VE)~\cite{song2020score}, DDIM~\cite{song2020denoising}, and EDM~\cite{karras2022elucidating}.
We adopt a unified formulation of these variants, and describe the forward noising process by the SDE:
\begin{equation} \label{eq:forward_SDE}
    x_0 \sim p_{\mathrm{data}}, \quad dx_t = -\beta_t x_t\, dt + \sqrt{2 \xi_t}\, dw_t\enspace.
\end{equation}
The solution has a marginal distribution given by \( p_t = p_{\mathrm{data}}(s_t \cdot) * \mathcal{N}\left(0, (s_t \sigma_t)^2 \id\right) \), where \( B_t \coloneqq \int_0^t \beta_s \, ds \) and \( s_t \coloneqq e^{-B_t} \) is a time-dependent scaling factor. This corresponds to a rescaled version of the Gaussian-smoothed data distribution \( p_{\sigma_t} \), with noise standard deviation \( \sigma_t \coloneqq \left(2 \int_0^t \xi_s e^{2B_s} \, ds \right)^{1/2} \).
The VP diffusion, where \( \beta_t = \xi_t \), converges to \( \mathcal{N}(0, \mathrm{Id}) \), while the VE model sets \( \beta_t = 0 \), leading to an exploding variance as \( t \to \infty \). As discussed in~\cite{song2020score}, sampling from $p_{\mathrm{data}}$ can then be done by reversing this process, from time $T>0$ to $0$. As shown in Appendix~\ref{app:reverse}, this reverse-time process  \( q_t \eqdef p_{T-t} \), follows the SDE
\begin{equation} \label{eq:backward_SDE}
    y_0 \sim p_T, \quad dy_t = \beta_{T-t} y_t\, dt + (1 + \alpha) \xi_{T-t} \nabla \log p_{T-t}(y_t)\, dt + \sqrt{2 \alpha \xi_{T-t}}\, dw_t.
\end{equation}
where \( \alpha \geq 0 \) is a parameter controlling the diffusion coefficient. In this work, we consider the Euler–Maruyama discretized version of this process: for a stepsize $\gamma > 0$, and timesteps $t_k \eqdef k\gamma$, with $k \in \llbracket 1, K \rrbracket$ and $t_K \leq T$:
\begin{align} \label{eq:euler_disc}
        y_{k+1} &= (1 + \gamma \beta_{T-t_k}) y_k + \gamma (1 + \alpha) \xi_{T-t_k} \nabla \log p_{T-t_k}(y_k) + \sqrt{2 \alpha \gamma \xi_{T-t_k}} \cdot w_k
\end{align}
As in plug-and-play Langevin methods, the unknown score \( \nabla \log p_{T-t} \) is approximated using denoising score matching~\eqref{eq:DSM_loss}, but with a time-dependent noise level \( \sigma_t > 0 \). In practice, for optimal sampling performance, the diffusion is early-stopped at $t_K < T$. The bias of diffusion sampling thus depends on the stepsize $\gamma$ and the final time $t_K$.

In the remainder of this paper, we assume that the following assumption holds.
\begin{ass}\label{ass:gaussian} The data distribution is Gaussian, $p_{\mathrm{data}} \sim \mathcal{N}(\mu_\mathrm{data}, C_\mathrm{data})$ with $C_\mathrm{data} \succ  0$. 
\end{ass}
We denote $(\lambda_i)_{1 \leq i \leq d}$ the positive eigenvalues of $C_\mathrm{data}$ with associated eigenvectors $(u_i)_{1 \leq i \leq d}$.

\paragraph{Goal of the paper.} We analyze the sampling error of Langevin and diffusion models as a function of 
\((\tau, 1/N, \gamma)\), which are assumed to be small, as well as the noise parameters \(\sigma\) (for Langevin) and the stopping time \(t_K\) (for diffusion), with explicit constants depending only on the power spectrum~$(\lambda_i)$ of the covariance of the data.

\section{Modeling the Error in Denoising Score Matching} \label{sec:DSM_error}

In this section, we analyze the training of a linear parametric model $v_{\sigma}(x, \theta)$ to approximate the score of the smoothed distribution $\nabla \log p_\sigma$ by minimizing the denoising score matching loss~\eqref{eq:DSM_loss}. More precisely, we consider a score parametrized as the affine operator 
\begin{equation*}
v_\sigma(x, \theta) = - A x + b \quad\text{with parameters } \theta = (A, b) \in \RR^{d \times d}\times \RR^d
\end{equation*}
Under Assumption~\ref{ass:gaussian}, denoting $C_\sigma = C_\mathrm{data} + \sigma^2 \id$, the true score $\nabla \log p_\sigma$ is the affine operator $ v_{\sigma}^\star(x) = \nabla \log p_\sigma = - C_\sigma^{-1} x + C_\sigma^{-1} \mu_\mathrm{data}$. This implies that, under the Gaussian assumption, the chosen affine parameterization of the score function does not introduce any approximation error.
We consider training the parameter vector \( \theta = \operatorname{vec} \left[ -A^\top \; b^\top \right]^\top \in \mathbb{R}^{(d+1)d} \) using the SGD algorithm~\eqref{eq:basic_sgd} with a fixed learning rate \( \tau > 0 \), and a finite dataset ${X_N = \{x_i \sim p_{\mathrm{data}}\}_{0 \leq i < N}}$ of size \( N \):
\begin{equation} \label{eq:score_matching_SGD}
    \text{For $k \geq 0$}, \quad  x_k \in X_N, \quad w_k \sim \mathcal{N}(0, \id), \quad \theta_{k+1} = \theta_k - \tau \nabla l(x_k, w_k, \theta_k), 
\end{equation}
The objective of this section is to provide an explicit approximation of the stationary distribution sampled by this process, with respect to the choice of learning rate $\tau$ and the number of data samples~$N$. We thus consider two sources of error.

\begin{figure}[h]
\centering
 \includegraphics[width=0.4\linewidth]{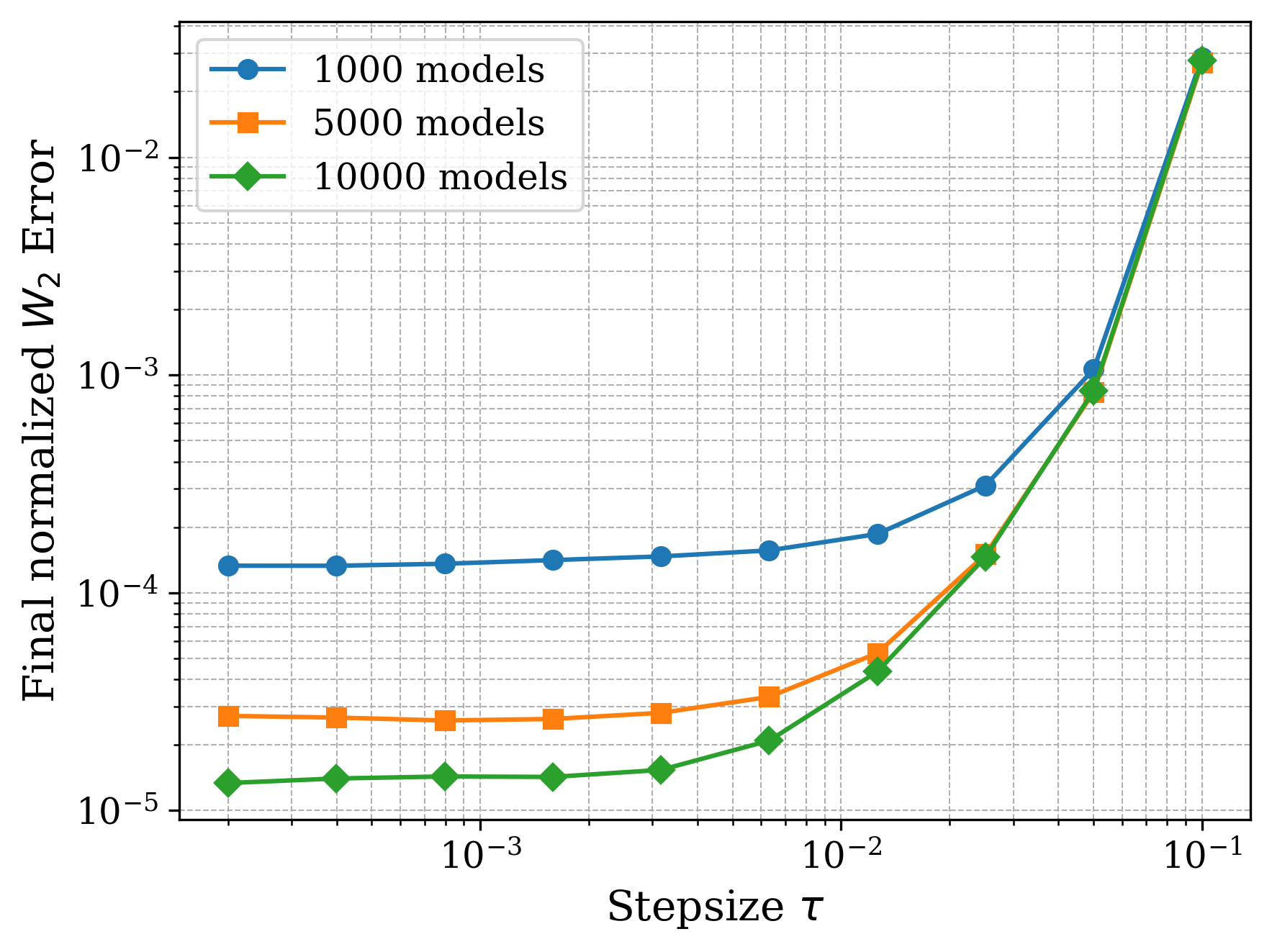}
\includegraphics[width=0.4\linewidth]{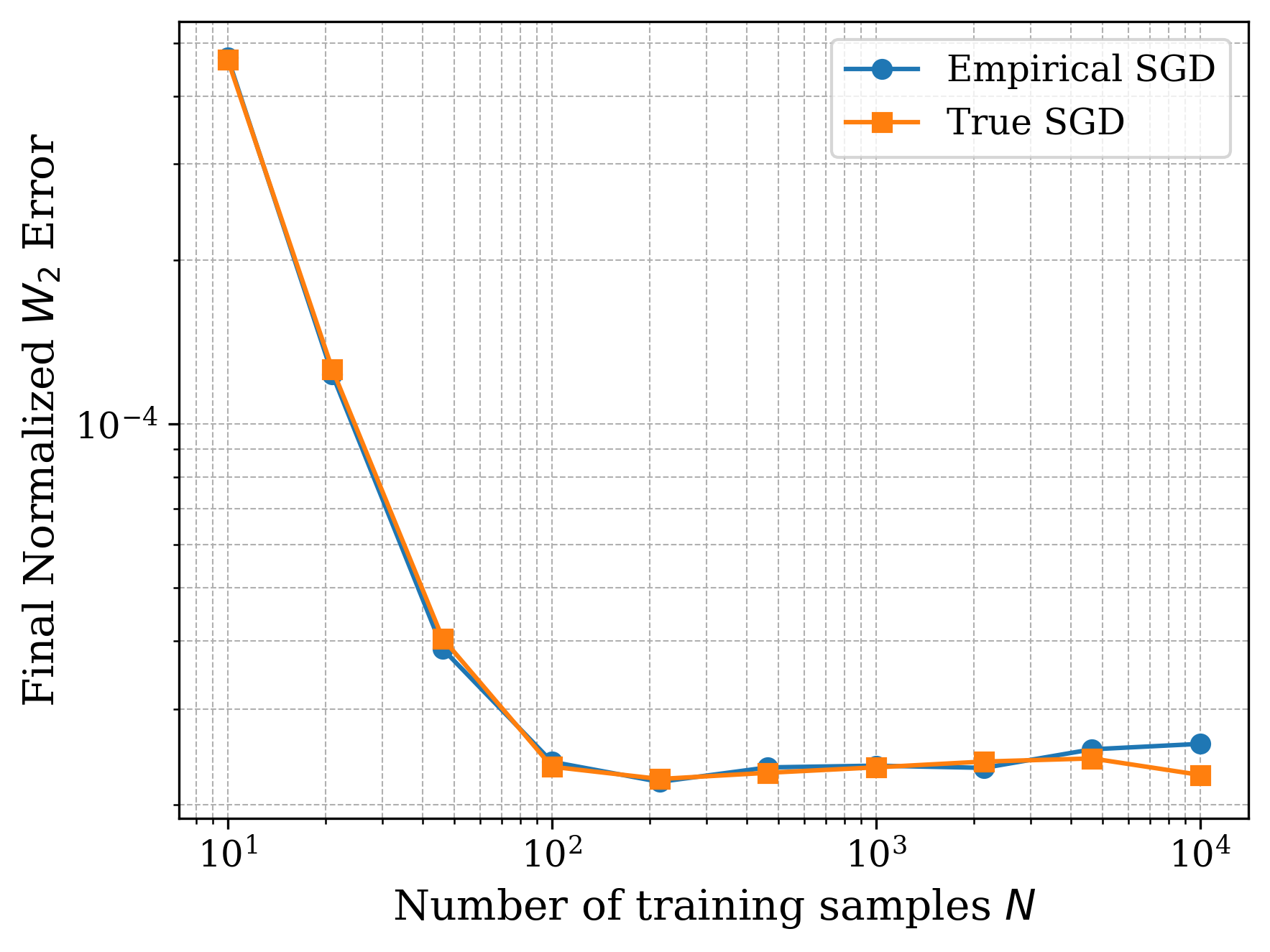}
  \caption{$W_2$ error, with respect to $\tau$ (left) and $N$ (right), between the theoretical (Theorem~\ref{thm:SGD_error}) and the empirical Gaussian approximations of the stationary distribution of the SGD algorithm~\eqref{eq:score_matching_SGD_emp}.}
  \label{fig:SGD} \vspace{-0.4cm}
\end{figure}

\paragraph{Generalization error}
It arises from the limited number $N$ of samples drawn from \( p_{\mathrm{data}} \) used for training the score model.
To account for this error source, we assume that, at each step of the SGD algorithm~\eqref{eq:score_matching_SGD}, the data sample is not chosen in $X_N$ but is drawn from the empirical Gaussian distribution $p_N \eqdef \mathcal{N}(\hat{\mu}_N, \hat{C}_N)$, where
\begin{equation} \label{eq:pN}
\hat{\mu}_N \eqdef \frac{1}{N}\sum_{x_i \in X_N} x_i \quad\text{and}\quad \hat{C}_N \eqdef \frac{1}{N}\sum_{x_i \in X_N} (x_i - \hat{\mu}_N)(x_i - \hat{\mu}_N)^\top.
\end{equation}
We thus analyze the following SGD iterations
\begin{equation} \label{eq:score_matching_SGD_emp}
    \text{For $k \geq 0$}, \quad  x_k \sim p_N, \quad w_k \sim \mathcal{N}(0, \id), \quad \theta_{k+1} = \theta_k - \tau \nabla l(x_k, w_k, \theta_k). 
\end{equation}
Note that this is an approximation of the actual algorithm~\eqref{eq:score_matching_SGD} used in practice, allowing for a simplified mathematical analysis. We empirically verify Figure~\ref{fig:SGD} the validity of this approximation.


\paragraph{Optimization error} The iterates \( (\theta_k) \) of~\eqref{eq:score_matching_SGD_emp} define a homogeneous Markov chain that, for small learning rate \( \tau \), converges to an invariant distribution \( \pi_{\tau, N}(\theta) \)~\cite{meyn2012markov, bach2013non}. As shown in~\cite{bach2013non}, for a quadratic loss, a constant learning rate  \( \tau > 0 \) introduces an error due to nonzero second and higher moments of \( \pi_{\tau, N} \). \cite[Proposition 3]{dieuleveut2020bridging} provides a closed-form second moment via the covariance of the per-step SGD error. However, this general result does not capture the specific structure of the denoising score matching loss~\eqref{eq:DSM_loss}. We refine this analysis by deriving a precise expression for the denoising case. The following theorem provides an explicit expansion, for small \( \tau \) and large \( N \), of the mean and covariance of the stationary distribution of the parameters \( A \) and \( b \), trained via~\eqref{eq:score_matching_SGD_emp}, in terms of the noise level \( \sigma \) and data geometry \( C_{\mathrm{data}} \). The proof is given in Appendix~\ref{app:SGD}.

\begin{thm}[Optimization and generalization errors in SGD for denoising score matching] \label{thm:SGD_error}
Under Assumption~\ref{ass:gaussian}, for fixed stepsize $\tau < \frac{2}{\max\left( \lambda_{max} + \sigma^2, 1 \right)},$ the SGD iterates~\eqref{eq:score_matching_SGD_emp} converge to a stationary distribution on the parameters $A$ and $b$. At stationarity, $b = \kappa + A \mu_\mathrm{data}$, with $\kappa$ and $A$ uncorrelated random variables with means
\begin{align*}
 \EE[\kappa] =  o\left( \frac{1}{N}\right)\quad\text{and}\quad \EE[A] = C_\sigma^{-1} + o\left( \frac{1}{N}\right)
\end{align*}
and covariances (denoting $P_\sigma \eqdef C_\sigma^{-1} C_\mathrm{data}$ and $Q_\sigma \eqdef C_\sigma^{-2}C_\mathrm{data}$)
\begin{equation*}
\begin{split}
\cov(\kappa) &= \frac{\tau}{2 \sigma^2} P_\sigma +  \frac{1}{N} Q_\sigma + \frac{\tau}{N \sigma^2} \left( \frac12 \langle P_\sigma, C_\mathrm{data} \rangle \id -  P_\sigma^2 \right) 
+ o \left(\tau, \frac{1}{N}\right) 
\end{split}
\end{equation*}
\begin{equation*}
\begin{split}
\cov(A)  = \frac{\tau}{2\sigma^2} P_\sigma \otimes \id + \frac{1}{N} \left( Q_\sigma \otimes Q_\sigma \right) (I_{d^2} + P_{trans})  -  \frac{\tau}{N \sigma^2} P_\sigma^2 \otimes \id  + o\left(\tau, \frac{1}{N}\right) .
\end{split}
\end{equation*}
$P_{trans}$ denotes the transpose in the vectorized space $\RR^{d^2}$  i.e. $P_{trans}[vec(X)] = vec(X^\top)$.
\end{thm}

The bias of SGD thus arises from the covariances of $A$ and $b$. We expressed these covariances as the sum of three terms, respectively, in $O(\tau)$, $O\left(\frac{1}{N}\right)$, and $O\left(\frac{\tau}{N}\right)$.
Note that the constants in each term depend only on the geometry of the data, characterized by \( C_{\mathrm{data}} \), and on the noise standard deviation~\( \sigma \). In particular, the covariances of \( \kappa \) and $A$ are diagonal, respectively in the eigenbasis of \( C_{\mathrm{data}} \) and in the eigenvector outer-product basis \( (u_i \otimes u_j^\top)_{ij} \) of \( \mathbb{R}^{d^2 \times d^2} \), with diagonal coefficients that depend (above $\tau$ and $N$), on the eigenvalues $\lambda_i$ of \( C_{\mathrm{data}} \) (the power spectrum) and on~\( \sigma \). 

\paragraph{Numerical validation}
\label{par:numerical_validation}
Figure~\ref{fig:SGD} confirms the results of Theorem~\ref{thm:SGD_error}. We consider Gaussian data in dimension \( d = 10 \), with eigenvalues \( \lambda_i \) decreasing in a quadratic power-law from \( 1 \) to \( 10^{-2} \). The SGD algorithm~\eqref{eq:score_matching_SGD_emp} is run on \( n_{\text{models}}  \) independent models until convergence, with varying learning rate \( \tau \) and sample size \( N \). For each configuration, we compute the empirical means and covariances of the estimators \( b \) and \( A \), and compare them to the theoretical predictions of Theorem~\ref{thm:SGD_error}, via the squared Wasserstein-2 ($W_2$) distance between the corresponding Gaussian distributions.  We report the sum of these $W_2$ distances for \( b \) and \( A \), after normalizing the covariances by their theoretical trace to ensure consistency across different \( N \) and \( \tau \). Experimental details are given Appendix~\ref{app:exp}.

Figure~\ref{fig:SGD} (left) shows the squared \( W_2 \) error as a function of the learning rate \( \tau \) (with fixed \( N = 10^3 \)), for various values of \( n_{\text{models}} \). For sufficiently small learning rates $\tau$, the theoretical distributions closely match the empirical ones, up to a residual error due to the finite number of \( n_{\text{models}} \). In the bottom plot, we show the error as a function of the number of training samples \( N \) (with fixed \( \tau = 10^{-3} \)). We compare the error between the theoretical distribution and the empirical results from two finite-sample SGD variants: \emph{Empirical SGD}~\eqref{eq:score_matching_SGD_emp}, where each iterate \( x_k \) is drawn from the empirical distribution \( p_N \) (as assumed in the theory), and \emph{True SGD}~\eqref{eq:score_matching_SGD}, where \( x_k \) is sampled from a fixed dataset \( X_N \). These results validate the theory for sufficiently large \( N \), and support the approximation of replacing the finite data distribution with its empirical Gaussian estimate in~\eqref{eq:score_matching_SGD_emp}.

\section{Langevin and diffusion model sampling errors}

\looseness=-1
In this section, we present a parallel analysis of the sampling errors introduced by Langevin dynamics (Section~\ref{sec:langevin}) and diffusion models (Section~\ref{sec:diffusion}) with linear scores. For each algorithm, Theorems~\ref{thm:langevin_distance_general} and~\ref{thm:distance_general_diff} provide similar closed-form asymptotic expressions for the resulting Wasserstein error, capturing the effects of: (i) discretization due to finite stepsize, (ii) nonzero noise level or finite diffusion time, and (iii) inexact linear score functions. These results are then specialized, in Corollaries~\ref{cor:langevin_final} and~\ref{cor:diffusion_final}, to a linear score trained by denoising score matching, based on the analysis from the previous section.

\subsection{Langevin sampling error}
\label{sec:langevin}

We first consider the Unadjusted Langevin (ULA) sampling algorithm~\eqref{eq:ULA} with constant stepsize $\gamma >0$ and with the unknown true score $\nabla \log p_\mathrm{data}$ replaced by the linear score $v(x, \theta) = -A x + b$.
When $A$ is symmetric, this is equivalent to assuming Gaussian data with mean $\mu = A^{-1}b$ and covariance~$A^{-1}$. 
We first analyze the sampling error of this algorithm arising from both the discretization of the Langevin dynamics and the inexact approximation of the score function. 

\paragraph{Discretization error} The ULA iterates~\eqref{eq:ULA} then sample a Gaussian distribution with explicit mean and variance given in the following Lemma. The discretization error
appears in the covariance of the sampled distribution, which differs from $A^{-1}$ due to the positive stepsize $\gamma>0$. This is not a new result, but an important result for the rest of our analysis. The covariance is written using the inverse of the Lyapunov operator $L^\gamma_C$, defined, for $C \in \RR^{p\times p}$ and $\gamma >0$ by $L^\gamma_C[X] = CX + XC^\top - \gamma C X C^\top$.

\begin{lem}[ULA solution with linear score, proof in Appendix~\ref{app:lemma_langevin_1})] \label{lem:lemma_langevin_1}
    Assuming $\operatorname{Re}(\operatorname{eig}(A)) > 0$, and ${\gamma < \frac{2}{\lambda_{max}(A)}}$, the ULA algorithm~\eqref{eq:ULA} with linear score $v(x, \theta) = -A x + b$ converges to 
     \begin{equation*}
    q = \mathcal{N}(A^{-1}b, \Sigma^\gamma_A) \quad \text{with} \quad  \Sigma^\gamma_A \eqdef (L^\gamma_A)^{-1}[2\id].
    \end{equation*}
\end{lem}

\paragraph{Score estimation error} We now analyze the impact of a small perturbation of the score on the above asymptotic distribution. We saw in Section~\ref{sec:DSM_error}, that under the Gaussian assumption~\ref{ass:gaussian}, the optimal score for the denoising score matching loss corresponds to $A^* = C_\sigma^{-1}$ and $b^* = C_\sigma^{-1} \mu_\mathrm{data}$. Instead of using this true score, and following the form of the error on $b$ identified in Theorem~\ref{thm:SGD_error}, we suppose that we have its following random perturbed version 
\begin{equation} \label{eq:gen_inexact_score}
    v_\sigma(x, \theta) = - A_\varepsilon x + b_\varepsilon, \quad \text{with}\quad 
    A_\varepsilon = A^* + \varepsilon \Delta, \quad b_\varepsilon = b^* + \varepsilon (\delta + \Delta \mu_\mathrm{data}),
\end{equation}
where \(\Delta\) and \(\delta\) are uncorrelated and zero-mean random variables taking values in $\mathbb{R}^{d \times d}$ and $\mathbb{R}^{d}$.
Using the score~\eqref{eq:gen_inexact_score}, applying Lemma~\ref{lem:lemma_langevin_1}, we get that, for small enough $\gamma$, the ULA iterates sample the invariant Gaussian distribution $q_\varepsilon \eqdef \mathcal{N}(\mu_\varepsilon, \Sigma_\varepsilon)$ with $\mu_\varepsilon \eqdef A_\varepsilon^{-1} b_\varepsilon$ and $\Sigma^\gamma_\varepsilon = (L^\gamma_{A_\varepsilon})^{-1}[2\id]$.

We derive in Theorem~\ref{thm:langevin_distance_general} the expansion of the averaged Wasserstein distance $\EE_{\delta, \Delta}  \left[ W_2^2(p_\mathrm{data},q_\varepsilon) \right]$
between this limiting distribution and the data distribution. The Wasserstein distance writes, for Gaussian distributions, as the sum of the $L^2$ norm between means $\norm{\mu_\mathrm{data} - \mu_\varepsilon}^2$ and the Bures squared distance between covariances. 
The theorem expresses the overall error as the sum of three distinct terms. The first term $\mathcal{E}^{(0)}_{\sigma, \gamma}$ corresponds to the zeroth-order error, which arises from the discrepancy between \(C_\sigma\) and \(C_{\mathrm{data}}\) (due to $\sigma>0$) and from the bias introduced by discretization when \(\gamma > 0\).  The error terms $\mathcal{E}_{\sigma, \gamma}^{\text{mean}}(\delta)$ and $\mathcal{E}_{\sigma, \gamma}^{\text{cov}}(\Delta)$ stem from the perturbation in $\varepsilon$ of the parameters $b$ and $A$ in the linear score~\eqref{eq:gen_inexact_score}, which respectively induce an error on the mean and on the covariance of the invariant distribution. These terms depend on both the spectral decomposition $(\lambda_i)$ of \(C_{\mathrm{data}}\) and the projection of the perturbations $\Delta$ and $\delta$ onto the eigenspace of \(C_{\mathrm{data}}\), thereby reflecting how these perturbations interact with the data distribution.
\begin{thm}[ULA sampling error with inexact linear score, proof in Appendix~\ref{app:langevin_distance_general}] \label{thm:langevin_distance_general}
For a small enough stepsize $\gamma$, the ULA algorithm~\eqref{eq:ULA} with inexact linear score~\eqref{eq:gen_inexact_score} converges to an invariant Gaussian distribution $q^\varepsilon$ that satisfies 
\begin{equation*} 
\EE_{\delta, \Delta}  \left[ W_2^2(p_\mathrm{data},q_\varepsilon) \right] =  \mathcal{E}^{(0)}_{\sigma, \gamma} + \varepsilon^2 \left( \mathcal{E}_{\sigma, \gamma}^{\text{mean}}(\delta) + \mathcal{E}_{\sigma, \gamma}^{\text{cov}}(\Delta) \right) + o(\varepsilon^2)
\end{equation*}
The explicit expressions of each term are provided in Appendix~\ref{app:langevin_distance_general}. They involve the parameters~\( \gamma \),~\( \sigma \), the eigenvalues \( (\lambda_i) \) of \( C_{\mathrm{data}} \), as well as, for $\mathcal{E}_{\sigma, \gamma}^{\text{mean}}(\delta)$, and $\mathcal{E}_{\sigma, \gamma}^{\text{cov}}(\Delta)$  the averaged projections of the perturbations $\delta$ and $\Delta$ onto the eigenvector outer-product basis \( (u_i u_j^\top)_{ij} \) of \(C_{\mathrm{data}}\):
\[
\mathbb{E}_{\delta} \left[ \langle \operatorname{Cov}(\delta), u_i u_i^\top \rangle \right], \quad 
\mathbb{E}_{\Delta} \left[ \langle \Delta, u_i u_j^\top \rangle^2 \right], \quad \text{and} \quad
\mathbb{E}_{\Delta} \left[ \langle \Delta, u_i u_j^\top \rangle \langle \Delta, u_j u_i^\top \rangle \right].
\]
\end{thm}
The proof combines the second-order expansion in~$\varepsilon$ of the mean and covariance of $q_\varepsilon$ (Lemma~\ref{lem:lemma_langevin_1}) with the second-order expansion of the $W_2$ distance between Gaussians (Proposition~\ref{prop:Langevin_2}). Although written for $A^* = C_\sigma^{-1}$, this theorem generalizes to any $A^*$ that co-diagonalizes with $C_\mathrm{data}$. 

\paragraph{Langevin sampling with a linear score trained by SGD} We now calculate the global error of Langevin sampling, combining the Langevin sampling error (Theorem~\ref{thm:langevin_distance_general}) and the SGD training error  (Theorem~\ref{thm:SGD_error}). We thus use as linear score $v_\sigma(x, \theta) = -A x + b$
with parameters \( A \) and \( b \) trained by minimizing the DSM loss~\eqref{eq:DSM_loss} at noise level \( \sigma \) with the SGD algorithm~\eqref{eq:score_matching_SGD_emp}. From Theorem~\ref{thm:SGD_error}, 
\begin{equation*}
\begin{split}
    b = C_\sigma^{-1}\mu_\mathrm{data} + \delta + \Delta \mu_\mathrm{data} \quad &\text{and} \quad
    A = C_\sigma^{-1} + \Delta, \\ 
    \quad\text{where}\quad \delta \sim \mathcal{N}(0, \cov(\kappa)) \quad &\text{and}\quad \Delta \sim \mathcal{N}(0, \cov(A)),
\end{split}
\end{equation*}
Using this score, we apply Theorem~\ref{thm:langevin_distance_general} and obtain an averaged error that depends on the values of $\langle \cov(\delta), u_i u_i^\top \rangle$, $\mathbb{E}[\langle \Delta, u_i u_j^\top \rangle^2]$, \text{and} $\mathbb{E}[\langle \Delta, u_i u_j^\top \rangle \langle \Delta, u_j u_i^\top \rangle]$. Using Theorem~\ref{thm:SGD_error}, the expansion of these quantities for small \( \tau \) (SGD learning rate) and large $N$ (number of training samples) can be computed with respect to $\sigma$ and the eigenvalues $\lambda_i$ of the $C_\mathrm{data}$.  This leads to the following corollary. 
\begin{cor}[ULA sampling error with linear score trained by SGD, proof in Appendix~\ref{app:langevin_final}]
\label{cor:langevin_final}
Under assumption~\ref{ass:gaussian}, consider a linear score $v_\sigma(x, \theta) = - A x + b$ with parameters \( \theta=(A,b) \) trained via the SGD algorithm~\eqref{eq:score_matching_SGD_emp}, with \( N \) data samples and constant learning rate \( \tau > 0 \). Then, for small enough stepsize $\gamma$, the ULA algorithm~\eqref{eq:ULA} samples an invariant Gaussian distribution $q$ that satisfies \vspace{-0.1cm}
\begin{equation*} \label{eq:langevin_W2}
 \! \! \mathbb{E}_{\theta}\left[  W_2\left(p, q\right)^2 \right] \! =  
 \mathcal{E}^{(0)}_{\sigma, \gamma} \! +  \! \!  \sum_{i,j = 1}^d  \! \left( \! \frac{\tau}{\sigma^2} k_{\sigma, \gamma}^\tau (\lambda_i, \lambda_j)  \! +  \!\frac{\tau}{N \sigma^2} k_{\sigma, \gamma}^{\tau, N}(\lambda_i, \lambda_j)  \!+  \! \frac{1}{N} k_{\sigma, \gamma}^N(\lambda_i, \lambda_j) \! \right)  \! + \! o\left(\! \tau, \frac{1}{N} \! \right).
 \end{equation*} 
$k_{\sigma, \gamma}^\tau, k_{\sigma, \gamma}^N, k_{\sigma, \gamma}^{\tau, N} : \RR^d \times \RR^d \to \RR$ are kernel functions capturing interactions between the eigenvalues of the $C_\mathrm{data}$. Their explicit expressions are given in Appendix~\ref{app:langevin_final}. 
\end{cor}

This result characterizes the relationship between the data spectral properties and the algorithmic parameters involved in score training (SGD learning rate $\tau$, number of data samples $N$, noise standard deviation $\sigma$) and in sampling (stepsize $\gamma$). The overall error takes the form of a kernel norm over the power spectrum of the data.  This result highlights a fundamental trade-off in the selection of the parameter \(\sigma\). Specifically, the first term $\mathcal{E}^{(0)}$ grows as \(O(\sigma^4)\) for large \(\sigma\), due to the difference between $p_\mathrm{data}$ and $p_\sigma$, while the second and third terms diverge as \(O(\frac{1}{\sigma^2})\) for small \(\sigma\).

\begin{figure}[h]
  \centering \vspace{-0.6cm}
   \includegraphics[width=0.4\linewidth]{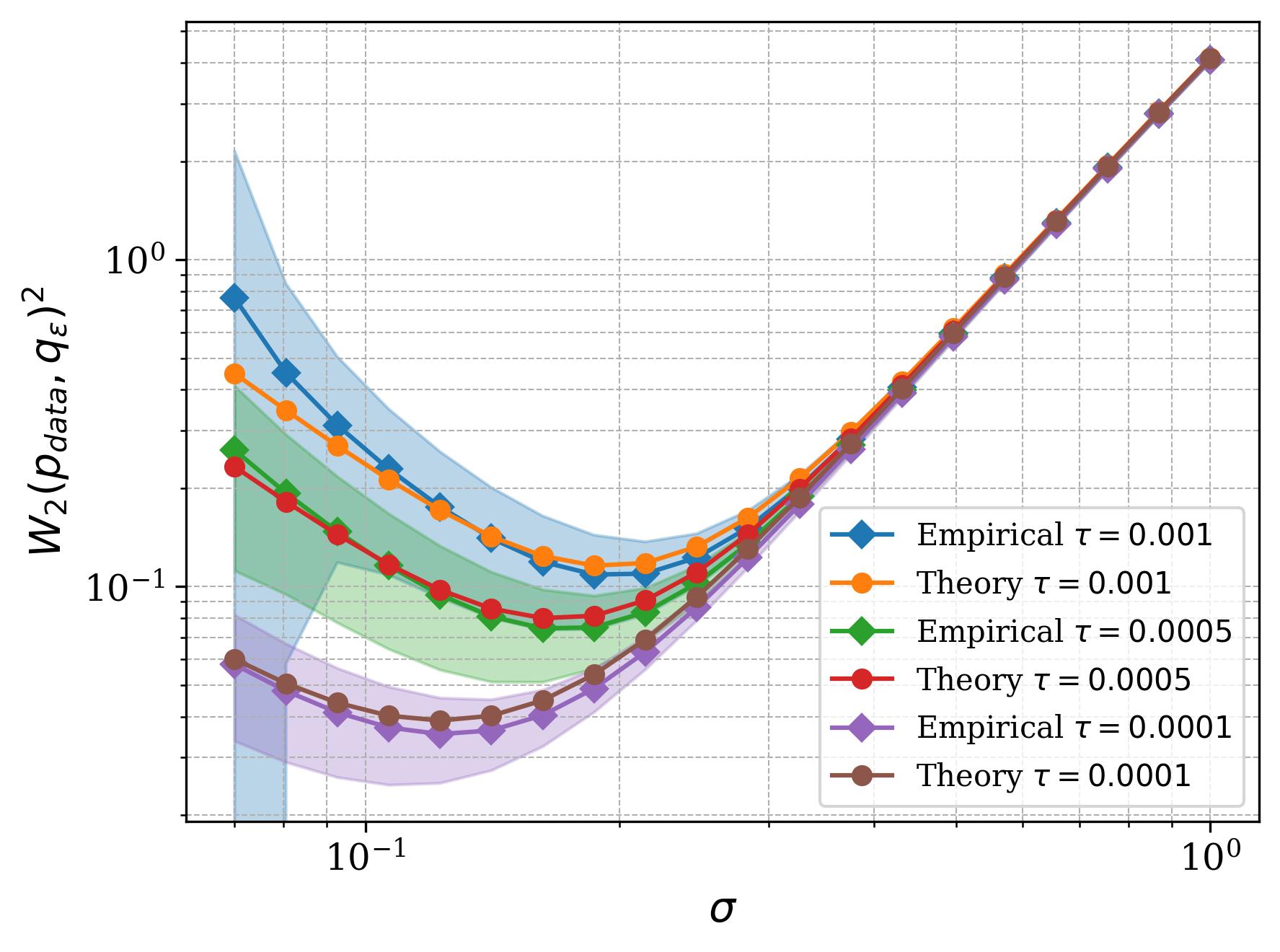} 
   \includegraphics[width=0.4\linewidth]{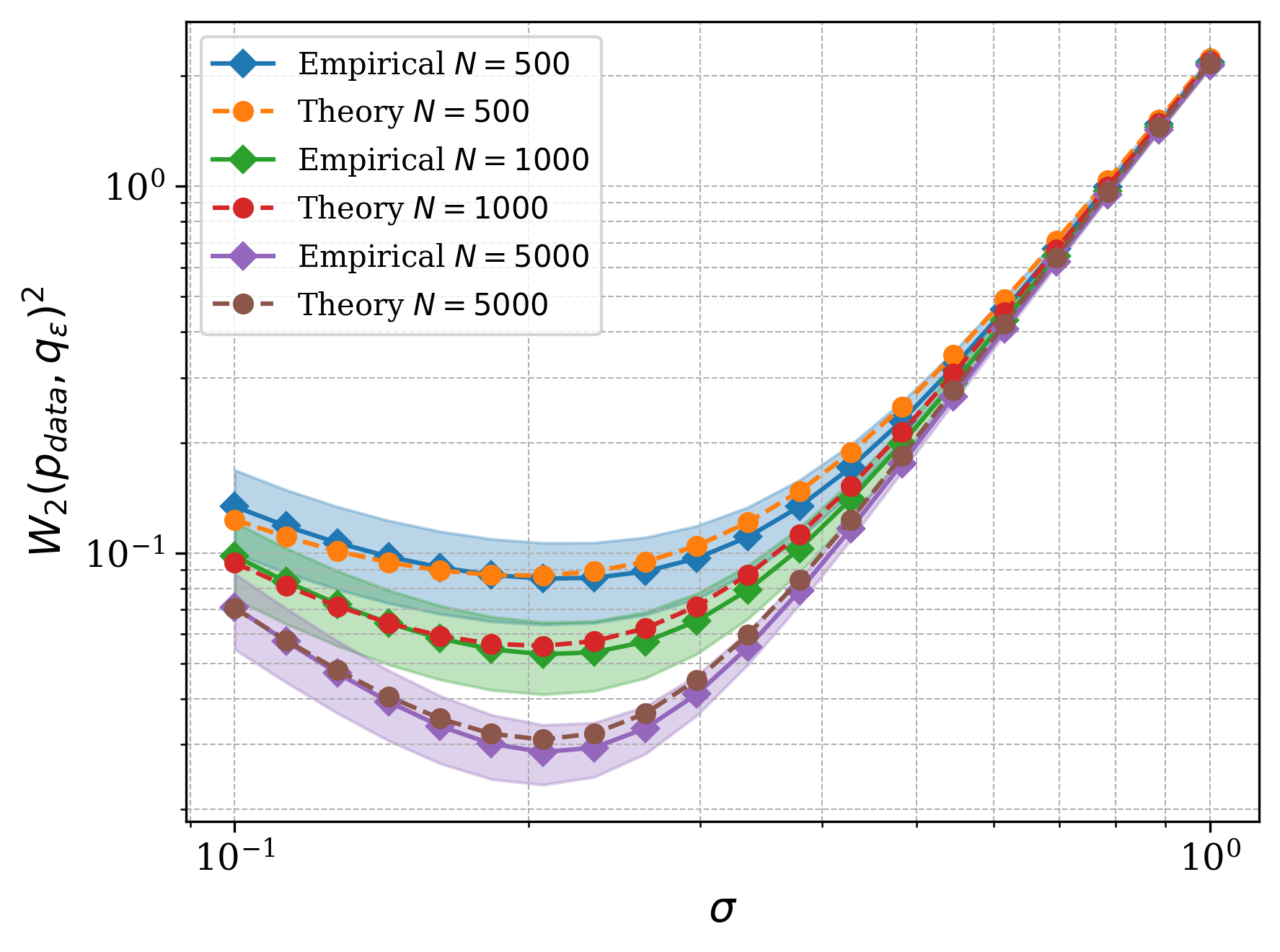}
  \caption{Theoretical vs Empirical Langevin sampling $W_2$ error as a function of $\sigma$, for various score training parameters $\tau$ (left) and $N$ (right).}
  \label{fig:langevin} 
\end{figure}

\paragraph{Numerical validation} We verify this result numerically in Figure~\ref{fig:langevin}. In the same setting (dimension $10$) as Figure~\ref{fig:SGD} (detailed at the end of Section~\ref{sec:DSM_error}), we take $n_{models} = 10^2$ linear score models $v_\sigma(x) = -A x +b$ with $A$ and $b$ sampled from the theoretical Gaussians from Theorem~\ref{thm:SGD_error} with noise level $\sigma$, learning rate $\tau$ and $N$ data samples. For each model, we sample $10^6$ particles with the Langevin iterations~\eqref{eq:ULA} with stepsize $\gamma$, compute the empirical mean and covariance, and estimate the empirical (squared) $W_2$ distance with the data distribution. We plot in Figure~\ref{fig:langevin} the comparison, for various $\tau$ (left) and $N$ (right), between the average (and standard deviation) empirical $W_2$ distance and the theoretical one from Corollary~\ref{cor:langevin_final}. 

The trade-off in the choice of the noise parameter $\sigma$ is illustrated in Figure~\ref{fig:langevin}. For a given data power spectrum and fixed parameters \(\tau\) and \(N\), there exists an optimal \(\sigma^*\) that minimizes the sampling error. We further  investigate in Appendix~\ref{app:analysis_Langevin} the behavior of \(\sigma^*\) with respect to the data power spectrum, and the parameters $\tau$, $N$, and $\gamma$.

\subsection{Diffusion model sampling error}
\label{sec:diffusion}

We now conduct a similar error analysis on the discretized diffusion sampling algorithm~\eqref{eq:euler_disc} where, at each timestep, the score $\nabla \log p_{t}$ is replaced by a linear time-dependent model ${v_{t}(x, \theta) = - A_{t}x + b_{t}}$. In particular, we analyze in detail the error arising from inexact approximation of the score function.

Using the notations introduced in Section~\ref{sec:background}, under Assumption~\ref{ass:gaussian}, the marginal~$p_t$ of the forward noising process~\eqref{eq:forward_SDE} is Gaussian with mean $\mu_t = s_t \mu_{\mathrm{data}}$ and covariance $\Sigma_t = s_t^2(C_{\mathrm{data}} + \sigma_t^2\id)$.  Gaussian diffusion with true linear score $\nabla \log p_t$ corresponds to $A_t = \Sigma_t^{-1}$ and $b_t = \Sigma_t^{-1} \mu_t$. In this case, the continuous backward SDE~\eqref{eq:backward_SDE} samples the Gaussian $q_t = \mathcal{N}(\mu_{T-t}, \Sigma_{T-t})$, which approaches $p_\mathrm{data}$ as $T - t \to 0$.

\paragraph{Discretization error} 
The discretized version~\eqref{eq:euler_disc} samples, at each timestep $t_k$, a Gaussian distribution, different from $p_{T-t_k}$ due to a first discretization bias. 
In this work, we do not explicitly study the discretization error of diffusion models. However, \cite{talay1990expansion} studied the first-order expansion, for small stepsize~$\gamma$, of the weak error due to Euler-Maruyama discretization of general SDEs. Denoting $(y_t)_{t\leq T}$ the continuous solution~\eqref{eq:backward_SDE} and $(y_k)_{k \leq K}$ the discrete solution of~\eqref{eq:euler_disc}, under broad assumptions on the drift $v_t$ and the diffusion-term $a_t$, they show that for a $\mathcal{C}^\infty$ test function $f : \RR^d \to \RR$:
\begin{align*}
\EE[f(y_k)] - \EE[f(y_{t_k})] = \gamma \int_0^{t_k} \psi_s(y_{s})ds + o(\gamma)
\end{align*}
for some time-dependent function $\psi_s : \RR^d \to \RR$. The specialization of this first-order error for Gaussian data is detailed in \cite{hurault2026geometry}. With this result, we get that the discretization error on the mean and covariance of the sampled distribution is of order $\gamma$. Given that the mean and covariance evolution along the discretized process decouples in the eigenbasis of the data, we get:
\begin{lem}[Discretized diffusion with linear score, proof in Appendix~\ref{app:diffusion_discretization_1}] \label{prop:diffusion_discretization_1}
Under Assumption~\ref{ass:gaussian}, the discretized diffusion process~\eqref{eq:euler_disc} samples, at step $k \in \llbracket 0, K \rrbracket$, a Gaussian distribution with 
\begin{align*} 
\EE[y_{k}] &= \mu_{T-t_k} + \gamma \, d_{T-t_k} \cdot \mu_{\mathrm{data}} + o(\gamma)  \\
\operatorname{Cov}(y_k) &=  \Sigma_{T-t_k} +  \gamma \,  D_{T-t_k} + o(\gamma) 
\end{align*}
for some time-dependent matrices $d_{T-t}, D_{T-t} \in \RR^{d \times d}$ co-diagonalizing with the data covariance $\Sigma_{\mathrm{data}}$. 
\end{lem} 
The explicit calculation of the above matrices $d_{t}, D_t$ is given in \cite{hurault2026geometry}.

\vspace{-0.2cm}
\paragraph{Score estimation error}
The objective of this section is to determine the influence of 
an inexact score on the trajectory mean $\EE[y_k]$ 
and covariance $\cov(y_k)$. Following the form of the score training  error shown in Theorem~\ref{thm:SGD_error}, we consider the following linear inexact score
\begin{equation} \label{eq:diff_inexact_score}
    v_{t}(x, \theta) = - A_{t}^\varepsilon x + b_{t}^\varepsilon, \quad \text{where}\quad 
    A_{t}^\varepsilon = A_{t} + \varepsilon \Delta_{t}, \quad b_{t}^\varepsilon = b_{t} + \varepsilon (\delta_{t} + \Delta_{t} \mu_{t}),
\end{equation}
where, for all $t \leq T$, \(\Delta_{t}\) and \(\delta_{t}\) are zero-mean random variables with values in $\RR^{d \times d}$ and $\RR^{d}$. Following Theorem~\ref{thm:SGD_error}, we assume that \(\Delta_{t}\) and \(\delta_{t}\) are uncorrelated. Moreover, we consider the idealized setting where $(\delta_t)_{t \geq 0}$ and $ (\Delta_t)_{t \geq 0}$ are  independent across time.  

In Proposition~\ref{prop:pertubed_diffusion} (Appendix~\ref{app:distance_general_diff}) we derive the second-order expansion, in $\varepsilon$, of the mean and covariance of the solution $q^\varepsilon_k = \mathcal{N}(\EE[y_{k}], \cov(y_k))$ of the perturbed discretized diffusion.
Using this expansion, we deduce in the following theorem the expansion of the average Wasserstein distance between $q_k^\varepsilon$ and the data distribution. It can be written as the sum of four distinct terms. The first term corresponds to the zeroth-order error, which arises from the fact that $T-t_k >0$. The second term represents the discretization error, which is not made explicit in this work. The third and fourth terms stem from the time-dependent perturbation of the parameters \(b_t\) and \(A_t\) in the linear score. 
\begin{thm}[Diffusion sampling error with inexact linear score, proof in Appendix~\ref{app:distance_general_diff}] \label{thm:distance_general_diff}
Under Assumption~\ref{ass:gaussian}, the discretized diffusion process~\eqref{eq:euler_disc} with score~\eqref{eq:diff_inexact_score} samples a Gaussian distribution ${q^\varepsilon_k = \mathcal{N}(\EE[y_{k}], \cov(y_k))}$ that satisfies, 
\begin{equation*}
\EE_{\delta, \Delta}  \left[ W_2^2(p_\mathrm{data},q_k^\varepsilon) \right] =  \mathcal{E}^{(0)}_{T-t_k} + \gamma \mathcal{E}^{\text{disc}}_{T-t_k} + \varepsilon^2 \gamma \left( \mathcal{E}_{T-t_k}^{\text{mean}}(\delta) + \mathcal{E}_{T-t_k}^{\text{cov}}(\Delta) \right) + o(\gamma, \varepsilon^2)
\end{equation*}
The explicit expressions of each term are provided in Appendix~\ref{app:distance_general_diff}. 
They involve the current timestep~$t_k$, the eigenvalues~\( (\lambda_i) \) of~\( C_{\mathrm{data}} \), as well as, for the terms $\mathcal{E}_{T-t}^{\text{mean}}(\delta)$ and $\mathcal{E}_{T-t}^{\text{cov}}(\Delta)$, the integral over $ s \in [T-t, T]$ of all the projections of the perturbations $\delta_s$ and $\Delta_s$ onto the eigenvector outer-product basis \( (u_i u_j^\top)_{ij} \) of the data.
\end{thm} 
This result closely resembles Theorem~\ref{thm:langevin_distance_general} for Langevin sampling. Unlike Langevin, it contains an additional discretization term from the small-$\gamma$ expansion. Moreover, the mean $\mu_{\mathrm{data}}$ of the data now influences the zeroth-order error $\mathcal{E}^{(0)}$. The score approximation errors $\mathcal{E}^{\text{mean}}$ and $\mathcal{E}^{\text{cov}}$ also share a similar structure. While Langevin uses a fixed noise level $\sigma$, diffusion models involve a time-varying noise level, making the score error timestep-dependent and yielding, at the first-order in~$\gamma$, an integral over the single-step errors, thereby reflecting the accumulation of the time-dependent score approximation error along the diffusion trajectory. 

\vspace{-0.2cm}
\paragraph{Diffusion model sampling with a linear score trained by SGD} We now make explicit the overall error of diffusion sampling, accounting for both the score approximation error from denoising score matching and the above diffusion error. More precisely, we use the linear score~\eqref{eq:diff_inexact_score} with parameters \( A_t \) and \( b_t \) trained, independently at each $t$, by SGD minimizing the loss~\eqref{eq:DSM_loss} at noise level \( \sigma_t \), with data rescaled by $s_t$. 
We use Theorem~\ref{thm:SGD_error} with $\mu \to s_t \mu = \mu_t$, $C_{\mathrm{data}} \to s_t^2 C_{\mathrm{data}}$, $\sigma \to s_t \sigma_t$ and get that, after SGD training, the parameters $(A_t, b_t)$ verify
\begin{equation*}
\begin{split}
    & \quad  \quad  \quad b_t = \Sigma_t^{-1}\mu_t + \delta_t + \Delta_t \mu_t \quad\text{and}\quad A_t = \Sigma_t^{-1} + \Delta_t, \\ 
    & \quad\text{where}\quad \delta_t \sim \mathcal{N}(0, \cov(\kappa_t)) \quad\text{and}\quad \Delta_t \sim \mathcal{N}(0, \cov(A_t)),
\end{split}
\end{equation*}
The expansions of $\cov(\kappa_t)$ and $\cov(A_t)$ for small SGD learning rate $\tau$ and large dataset size $N$ are calculated in Appendix~\ref{app:diffusion_final}. Using this score, we apply Theorem~\ref{thm:distance_general_diff} to get the following corollary.
\begin{cor}
\label{cor:diffusion_final}[Diffusion sampling error with linear score trained by SGD, proof in Appendix~\ref{app:diffusion_final}]
Under Assumption~\ref{ass:gaussian}, consider a linear score $v_t(x, \theta) = - A_t x + b_t$ with parameters \( \theta=(A_t,b_t)\) trained at each $t$ via the SGD algorithm~\eqref{eq:score_matching_SGD_emp}, with \( N \) data samples and constant learning rate \( \tau > 0 \). Then, at step $k$, the discretized diffusion algorithm~\eqref{eq:euler_disc} with score $v_t$ samples a Gaussian distribution $q_k$ that satisfies \vspace{-0.cm}
\begin{equation*}
\begin{split}
     \mathbb{E}_{\theta}&\left[W_2\left(p, q_k\right)^2\right] = 
 \mathcal{E}^{(0)}_{T-t_k} + \gamma \mathcal{E}^{\text{disc}}_{T-t_k} \\[-0.2cm] &+ \gamma \sum_{i,j = 1}^d \left(\tau k_{T-t_k}^\tau (\lambda_i, \lambda_j)  + \frac{\tau}{N} k_{T-t_k}^{\tau, N}(\lambda_i, \lambda_j) +\frac{1}{N} k_{T-t_k}^N(\lambda_i, \lambda_j)\right)  + o\left( \gamma, \tau, \frac{1}{N} \right)
\end{split} 
 \end{equation*}
where, for $t \leq T$, $k_t^\tau, k_t^N, k_t^{\tau, N} : \RR^d \times \RR^d \to \RR$ are kernel functions that capture the interactions between the eigenvalues of $C_\mathrm{data}$. Their explicit expressions are given in Appendix~\ref{app:diffusion_final}. 
\end{cor}
Similar to the Langevin error in Corollary~\ref{cor:langevin_final}, this result characterizes how the data spectrum, the score training parameters (learning rate $\tau$, number of data samples $N$), and the sampling parameters (stepsize $\gamma$; diffusion parameters $\beta_t$, $\xi_t$, $\alpha$) jointly influence performance. The score-related error takes again the form of kernel norms over the power spectrum of the data. Each kernel can be written as the time integral over \( s \in [T - t, T] \) of a time-dependent rational function of the \(\lambda_i\), capturing how errors interact with the data geometry progressively over the dynamics. In particular, this result highlights a trade-off on the choice of the final diffusion timestep $t_k$. More precisely, for small $T-t \to 0$, the term $\mathcal{E}^{(0)}_{T-t}$ behaves as
$\mathcal{O}\left(\sigma_{T-t}^4\right)$ in both the diffusion VE and VP settings discussed in Section~\ref{sec:background}.
On the other hand, the terms $k_{T-t}^\tau$ and $k_{T-t}^{\tau,N}$ diverge at rate $\mathcal{O}\left(\int_{T-t}^T \frac{1}{\sigma_s^2}ds\right)$. 

\begin{figure}[h]
  \centering 
  \includegraphics[width=0.45\textwidth]{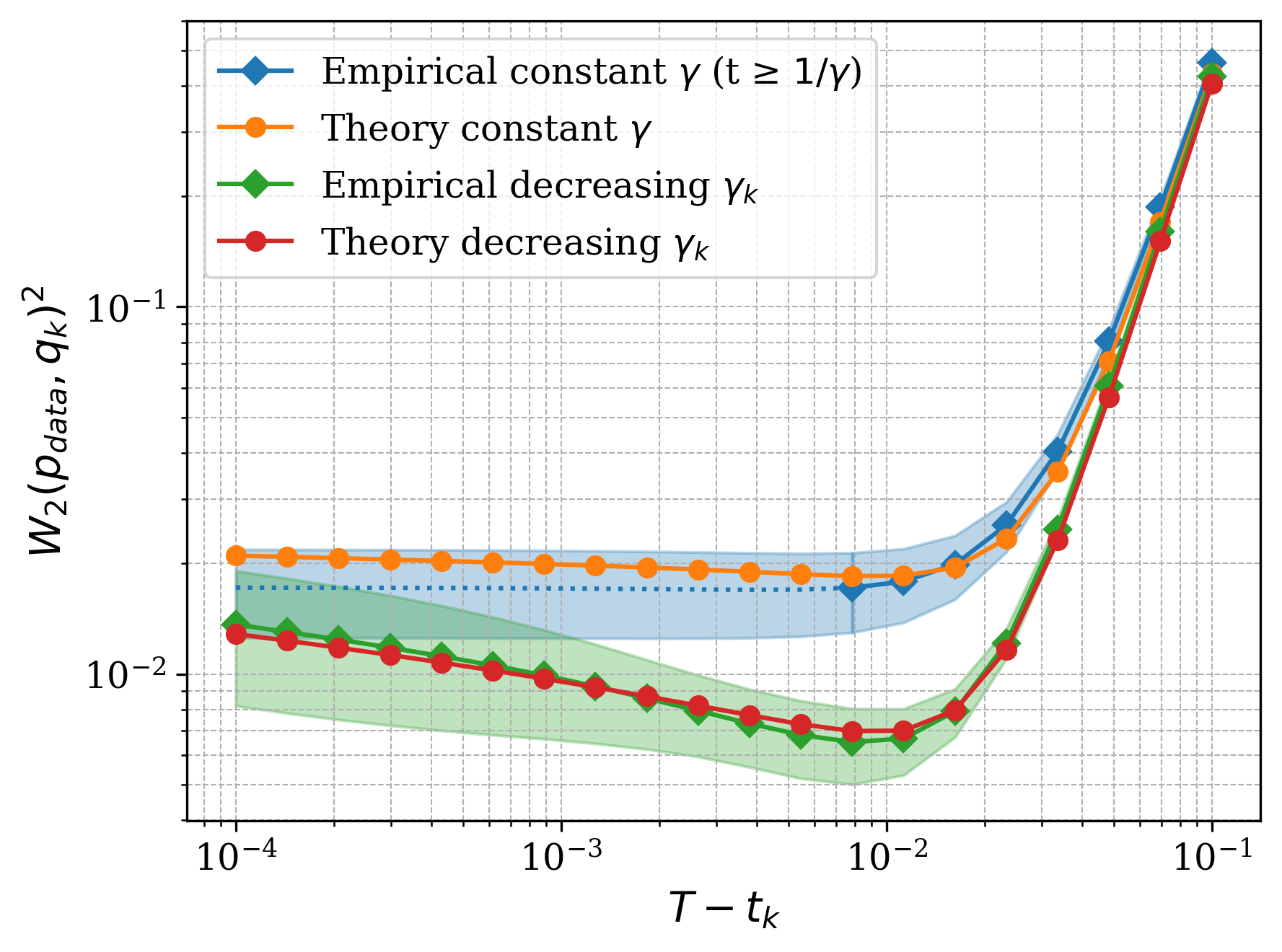}
  \caption{Theoretical and empirical $W_2$ errors of discretized diffusion at step $k$, with respect to $T-t_k$,  with constant stepsize~$\gamma$ and (exponentially) decreasing stepsize~$\gamma_k$.}
  \label{fig:diffusion} 
\end{figure}

\paragraph{Numerical validation} We verify numerically this result in Figure~\ref{fig:diffusion} with the same setting as in Figure~\ref{fig:langevin}. We consider here the VP diffusion model ($\xi_t = \beta_t$, with the same linear evolution of $\beta_t$ as the one proposed by~\cite{song2020score}) and with $\alpha = 1$. For each timestep $t$, $A_{t}$ and $b_{t}$ are sampled from the theoretical Gaussians from Theorem~\ref{thm:SGD_error} with ${\tau = 10^{-2}}$ and ${N=100}$. For each model, we sample $10^6$ particles, with $k$ diffusion iterations of the algorithm~\eqref{eq:euler_disc} with stepsize $\gamma = 10^{-2}$, and show in Figures~\ref{fig:diffusion} the empirical $W_2$ distance with the data distribution, with respect to $T-t_k$. This gives the curve \emph{Empirical constant~$\gamma$} which is compared against the theoretical error from Corollary~\ref{cor:diffusion_final} \emph{(Theory constant~$\gamma$)}. 

In this setting, the rates of convergence identified below Corollary~\ref{cor:diffusion_final} simplify to $\mathcal{O}\left(\sigma_{T-t}^4\right) = \mathcal{O}((T-t)^2)$ and $\mathcal{O}\left(\int_{T-t}^T \frac{1}{\sigma_s^2}ds\right) = \mathcal{O}(\log(T-t))$. We indeed verify an initial \emph{quadratic} decrease of the error as $T-t \to 0$. However, using a uniform discretization stepsize \(\gamma\), we cannot have~\(T - t_k\) smaller than~\(\gamma\), and the divergence for small $T-t_k$ identified in Corollary~\ref{cor:diffusion_final} is not observed. To address this, we also plot the error obtained with a diffusion discretized with non-uniform exponentially decreasing stepsizes~\(\gamma_k\). The adaptation of the theoretical analysis to this case is discussed in Appendix~\ref{app:non-uniform-gamma}. With this discretization, we reduce the optimal sampling error and numerically observe the logarithmic increase of the $W_2$ error as $T-t_k \to 0$. We give further insights on this trade-off in Appendix~\ref{app:analysis_diffusion}.

\section*{Conclusion}

This work provides a first analytical step toward understanding how the data geometry, encoded by the power spectrum, interacts with key parameters of Langevin and diffusion model sampling pipelines. While limited to linear models and Gaussian data, the framework uncovers structural dependencies that offer valuable guidance about the trade-off to select various parameters such as the noise parameter of Langevin or the stopping times of diffusion models. Extending these results to nonlinear neural networks remains an open and difficult challenge.

\bibliographystyle{plainnat}
\bibliography{refs}

\appendix
\newpage

\section{Denoising Score Matching error}
\subsection{Proof of Theorem~\ref{thm:SGD_error}} 
\label{app:SGD}
\newtheorem*{repeatthm1}{Theorem 1}
\begin{repeatthm1}[Optimization and generalization errors in SGD for denoising score matching] 
Under Assumption~\ref{ass:gaussian}, for fixed stepsize $\tau < \frac{2}{\max\left( \lambda_{max} + \sigma^2, 1 \right)},$ the SGD iterates~\eqref{eq:score_matching_SGD_emp} converge to a stationary distribution on the parameters $A$ and $b$. At stationarity, $b = \kappa + A \mu_\mathrm{data}$, with $\kappa$ and $A$ uncorrelated random variables with means
\begin{align*}
 \EE[\kappa] =  o\left( \frac{1}{N}\right)\quad\text{and}\quad \EE[A] = C_\sigma^{-1} + o\left( \frac{1}{N}\right)
\end{align*}
and covariances (denoting $P_\sigma \eqdef C_\sigma^{-1} C_\mathrm{data}$ and $Q_\sigma \eqdef C_\sigma^{-2}C_\mathrm{data}$)
\begin{equation*}
\begin{split}
\cov(\kappa) &= \frac{\tau}{2 \sigma^2} P_\sigma +  \frac{1}{N} Q_\sigma + \frac{\tau}{N \sigma^2} \left( \frac12 \langle P_\sigma, C_\mathrm{data} \rangle \id -  P_\sigma^2 \right) 
+ o \left(\tau, \frac{1}{N}\right) 
\end{split}
\end{equation*}
\begin{equation*}
\begin{split}
\cov(A)  = \frac{\tau}{2\sigma^2} P_\sigma \otimes \id + \frac{1}{N} \left( Q_\sigma \otimes Q_\sigma \right) (I_{d^2} + P_{trans})  -  \frac{\tau}{N \sigma^2} P_\sigma^2 \otimes \id  + o\left(\tau, \frac{1}{N}\right) .
\end{split}
\end{equation*}
$P_{trans}$ denotes the transpose in the vectorized space $\RR^{d^2}$  i.e. $P_{trans}[vec(X)] = vec(X^\top)$.
\end{repeatthm1}

\begin{proof}
We first treat only the optimization error ($N = \infty$) and then consider the generalization error in a second part ($N < \infty$). 
\paragraph{Optimization error} We first consider that, in the SGD iterates~\eqref{eq:score_matching_SGD_emp}, that the $x_k$ are sampled from the true data distribution $p_\mathrm{data}$, as in~\eqref{eq:basic_sgd}. This is the limit $N \to \infty$. The optimization error is then given in the following Proposition~\ref{prop:optim_error}. 
\begin{prop}[Optimization error of SGD] \label{prop:optim_error}
Assuming a Gaussian data distribution ${p \sim \mathcal{N}(\mu_\mathrm{data}, C_\mathrm{data})}$, for fixed stepsize $$\tau < \frac{2}{\max\left( \max_{k\leq d}{\lambda_k} + \sigma^2, 1 \right) },$$ the SGD iterates~\eqref{eq:basic_sgd} converge  to a stationary distribution on the parameters $A$ and $b$. At stationarity, there is $\kappa$, uncorrelated to $A$, such that
$$ b = \kappa + A \mu_\mathrm{data} .$$
The means and covariances of $\kappa$ and $A$ are
\begin{align*}
 \EE[\kappa] = 0 \quad\text{and}\quad \EE[A] = C_\sigma^{-1}
\end{align*}
\begin{align*}
\cov(\kappa) &=  \frac{\tau}{2\sigma^2} C_\sigma^{-1}C_\mathrm{data} + o(\tau) \\
\cov(A) &=  \frac{\tau}{2\sigma^2}  (C_\sigma^{-1}C_\mathrm{data}) \otimes \id + o(\tau).
\end{align*}
\end{prop}
\begin{proof}
In the first part of the following proof (until Lemma~\ref{lem:SGD_2}), we do not need to assume that the data are Gaussian. We start with the change of variable
\begin{align*}
    \left \{ \begin{aligned}
        \hat x &= x - \mu \\
        z &= (\hat x+\sigma w, 1) \\
        y &= - \frac{1}{\sigma}w .
    \end{aligned}
    \right.
\end{align*}
We define 
\begin{align*}
    C_z \eqdef  \EE[zz^\top] = 
 \begin{bmatrix}
E[\hat x \hat x^\top] + \sigma^2 \id  & 0 \\
0 & 1
\end{bmatrix}
= 
 \begin{bmatrix}
C_\sigma   & 0 \\
0 & 1
\end{bmatrix}  
\end{align*}
and for $i \in \llbracket 1, d \rrbracket$,
\begin{align*}
\theta^*_i \eqdef C_z^{-1} \nu_i, \quad\text{where}\quad
\nu_i \eqdef \mathbb{E}[y_i z] = - \mathbb{E}[(w_i(\hat x + \sigma w), w_i)] = - (e_i , 0) \in \RR^{d+1} .
\end{align*}

We also denote, for $i \in \llbracket 1, d \rrbracket$,  $\theta_i =
\begin{bmatrix}
-a_i \\ b_i - a_i^\top \mu
\end{bmatrix} \in \RR^{d+1}$, where $a_i$ is the $i^{th}$ column of $A$, such that ${\theta = [\theta_1^\top, .., \theta_d^\top]^\top  =\operatorname{vec} \begin{bmatrix} -A^\top \\ (b - A \mu)^\top \end{bmatrix} \in \mathbb{R}^{n = (d+1)d}}$. The score matching loss~\eqref{eq:DSM_loss} can be rewritten as the sum of the $d$ least square regression objectives:
\begin{align*}
    F(\theta) =  \sum_{i=1}^d    \mathbb{E}_{z, y_i} \left[ f(z, y_i, \theta_i) \right] 
    = \frac{1}{2} \sum_{i=1}^d  \|  C_z^{1/2}( \theta_i - \theta_i^*) \|_{\RR^{d+1}}^2 + \text{cst}
\end{align*}
where $f : \RR^{d+1} \times \RR \times \RR^n : (z, y, \theta) \to \frac{1}{2}( \dotp{z}{\theta_i} - y_i )^2$.
The SGD iterations \eqref{eq:basic_sgd} can be rewritten for each~$\theta_i$ as: 
\begin{align}
\text{For $0 \leq i \leq d$,} \quad (\theta_i)_{k+1} &= (\theta_i)_k - \tau \nabla_\theta f(z_k, (y_i)_k, (\theta_i)_k) \\
    &=  (\theta_i)_k - \tau \mathbb{E}_{y,z}[\nabla_\theta f(z, y, (\theta_i)_k)] + \tau \varepsilon_k \\
    &= (\theta_i)_k - \tau \nabla_{\theta_i} F(\theta) + \tau \varepsilon_k \label{eq:SGD_general} .
\end{align}
It corresponds to a gradient descent step on $F$ plus a zero-mean random noise $\varepsilon_k =  \varepsilon((y_i)_k, z_k, (\theta_i)_k)$ defined by
\begin{align} \label{eq:eps_fn}
    \varepsilon(y_i, z, \theta_i) &=  \nabla_{\theta_i} f(y_i,z,\theta_i) - \EE_{y, z}[\nabla_{\theta_i} f(y_i,z,\theta_i)] \in \RR^{d+1} .
\end{align}
The following lemma indicates that the $(\theta_i)_k$ converge towards the sample of a distribution with mean the optimal $\theta_i^* = C_z^{-1} \mathbb{E}[y_i z]$ and with a covariance that depends on the fourth moment of the data. In this lemma, we make use of the matrix operator, defined for $C \in \RR^{p \times p}$ as
\begin{equation} \label{eq:lyap_extend}
\begin{split}
        L^\tau_C :  \  \RR^{p\times p} &\to  \RR^{p\times p} \\
     X &\mapsto  CX + XC^\top - \tau C X C^\top.
\end{split}
\end{equation}
Note that for $C \in S^{++}_d$ and $\tau < \frac{2}{\max eig(C)}$, its inverse is well defined. 
\begin{lem} \label{lem:SGD_1}
The algorithm~\eqref{eq:SGD_general} defines a Markov chain that admits, for fixed stepsize $$\tau < \frac{2}{\max\left( \max_{k\leq d}{\lambda_k} + \sigma^2, 1 \right) }$$ a unique stationary distribution $\pi_\tau$, with first and second moments, for 
$i,j \in \llbracket 1, d \rrbracket$,
\begin{align*}
\EE_{\pi_\tau}[\theta_i]  = \theta_i^* 
\end{align*}
and
\begin{align} \label{eq:sigma_Linv}
\Sigma^{ij} = \EE_{\pi_\tau}\left[(\theta_i -  \theta_i^*)(\theta_j - \theta_j^*)^\top\right] = \tau (L_{ C_z}^\tau)^{-1} [\Sigma_\epsilon^{ij}] 
\end{align}
where
\begin{align}
\Sigma_\epsilon^{ij} &\eqdef \EE_{\theta \sim \pi, (y, z)} [  \varepsilon(y_i,z,\theta_i) \varepsilon(y_j, z, \theta_j)^\top ] \\
&= \EE[y_i y_j z z^\top] + \EE[z z^\top \theta_i^*(\theta_j^*)^\top z z^\top] - \EE[y_i z (\theta_j^*)^\top z z^\top] - \EE[y_j z (\theta_i^*)^\top z z^\top] \label{eq:sigma_eps_lemma}
\end{align}
\end{lem}
\begin{proof}
We start by making explicit the noise function $\varepsilon$ in \eqref{eq:eps_fn}
\begin{align*}
    \epsilon(y_i, z, \theta_i) &=  \nabla_{\theta_i} f(y_i,z,\theta_i) - \EE[\nabla_{\theta_i} f(y_i,z,\theta_i)] \in \RR^{d+1} \\
    &= \langle z, \theta_i \rangle z - y_i z - \EE[\langle z, \theta_i \rangle z - y_i z] \\
    &= (zz^\top  - C_z) \theta_i - y_i z + \nu_i \\
    &=  (zz^\top  - C_z) \theta_i - y_i z + C_z \theta^*_i  \\
    &= (zz^\top - C_z)(\theta_i - \theta^*_i) + (z^\top \theta^*_i- y_i)z .
\end{align*}
It satisfies $\mathbb{E}_{y,z}[\epsilon(y_i,z,\theta_i)] = 0$ and, by independence between $\theta_k$ and $z_k$,  $\epsilon_k = \epsilon((y_i)_k, z_k, (\theta_i)_k)$ is independent of $\theta_k$.

From~\cite[Proposition 2, Example 1]{dieuleveut2020bridging}, for $\gamma < \frac{2}{L}$, with $L = \lambda_{max}(C_z)$, the Markov chain $(\theta_k)_{k \geq 0}$ defined in~\eqref{eq:SGD_general} admits a unique stationary distribution $\pi_\tau$. We now make explicit the first and second-order moments of this distribution.  Using that $\EE[\nabla_{\theta_i} F(\theta)] = C_z( \EE[\theta_i] - \theta_i^*) $, taking the expectation on both sides of~\eqref{eq:SGD_general}, we have after convergence, for each $0 \leq i \leq d$,
\begin{align*}
\EE_{\pi_\tau}[\theta_i] = \EE_{\pi_i}[\theta_i] - \tau C_z( \EE_{\pi}[\theta_i] - \theta_i^*) \Longrightarrow \EE_{\pi}[\theta_i] = \theta_i^* .
\end{align*}
For the second-order moment, again from~\eqref{eq:SGD_general}, for $0 \leq i,j \leq d$, the covariance ${\Sigma^{ij} \eqdef \EE_{\pi_\tau}\left[(\theta_i -  \theta_i^*)(\theta_j - \theta_j^*)^\top\right]}$ satisfies, by independence between $\theta_k$ and $\epsilon_k$,
\begin{align*}
&\Sigma^{ij} = \Sigma^{ij} - \tau \left( \EE\left[(\nabla_{\theta_i} F(\theta))(\theta_j -  \theta_j^*)^\top\right] + \EE\left[(\theta_i -  \theta_i^*)  (\nabla_{\theta_j} F(\theta))^\top\right)\right] \\ \nonumber &+ \tau^2 \EE\left[(\nabla_{\theta_i} F(\theta)) (\nabla_{\theta_j} F(\theta))^\top \right] + \tau^2   \EE_{\pi_\tau, y,z}[\epsilon(y_i,z, \theta_i) \epsilon(y_j, z, \theta_j)^\top ] \\
&\Leftrightarrow C_z \Sigma^{ij} +  \Sigma^{ij} C_z - \tau C_z \Sigma^{ij} C_z  = \tau \EE_{\pi_\tau, y,z}[\epsilon(y_i,z, \theta_i) \epsilon(y_j, z, \theta_j)^\top ].
\end{align*} 
This can be rewritten as 
\begin{align*}
\Sigma^{ij} = \tau (L_{ C_z}^\tau)^{-1} [\Sigma_\epsilon^{ij}] 
\end{align*}
where, by independence between $\theta_i$ and $z$,
\begin{align*}
\Sigma_\epsilon^{ij} &= \EE_{\theta \sim \pi_\tau, (y, z)} [  \epsilon(y_i,z,\theta_i) \epsilon(y_j, z, \theta_j)^\top ] \\
&= \EE_{(y, z)} [  (z^\top \theta^*_i- y_i)z z^\top (z^\top \theta^*_j- y_j)^\top] \\
&= \EE[y_i y_j z z^\top] + \EE[z z^\top \theta_i^*(\theta_j^*)^\top z z^\top] - \EE[y_i z (\theta_j^*)^\top z z^\top] - \EE[y_j z (\theta_i^*)^\top z z^\top] .
\end{align*}
\end{proof}

 It is important to note that \textbf{at this stage, no Gaussian assumption on the data is required}. However, assuming Gaussian data becomes necessary to explicitly compute the fourth moment of the data, which appears in the second term of~\eqref{eq:sigma_eps_lemma}. Next, we evaluate each term in~\eqref{eq:sigma_eps_lemma} using their index-wise decomposition. Specifically, we apply Isserlis' theorem to compute the fourth moment of the Gaussian data distribution. The result is summarized in the following lemma. 
\begin{lem} \label{lem:SGD_2} Under the Gaussian data assumption~\ref{ass:gaussian}, using the previous notations, the noise matrix $\Sigma_\epsilon^{ij}$ defined for $i,j \in \llbracket 1, d \rrbracket$ in equation~\eqref{eq:sigma_eps_lemma} simplifies to
\begin{align*}
\Sigma^{ij}_\epsilon = (C_\sigma^{-1}C_\mathrm{data})_{ij} C_z.
 \end{align*}
\end{lem}
\begin{proof}
We calculate each term inside 
\begin{align} \label{eq:sigma_eps}
   \Sigma_\epsilon^{ij}(\theta^*) = \underbrace{\EE[y_i y_j z z^\top]}_{\text{term } 1} + \underbrace{\EE[z z^\top \theta_i^*(\theta_j^*)^\top z z^\top]}_{\text{term } 2} - \underbrace{\EE[y_i z (\theta_j^*)^\top z z^\top]}_{\text{term } 3} - \underbrace{\EE[y_j z (\theta_i^*)^\top z z^\top]}_{\text{term } 4}
\end{align}
We first rewrite $z$ as 
\begin{align*}
z = 
\begin{bmatrix}
\hat x + \sigma  w\\
0
\end{bmatrix}
 + 
\begin{bmatrix}
0 \\
1
\end{bmatrix}
 = 
\hat{z}
 + 
r
\end{align*}
where $\hat z \sim \mathcal{N}(0, \id_{\RR^{d+1}})$ has zero mean. Recall that each $y_i = -\frac{w_i}{\sigma} \sim \mathcal{N}(0, \sigma^{-2}\id)$ and that we denoted $\nu_i = \EE[y_i z]$. We compute each term in~\eqref{eq:sigma_eps} using its index decomposition. 
\paragraph{Term 1.} Developing $z$ and using that the expectation of the product of three zero-mean Gaussian is~$0$, we get, for $0 \leq k,l \leq d+1$,
    \begin{align*}
    \EE[y_i y_j z z^\top]_{kl} &=\EE[y_i y_j z_k z_l]   \\
    &= \EE[y_i y_j]\EE[\hat z_k \hat z_l] + \EE[y_i \hat z_k]\EE[y_j \hat z_l] + \EE[y_i \hat z_l]\EE[y_j \hat z_k] + \EE[r_k r_l y_i y_j]   \\
    &= \frac{1}{\sigma^2}\mathbf{1}_{i=j} (\EE[\hat z_k \hat z_l] + r_k r_l) + (\nu_i)_k (\nu_j)_l +  (\nu_i)_l (\nu_j)_k ,
    \end{align*}
such that 
\begin{align*}
 \EE[y_i y_j z z^\top] = \frac{1}{\sigma^2} \mathbf{1}_{i=j} C_z  + \nu_i \nu_j^\top + \nu_j \nu_i^\top.
 \end{align*}
\paragraph{Term 2.} It involves the fourth moment of $z$. 
\begin{align*}
   \EE[z z^\top \theta_i^*(\theta_j^*)^\top z z^\top] 
    &= \left( \sum_{k,l = 1}^{d+1} (C_z^{-1} \nu_i \nu_j^\top C_z^{-1})_{k,l}  \mathbb{E}(z_p z_q z_k z_{\ell}) \right)_{p,q} \\ &= \Lambda(\theta_i^*(\theta_j^*)^\top)
\end{align*}
where $\Lambda(S) := E[z z^\top S z z^\top] $. We denote $C_{\hat z} = \EE[\hat z \hat z^\top]$. For  $0 \leq p,q,k,l \leq d+1$, using Isserlis' theorem and the fact that the expectation of the product of three zero-mean Gaussian variables is zero,
\begin{align*}
 \mathbb{E}(z_p z_q z_k z_{\ell}) &=  (C_{\hat z})_{p,q} (C_{\hat z})_{k,\ell} + (C_{\hat z})_{p,k} (C_{\hat z})_{q,\ell} + (C_{\hat z})_{p,\ell} (C_{\hat z})_{q,k} + r_p r_q r_k r_l \\
 &+ r_p r_q (C_{\hat z})_{k,\ell} + r_p r_k (C_{\hat z})_{q,\ell} + r_p r_l (C_{\hat z})_{q,k} \\
 &+  r_q r_k (C_{\hat z})_{p,\ell} + r_q r_l (C_{\hat z})_{p,k} + r_k r_l (C_{\hat z})_{p,q} .
\end{align*}
Using that 
\begin{align*}
C_{\hat z} = C_z - rr^\top,
\end{align*}
the above expression simplifies to 
\begin{align*}
 \mathbb{E}(z_p z_q z_k z_{\ell}) &=  (C_{ z})_{p,q} (C_{ z})_{k,\ell} + (C_{ z})_{p,k} (C_{z})_{q,\ell} + (C_{z})_{p,\ell} (C_{z})_{q,k} - 2 r_p r_q r_k r_l.
\end{align*}
We can then  simplify $\Lambda(S)$
\begin{align*} 
	\Lambda(S) &= C_{z} S C_{z} + C_{z} S^\top C_{z} +  \dotp{C_{z}}{S} C_{z} - 2 \langle rr^\top, S \rangle  r r^\top  ,
\end{align*}
and for $S = \theta_i^*(\theta_j^*)^\top$, 
\begin{align} 
	\text{term 2} = \Lambda(\theta_i^*(\theta_j^*)^\top) &= C_z S C_z + C_z S^\top C_z + \dotp{C_z}{S} C_z - 2 \langle rr^\top, S \rangle  r r^\top \\
	&= \nu_i \nu_j^\top + \nu_j \nu_i^\top + \langle C_z^{-1}, \nu_i \nu_j^\top \rangle C_z - 2 \langle C_z^{-1} rr^\top C_z^{-1},   \nu_i \nu_j^\top  \rangle  r r^\top \label{eq:Lambda} .
\end{align}
We have that
$C_z^{-1}
 = 
\begin{bmatrix} 
C_\sigma^{-1} & 0 \\
0 & 1
\end{bmatrix}$, such that $C_z^{-1}  r = \begin{bmatrix}
0\\
1
\end{bmatrix}$ and
\begin{align*}
C_z^{-1}  r  r^\top  C_z^{-1}
 = 
\bigl(C_z^{-1}r\bigr)  \bigl(r^\top C_z^{-1}\bigr)
 = 
\underbrace{
\begin{bmatrix}
0 & \cdots & 0 & 0\\
\vdots & \ddots & \vdots & \vdots\\
0 & \cdots & 0 & 0\\
0 & \cdots & 0 & 1
\end{bmatrix}
}_{d+1}.
\end{align*}
Using the fact that $\nu_i \nu_j^\top$ is zero everywhere except at the position $(i,j)$ for $1 \leq i,j \leq d$, we get that
\begin{align*}
   \langle  C_z^{-1}  r  r^\top  C_z^{-1}, \nu_i \nu_j^\top \rangle = 0
\end{align*}
and thus the last term in \eqref{eq:Lambda} is zero and 
\begin{align*}
	\Lambda(S) &= \nu_i \nu_j^\top + \nu_j \nu_i^\top + \langle C_z^{-1}, \nu_i \nu_j^\top \rangle C_z . 
\end{align*}
\paragraph{Terms 3 and 4.} For $0 \leq k,l \leq d+1$,
\begin{align*}
(E[y_i (\theta_j^*)^\top z  z z^\top])_{kl} = \sum_{m=1}^{d+1}
 (\theta_j^*)_m\EE[y^i z_m z_k z_l].
 \end{align*}
Now again with Isserlis' theorem, using that $y$ has zero-mean, for $0 \leq k,l,m \leq d+1$,
\begin{align*}
\EE[y_i z_m z_k z_l] 
&= \EE[y_i z_m] (C_z)_{k,l} + \EE[y_i  z_k]  (C_z)_{m,l} +  \EE[y_i z_l]  (C_z)_{k,m} \\
&= (\nu_i)_m (C_z)_{k,l} + (\nu_i)_k (C_z)_{m,l} + (\nu_i)_l (C_z)_{k,m}
 \end{align*}
 and therefore, using $\theta_j^* = C_z^{-1}\nu_j$, 
 \begin{align*}
 E[y^i (\theta_j^*)^\top z  z z^\top] = \langle C_z^{-1}, \nu_i \nu_j^\top \rangle C_z + \nu_i \nu_j^\top + \nu_j \nu_i^\top.
 \end{align*}
We get the fourth term by inverting $i$ and $j$:
 \begin{align*}
 E[y^j (\theta_i^*)^\top z  z z^\top] = \langle C_z^{-1}, \nu_j \nu_i^\top \rangle C_z + \nu_i \nu_j^\top + \nu_j \nu_i^\top .
 \end{align*}
All in all, merging the four terms, we get
\begin{align*}
\Sigma^{ij}_\epsilon &= \left( \frac{1}{\sigma^2}\mathbf{1}_{i=j} - \langle C_z^{-1}, \nu_j \nu_i^\top \rangle \right) C_z \\
&= \frac{1}{\sigma^2} (\id - \sigma^2 C_\sigma^{-1})_{ij} C_z \\
&= \frac{1}{\sigma^2} (C_\sigma^{-1}C_\mathrm{data})_{ij} C_z .
 \end{align*}

We finally turn to the calculation of the covariance~\eqref{eq:sigma_Linv}. First, we decompose $(L_{ C}^\tau)^{-1}$ in the eigenbasis of the data using the following lemma.
\begin{lem} \label{lem:SGD_3} For $C \in S^{++}_d$, denoting $C = U \diag{\gamma_i}_{1 \leq i\leq p} U^\top$ the diagonalization of $C$ in its eigenbasis, the inverse operator of $L^\tau_C$ defined in ~\eqref{eq:lyap_extend} decomposes as 
\begin{align*}
   (L_{ C}^\tau)^{-1}[X] = U \left( U^\top X U \odot \left(\frac{1}{\gamma_i + \gamma_j - \tau \gamma_i \gamma_j}\right)_{1 \leq i,j \leq p} \right) U^\top.
\end{align*}
\end{lem}
Denoting $\lambda^z_k$ the eigenvalues of $C_z$, applying the above lemma applied to \eqref{eq:sigma_Linv}, and keeping only the first-order expansion in $\tau$,
\begin{align*}
    \Sigma^{ij} &= \EE_{\pi_\tau}\left[(\theta_i -  \theta_i^*)(\theta_j - \theta_j^*)^\top\right] \\&= \frac{\tau}{\sigma^2} \left(C_\sigma^{-1} C_\mathrm{data}\right)_{ij} U \left(\operatorname{Diag}\left(\lambda^z_k\right) \odot \left(\frac{1}{\lambda^z_k + \lambda^z_l -  \tau \lambda_k^z \lambda^z_k}\right)_{1\leq k,l\leq d+1} \right)U^\top  \\
    &= \frac{\tau}{\sigma^2} \left(C_\sigma^{-1} C_\mathrm{data}\right)_{ij} U \operatorname{Diag}\left(\frac{1}{2 - \tau\lambda^z_k}\right)_{1\leq k,l\leq d+1} U^\top \\
    &= \frac{\tau}{2 \sigma^2} \left(C_\sigma^{-1} C_\mathrm{data}\right)_{ij} \id_{\RR^{d+1}} + o(\tau) . 
\end{align*}

We can finally extract the covariances of $A$ and $b$ from $\Sigma^{ij}$ using the fact that $\theta_i =
\begin{bmatrix}
-a_i \\ b_i - a_i^\top \mu
\end{bmatrix}$.
As $\Sigma^{ij}$ is diagonal, for each $i,j \in \llbracket 1, d \rrbracket$, $a_i$ and $b_j - a_j^\top \mu$ are uncorrelated. Denoting, $\kappa =  b - A \mu$, $\kappa$ is uncorrelated to $A$ and we get
\begin{align*}
\cov(A) &= \left( \cov(a_i, a_j) \right)_{ij} = (\Sigma^{ij})_{k,l\leq d}= \frac{\tau}{2\sigma^2} (C_\sigma^{-1} C_\mathrm{data}) \otimes \id + o(\tau) 
\end{align*}
and
\begin{align*}
\cov(\kappa) &= \left( \cov(\kappa_i, \kappa_j) \right)_{ij} =(\Sigma^{ij})_{k,l\leq d}
= \frac{\tau}{2 \sigma^2} C_\sigma^{-1} C_\mathrm{data}.  
+ o(\tau) 
\end{align*}
\end{proof}

\paragraph{Generalization error} 
Equipped with the previous optimization error estimation, we now consider that at each iteration the $x_k$ are sampled from the empirical Gaussian distribution~\eqref{eq:pN}. First, for a fixed dataset $\{x_i\}_{i \leq N}$, we can follow the same calculations as in Proposition~\ref{prop:optim_error} by replacing the true data mean and data covariance by these empirical ones. By denoting $A_N$ and $b_N$ the parameters trained in this context, we get that, after convergence, the SGD updates converge to a stationary distribution on the parameters $A_N$ and $b_N$ that satisfies 
$b_N = \kappa_N + A_N \hat \mu_N$, with $\kappa_N$ and $A_N$ uncorrelated and with means
\begin{align} \label{eq:exp_bN}
 \EE_{SGD}[\kappa_N] &= 0
 \\ \EE_{SGD}[A_N] &= (\hat C_\sigma^N)^{-1} 
\end{align}
and covariances
\begin{align}
\cov_{SGD}(\kappa_N) &=  \frac{\tau}{2\sigma^2} (\hat C^N_\sigma)^{-1} \hat C_N \\
\cov_{SGD}(A_N) &=  \frac{\tau}{2\sigma^2}  \left( (\hat C^N_\sigma)^{-1} \hat C_N \right) \otimes \id  \label{eq:exp_AN_h}
\end{align}
where $\hat C^N_\sigma \eqdef \hat C_N + \sigma^2 \id$.
We now also take into account the randomness from the sampling of the dataset~$\{x_i\}_{i \leq N}$. Each $x_i$ is an i.i.d. random variable with Gaussian distribution $p_\mathrm{data}$. The empirical mean and covariance are then random variables. Using the central limit theorem, we can get the first-order Taylor expansion of these empirical means and covariances (and know that the second-order term has zero-mean) :
\begin{lem} \label{lem:SGD_4}
Let $\{x_i\}_{i \leq N}$  i.i.d. random variables with distribution $p_\mathrm{data}$. Then $\hat \mu_N$ and $\hat C_N$ verify
\begin{align} \label{eq:muN}
\hat  \mu_N &= \mu + \frac{1}{\sqrt{N}} \mu^{(1)} + \frac{1}{N}\mu^{(2)} + o\left(\frac{1}{N}\right) \quad \text{where} \quad \mu^{(1)} \sim \mathcal{N} (0, C_\mathrm{data}) \quad \text{and} \quad \EE[\mu^{(2)}] = 0 \\
 \label{eq:CN}
\hat C_N &=  C_\mathrm{data} + \frac{1}{\sqrt{N}}   C^{(1)} + \frac{1}{N}   C^{(2)} + o\left( \frac{1}{N}\right) \quad \text{where} \quad   C^{(1)} \sim \mathcal{N} (0, \mathcal{V}) \quad \text{and} \quad \EE[  C^{(2)}] = 0
\end{align}
and where $\mu^{(1)}$, $\mu^{(2)}$, $C^{(1)}$ and $C^{(2)}$ are uncorrelated. We denote
\begin{align*}
    \mathcal{V} = (C_\mathrm{data} \otimes C_\mathrm{data})  ( I_{d^2} + P_{trans}) \in \RR^{d^2 \times d^2}
\end{align*}
where $P_{trans}$ is the commutation matrix that realizes the transpose operation in vectorized form, and $ I_{d^2}$ is the identity matrix on the vectorized space.
\end{lem}
\begin{proof}
By the central limit theorem, the empirical mean satisfies 
\begin{align*}
    \sqrt{N}  (\hat\mu_N - \mu)  \xrightarrow{d} \mathcal{N}(0, C_\mathrm{data}) 
\end{align*}
that is to say 
\begin{align*} 
\hat \mu_N = \mu + \frac{1}{\sqrt{N}} \mu^{(1)} + o\left(\frac{1}{\sqrt{N}}\right) \quad \text{where} \quad \mu^{(1)} \sim \mathcal{N} (0, C_\mathrm{data})
\end{align*}

Note that the higher-order terms are not necessarily Gaussian but will always have zero-mean. In the following, in order to fully take into account the generalization error, we will need the second-order decomposition of $\mu_N$ in $\frac{1}{\sqrt{N}}$. Therefore, we write 
\begin{align*} 
\hat  \mu_N = \mu + \frac{1}{\sqrt{N}} \mu^{(1)} + \frac{1}{N}\mu^{(2)} + o\left(\frac{1}{N}\right) \quad \text{where} \quad \mu^{(1)} \sim \mathcal{N} (0, C_\mathrm{data}) \quad \text{and} \quad \EE[\mu^{(2)}] = 0
\end{align*}
We can also get a similar central limit result for the empirical covariance. 
\begin{align*}
    \sqrt{N} \operatorname{vec} ( \hat C_N - C_\mathrm{data}) \xrightarrow{d} \mathcal{N} (0, \mathcal{V}),
\end{align*}
where
\begin{align*}
\mathcal{V} &= \EE[(xx^\top - C_\mathrm{data}) \otimes (xx^\top - C_\mathrm{data}) ] \\ 
&= \EE[(xx^\top) \otimes (xx^\top) ] - C_\mathrm{data} \otimes C_\mathrm{data}.
\end{align*}
Using Isserlis' theorem, for $1 \leq i,j,k,l \leq d$, we get indexwise
\begin{align*}
\mathcal{V}_{ijkl} = (C_\mathrm{data})_{jl}  (C_\mathrm{data})_{ik} + (C_\mathrm{data})_{il}  (C_\mathrm{data})_{jk},
\end{align*}
i.e.~for $Z \in \RR^{d \times d}$
\begin{align*}
    \mathcal{V} vec(Z) = (C_\mathrm{data} \otimes C_\mathrm{data}) vec(Z + Z^\top) .
\end{align*}
We can thus rewrite $\mathcal{V}$ as 
\begin{align*} 
    \mathcal{V} = (C_\mathrm{data} \otimes C_\mathrm{data})  ( I_{d^2} + P_{trans})
\end{align*}
where $P_{trans}$ is the commutation matrix that realizes the transpose operation in vectorized form, and $I_{d^2}$ is the identity matrix on the vectorized space. Using again that higher-order terms have zero-mean, we can write $\hat C_N$ as 
\begin{align*} 
\hat C_N =  C_\mathrm{data} + \frac{1}{\sqrt{N}}   C^{(1)} + \frac{1}{N}   C^{(2)} + o\left( \frac{1}{N}\right) \quad \text{where} \quad   C^{(1)} \sim \mathcal{N} (0, \mathcal{V}) \quad \text{and} \quad \EE[  C^{(2)}] = 0
\end{align*}
In particular $C^{(1)}$ is symmetric.
\end{proof}

Considering the two sources of randomness—the optimization error from SGD (denoted by the subscript \(SGD\)) and the randomness due to data sampling (denoted by the subscript \(gen\))—we obtain the following expressions for the expectations:
\begin{align*} 
\mathbb{E}[A] &= \mathbb{E}_{gen} \mathbb{E}_{SGD} [A_N], \\
\mathbb{E}[b] &= \mathbb{E}_{gen} \mathbb{E}_{SGD} [b_N] = \mathbb{E}_{gen} \mathbb{E}_{SGD} [A_N \hat \mu_N] .
\end{align*}
Furthermore, applying the law of total covariance, we derive:
\begin{align*}
\operatorname{Cov}(A) &= \mathbb{E}_{gen} [\operatorname{Cov}_{SGD} (A_N)] + \operatorname{Cov}_{gen} [\mathbb{E}_{SGD} (A_N)], \\
\operatorname{Cov}(b) &= \mathbb{E}_{gen} [\operatorname{Cov}_{SGD} (b_N)] + \operatorname{Cov}_{gen} [\mathbb{E}_{SGD} (b_N)] .
\end{align*}
Note that for the covariance of $b$,
\begin{align*}
    \cov(b) &= \EE_{gen}[\cov_{SGD}[b_N]] + \cov_{gen}[\EE_{SGD}[b_N]] \\
    &= \EE_{gen}[\cov_{SGD}[\kappa_N]] + \EE_{gen}[\cov_{SGD}[A_N \hat \mu_N]] + \cov_{gen}[\EE_{SGD}[A_N] \hat \mu_N].
\end{align*}
Combining the expressions of the SGD expectations and covariances of $\kappa_N$ and $A_N$ from Equations~\eqref{eq:exp_bN} to \eqref{eq:exp_AN_h} and the second-order expansion of these quantities given in Lemma~\ref{lem:SGD_4}, we can calculate the second-order expansion of the above quantities, to get the final result of Theorem~\ref{thm:SGD_error}. 
The calculations are now detailed.

\paragraph{Mean of $b$}
\begin{align}
    \EE[b] &=  \EE_{gen}\EE_{SGD}[b_N] \\
    &= \EE_{gen}[(\hat C_\sigma^N)^{-1} \hat \mu_N] \\
    &= \EE_{gen} \left(C_\sigma^{-1} - \frac{1}{\sqrt{N}} C_\sigma^{-1}   C^{(1)} C_\sigma^{-1} +  \frac1N X^{(2)} + o\left( \frac{1}{N}\right) \right) \cdot  \label{eq:exp_bN2} \\ &\qquad  \left(\mu + \frac{1}{\sqrt{N}} \mu^{(1)} + \frac{1}{N}\mu^{(2)} + o\left(\frac{1}{N}\right)\right) \nonumber
\end{align}
where $X^{(2)}$ is a term linear in $C^{(2)}$, and thus that satisfies $\EE_{gen}[X^{(2)}] = 0$. Canceling the zero-mean terms we get 
\begin{align*}
\EE[b] &=  C_\sigma^{-1}\mu - \frac{1}{N} \EE[C_\sigma^{-1} C^{(1)} C_\sigma^{-1} \mu^{(1)}] + o\left(\frac{1}{N}\right) \\
&= C_\sigma^{-1}\mu + o\left(\frac{1}{N}\right)
\end{align*}
by uncorrelation between $\mu^{(1)}$ and $C^{(1)}$. 

\paragraph{Covariance of $b$}
\begin{align} \label{eq:cov_b_first}
    \cov(b) &= \mathbb{E}_{gen} [\operatorname{Cov}_{SGD} (\kappa_N)] + \mathbb{E}_{gen} [\operatorname{Cov}_{SGD} (A_N \hat \mu_N)] + \operatorname{Cov}_{gen} [\mathbb{E}_{SGD} (A_N \hat \mu_N)] 
\end{align}
Using that $A_N \hat \mu_N = (\hat \mu_N^\top \otimes \id) vec(A_N)$, 
\begin{align*}
\operatorname{Cov}_{SGD} (A_N \hat \mu_N) &= \EE_{SGD}\left[ A_N \hat \mu_N \hat \mu_N^\top A_N \right] \\
&= \EE_{SGD}\left[ (\hat \mu_N^\top \otimes Id) \cov_{SGD}(A_N)   (\hat \mu_N \otimes Id) \right] \\
&= \frac{1}{\sigma^2} \frac{\tau}{2} \left(\mu_N^\top  (\hat C^N_\sigma)^{-1} \hat C_N  \mu_N \right)  \id .
\end{align*}
We get for~\eqref{eq:cov_b_first}
\begin{align}
    \cov(b) &= \mathbb{E}_{gen} [\operatorname{Cov}_{SGD} (\kappa_N)] + \mathbb{E}_{gen} [\operatorname{Cov}_{SGD} (A_N \hat \mu_N)] + \operatorname{Cov}_{gen} [\mathbb{E}_{SGD} (A_N \hat \mu_N)] \\
    &= \label{eq:124}  \frac{1}{\sigma^2} \frac{\tau}{2}\EE_{gen} \left[\left( (\hat C^N_\sigma)^{-1} \hat C_N \right) \right] + \frac{1}{\sigma^2} \frac{\tau}{2}\mathbb{E}_{gen} \left[ \left(\mu_N^\top  (\hat C^N_\sigma)^{-1} \hat C_N  \mu_N \right) \id\right]  \\&+ \cov_{gen}[(\hat C_\sigma^N)^{-1} \hat \mu_N] + o(\tau) \nonumber .
\end{align}
For each term of~\eqref{eq:124}, we derive its expansion at the order $2$ in $\frac{1}{\sqrt{N}}$. First, for the first term,
\begin{align*}
\left( (\hat C^N_\sigma)^{-1} \hat C_N \right) &=  \left(C_\sigma + \frac{1}{\sqrt{N}}   C^{(1)} + \frac{1}{N}  C^{(2)} + o\left( \frac{1}{N}\right) \right)^{-1} \cdot  \\
&\qquad \nonumber \left(C_\mathrm{data} + \frac{1}{\sqrt{N}}   C^{(1)} + \frac{1}{N}  C^{(2)} + o\left( \frac{1}{N}\right)  \right) \\
&= \left(C_\sigma^{-1} - \frac{1}{\sqrt{N}} C_\sigma^{-1}   C^{(1)} C_\sigma^{-1} +  \frac1N X^{(2)} + o\left( \frac{1}{{N}}\right) \right) \cdot  \\
&\qquad \nonumber \left(C_\mathrm{data} + \frac{1}{\sqrt{N}}   C^{(1)} + \frac{1}{N}  C^{(2)} + o\left( \frac{1}{N}\right)  \right) 
\end{align*}
where $X^{(2)}$ is a term linear in $C^{(2)}$, and thus that satisfies $\EE_{gen}[X^{(2)}] = 0$.
Taking the expectation of the terms up to the order $o\left( \frac{1}{N}\right)$, we get
\begin{align*}
\EE_{gen}\left[ (\hat C^N_\sigma)^{-1} \hat C_N \right] &= C_\sigma^{-1} C_\mathrm{data} + \frac{1}{\sqrt{N}} \EE_{gen}\left[\left(C_\sigma^{-1} C^{(1)} - C_\sigma^{-1} C^{(1)} C_\sigma^{-1} C_\mathrm{data}  \right)\right] \\&- \frac1N \EE_{gen}\left[(C_\sigma^{-1}   C^{(1)})^2 + A^{(2)}C_\mathrm{data} + C^{-1}_\sigma C^{(2)}\right] +   o\left( \frac1N\right) \\
&= C_\sigma^{-1} C_\mathrm{data} - \frac1N C_\sigma^{-1} \EE_{gen}\left[  C^{(1)} C_\sigma^{-1} C^{(1)}  \right] +   o\left( \frac1N\right).
\end{align*}
From the property $vec(ABC) = (C^\top \otimes A)vec(B)$, 
\begin{align*}
\EE_{gen}\left[vec(C^{(1)} C_\sigma^{-1} C^{(1)})\right] &= \EE_{gen}\left[C^{(1)\top} \otimes C^{(1)} \right] vec(C_\sigma^{-1}) \\
&= \mathcal{V} \, vec(C_\sigma^{-1}) \\
&= (C_\mathrm{data} \otimes C_\mathrm{data}) (vec(C_\sigma^{-1}) + vec(C_\sigma^{-1})) \\
&= 2 vec (C_\mathrm{data} C_\sigma^{-1} C_\mathrm{data}),
\end{align*}
that is to say
\begin{align} \label{eq:ExpCmC}
\EE_{gen} \left[ (\hat C^N_\sigma)^{-1} \hat C_N \right] &= C_\sigma^{-1} C_\mathrm{data} - \frac2N (C_\sigma^{-1} C_\mathrm{data})^2 + o\left( \frac1N\right) .
\end{align}
Then, for the second term of~\eqref{eq:124}, we have by uncorrelation between $\hat \mu_N$ and $\hat C_N$, 
\begin{align*}
\mathbb{E}_{gen} \left[ \mu_N^\top  (\hat C^N_\sigma)^{-1} \hat C_N  \mu_N \right] &=  \left\langle \mathbb{E}_{gen} \left[ \mu_N\mu_N^\top \right],  \mathbb{E}_{gen} \left[ (\hat C^N_\sigma)^{-1} \hat C_N \right]  \right\rangle .
\end{align*}
But,
\begin{align*}
\mathbb{E}_{gen} [ \hat \mu_N \hat \mu_N^\top )] &= \mu \mu^\top + \frac{1}{N} \EE_{gen}[\mu^{(1)}(\mu^{(1)})^\top] + o \left( \frac{1}{N}\right) = \mu \mu^\top +  \frac{1}{N} C_\mathrm{data} + o \left( \frac{1}{N}\right) ,
\end{align*}
and thus 
\begin{align*}
\mathbb{E}_{gen} \left[ \mu_N^\top  (\hat C^N_\sigma)^{-1} \hat C_N  \mu_N \right] &=   \left\langle  \mu\mu^\top, C_\sigma^{-1} C_\mathrm{data} \right\rangle - \frac2N \left\langle  \mu\mu^\top, (C_\sigma^{-1} C_\mathrm{data})^2 \right\rangle \\
&+ \nonumber \frac1N \left\langle C_\mathrm{data}, C_\sigma^{-1} C_\mathrm{data} \right\rangle + o \left( \frac{1}{N}\right) .
\end{align*}
Finally, for the third term of~\eqref{eq:124}, from~\eqref{eq:exp_bN2} we have 
\begin{align}
\cov_{gen}[(\hat C_N)^{-1} \hat \mu_N] &= \cov_{gen}\left[C_\sigma^{-1} \mu + \frac{1}{\sqrt{N}} (C_\sigma^{-1} \mu^{(1)} - C_\sigma^{-1} C^{(1)} C_\sigma^{-1} \mu) + o\left(\frac{1}{\sqrt{N}}\right)   \right] \\
&= \frac1N \left( \cov_{gen}[C_\sigma^{-1} \mu^{(1)}] + \cov_{gen}[C_\sigma^{-1} C^{(1)} C_\sigma^{-1} \mu] \right) \\
&=  \frac1N C_\sigma^{-1} \cov_{gen}(\mu^{(1)}) C_\sigma^{-1}  \\
&+ \nonumber \frac1N   \EE_{gen}\left[ C_\sigma^{-1} C^{(1)} C_\sigma^{-1} \mu \mu^\top C_\sigma^{-1} (C^{(1)})^\top C_\sigma^{-1}\right]  \\
&=  \frac1N C_\sigma^{-2} C_\mathrm{data}  + \frac1N \EE_{gen}\left[ C_\sigma^{-1} C^{(1)} C_\sigma^{-1} \mu \mu^\top C_\sigma^{-1} (C^{(1)})^\top C_\sigma^{-1}\right] \label{eq:141}
\end{align}
We simplify the second term:
\begin{align*}
& vec\left( \EE_{gen}\left[ C_\sigma^{-1} C^{(1)} C_\sigma^{-1} \mu \mu^\top C_\sigma^{-1} (C^{(1)})^\top C_\sigma^{-1}\right] \right) \\
&= (\mu^\top \otimes \id) \cov_{gen}(C_\sigma^{-1} C^{(1)} C_\sigma^{-1})  (\mu \otimes \id)  \\
&=  (\mu^\top \otimes \id)  (C_\sigma^{-1} \otimes C_\sigma^{-1})  \cov_{gen}(C^{(1)}) (C_\sigma^{-1} \otimes C_\sigma^{-1}) (\mu \otimes \id) \\
&=  (\mu^\top \otimes \id)  (C_\sigma^{-1} \otimes C_\sigma^{-1})  (C_\mathrm{data} \otimes C_\mathrm{data})  ( I + P_{trans})   (C_\sigma^{-1} \otimes C_\sigma^{-1})  (\mu\otimes \id) \\
&=  (\mu^\top \otimes \id)  (C_\sigma^{-1} \otimes C_\sigma^{-1})  (C_\mathrm{data} \otimes C_\mathrm{data})  (C_\sigma^{-1} \otimes C_\sigma^{-1}) ( I + P_{trans})  (\mu\otimes \id) \\
&= (\mu^\top \otimes \id)  (C_\sigma^{-2} C_\mathrm{data} \otimes C_\sigma^{-2} C_\mathrm{data})  ( I + P_{trans})  (\mu\otimes \id). 
\end{align*}
Plugging back into~\eqref{eq:141}:
\begin{align*}
\cov_{gen}[(\hat C_N)^{-1} \hat \mu_N] &= \frac1N  C_\sigma^{-2} C_\mathrm{data}  + \frac1N (\mu^\top \otimes \id)  (C_\sigma^{-2} C_\mathrm{data} \otimes C_\sigma^{-2} C_\mathrm{data})  ( I + P_{trans})  (\mu\otimes \id).
\end{align*}
The covariance of $b$~\eqref{eq:124} eventually simplifies to
\begin{align} \label{eq:cov_b_app}
\cov(b) &= \frac{\tau}{2 \sigma^2}C_\sigma^{-1} C_\mathrm{data} - \frac{\tau}{N \sigma^2} (C_\sigma^{-1} C_\mathrm{data})^2 + \frac{1}{N}  C_\sigma^{-2}C_\mathrm{data} \\ \nonumber
&+ \frac{\tau}{2 \sigma^2} \langle \mu\mu^\top, C_\sigma^{-1} C_\mathrm{data} \rangle \id - \frac{\tau}{ N\sigma^2} \langle\mu\mu^\top,  (C_\sigma^{-1}C_\mathrm{data})^2 \rangle \id + \frac{\tau}{ 2N\sigma^2} \langle C_\mathrm{data}^2, C_\sigma^{-1} \rangle \id \\ \nonumber
&+ \frac1N (\mu^\top \otimes \id)  (C_\sigma^{-2} C_\mathrm{data} \otimes C_\sigma^{-2} C_\mathrm{data})  ( I + P_{trans})  (\mu\otimes \id).
\end{align}
This covariance will be later further simplified using the expression of the covariance of $A$, that we now establish.
\paragraph{Mean of $A$} 
\begin{align*}
    \EE[A] &=  \EE_{gen}\EE_{SGD}[A_N] = \EE_{data}[(\hat C_N + \sigma^2\id)^{-1}] = C_\sigma^{-1} + o \left( \frac{1}{\sqrt{N}} \right)
\end{align*}
\paragraph{Covariance of $A$} 
\begin{align*}
    \cov(A) &= \EE_{gen}[Cov_{SGD}[A_N]] + \cov_{gen}[\EE_{SGD}[A_N]] \\
    &=  \underbrace{\frac{\tau}{2 \sigma^2}  \EE_{gen}\left[ \left( (\hat C^N_\sigma)^{-1} \hat C_N \right) \otimes \id \right]}_{\text{term 1}} + \underbrace{\cov_{gen}\left[(\hat C_N + \sigma^2\id)^{-1} \right]}_{\text{term 2}}
\end{align*}
Our goal is to compute the expansion of these terms at order 1 in $\frac{1}{N}$. 
Using the expansion of $\EE\left[ (\hat C^N_\sigma)^{-1} \hat C_N \right]$ from~\eqref{eq:ExpCmC}, we have
\begin{align*}
\text{term 1} 
&= \frac{\tau}{2 \sigma^2} \left( C_\sigma^{-1} C_\mathrm{data} \right) \otimes  \id - \frac{\tau}{N \sigma^2} \left( C_\sigma^{-1} C_\mathrm{data}\right)^2 \otimes \id + o \left(\tau, \frac{1}{N}\right) .
\end{align*}
For the term 2, from the expression~\eqref{eq:CN}, we have 
\begin{align*}
    (\hat C_N + \sigma^2\id)^{-1} = C_\sigma^{-1} - \frac{1}{\sqrt{N}}  C_\sigma^{-1} C^{(1)} C_\sigma^{-1}  + o( \frac{1}{\sqrt{N}}) 
\end{align*}
and thus 
\begin{align*}
\cov_{gen}\left[(\hat C_N + \sigma^2\id)^{-1} \right]  &= \frac{1}{N} \cov_{gen}(C_\sigma^{-1} C^{(1)} C_\sigma^{-1}).
\end{align*}
Now, using that 
\begin{align*}
vec(C_\sigma^{-1} C^{(1)}  C_\sigma^{-1}) = (C_\sigma^{-1} \otimes C_\sigma^{-1})vec(C^{(1)}), 
\end{align*}
we have 
\begin{align*}
\cov_{gen}(C_\sigma^{-1} C^{(1)} C_\sigma^{-1}) &= \EE_{gen}[vec(C_\sigma^{-1} C^{(1)} C_\sigma^{-1})vec(C_\sigma^{-1} C^{(1)} C_\sigma^{-1})^\top] \\
&=(C_\sigma^{-1} \otimes C_\sigma^{-1})  \EE_{gen}[ vec(C^{(1)})vec(C^{(1)})^\top]   (C_\sigma^{-1} \otimes C_\sigma^{-1})^\top   \\
&= (C_\sigma^{-1} \otimes C_\sigma^{-1})  \mathcal{V}   (C_\sigma^{-1} \otimes C_\sigma^{-1}) \\
&= (C_\sigma^{-1} \otimes C_\sigma^{-1})  (C_\mathrm{data} \otimes C_\mathrm{data}) (I + P_{trans})   (C_\sigma^{-1} \otimes C_\sigma^{-1}) \\
&= \left(\left( C_\sigma^{-1} C_\mathrm{data} \right) \otimes \left( C_\sigma^{-1} C_\mathrm{data} \right)\right) (I + P_{trans})  (C_\sigma^{-1} \otimes C_\sigma^{-1}) .
\end{align*}
Note that for $Z \in \RR^{d\times d}$, $(C_\sigma^{-1} \otimes C_\sigma^{-1})vec(Z) = vec(C_\sigma^{-1} Z C_\sigma^{-1})$, and thus  
\begin{align*}
    \left(\left( C_\sigma^{-1} \right.\right. & C_\mathrm{data} \left. \left. \right) \otimes \left( C_\sigma^{-1} C_\mathrm{data} \right)\right) (I + P_{trans})(C_\sigma^{-1} \otimes C_\sigma^{-1})vec(Z) \\
    &=  (\left( C_\sigma^{-1} C_\mathrm{data} \right) \otimes \left( C_\sigma^{-1} C_\mathrm{data} \right))\left(vec(C_\sigma^{-1} Z C_\sigma^{-1}) + vec(C_\sigma^{-1} Z^\top C_\sigma^{-1})\right) \\
    &= \left( C_\sigma^{-1} C_\mathrm{data} \right) C_\sigma^{-1} Z C_\sigma^{-1}\left( C_\sigma^{-1} C_\mathrm{data} \right) + \left( C_\sigma^{-1} C_\mathrm{data} \right) C_\sigma^{-1} Z^\top C_\sigma^{-1}\left( C_\sigma^{-1} C_\mathrm{data} \right) \\
    &= ((C_\sigma^{-2} C_\mathrm{data}) \otimes (C_\sigma^{-2}C_\mathrm{data})) (I + P_{trans}) vec(Z) .
\end{align*}
We have eventually 
\begin{equation*}
\begin{split}
    \cov(A) &= \frac{\tau}{2 \sigma^2} \left( C_\sigma^{-1} C_\mathrm{data} \right) \otimes  \id - \frac{\tau}{N \sigma^2} \left( C_\sigma^{-1} C_\mathrm{data} \right)^2 \otimes \id \\
&+ \frac{1}{N} ((C_\sigma^{-2} C_\mathrm{data}) \otimes (C_\sigma^{-2}C_\mathrm{data})) (I + P_{trans})
+ o \left(\tau, \frac{1}{N}\right)  .
\end{split}
\end{equation*}
Note that comparing $\cov(b)$ from \eqref{eq:cov_b_app} with $\cov(A)$, the covariance of $b$ simplifies to
\begin{align*}
\cov(b) &= \cov(A \mu) + \frac{\tau}{2 \sigma^2}C_\sigma^{-1} C_\mathrm{data} + \frac{\tau}{N \sigma^2} \left( \frac12C_\sigma^{-1} C_\mathrm{data}^2 -  (C_\sigma^{-1} C_\mathrm{data})^2 \right) \\ &+ \nonumber  \frac{1}{N} C_\sigma^{-2}C_\mathrm{data} 
\end{align*}
i.e. $b = \hat \kappa + A\mu$ with $\hat \kappa$ uncorrelated to $A$ and 
\begin{align*}
\EE[\hat \kappa] &= 0 \\
\cov(\hat \kappa) &= \frac{\tau}{2 \sigma^2}C_\sigma^{-1} C_\mathrm{data} + \frac{\tau}{N \sigma^2} \left( \frac12 \langle C_\sigma^{-1},  C_\mathrm{data}^2 \rangle \id -  (C_\sigma^{-1} C_\mathrm{data})^2 \right) +  \frac{1}{N} C_\sigma^{-2}C_\mathrm{data} .
\end{align*}
\end{proof}
\end{proof}

\subsection{Experimental details}
\label{app:exp}
All the numerical experiments of the manuscript for SGD (Section~\ref{sec:DSM_error}), Langevin dynamics (Section~\ref{sec:langevin}), and diffusion-based sampling (Section~\ref{sec:diffusion}) were conducted on a single NVIDIA V100 GPU. These experiments utilized synthetic Gaussian data across various dimensions, ranging from 2 to 100. As a direction for future work, we intend to compare our theoretical findings with empirical results obtained from real image datasets and non-linear neural networks.

\section{Langevin sampling error}
\subsection{Proof of Lemma~\ref{lem:lemma_langevin_1}} \label{app:lemma_langevin_1}

Recall that the Lyapunov operator $L^\gamma_C$ is defined, for $C \in \RR^{p\times p}$ and $\gamma >0$ by 
\begin{align} \label{eq:inv_lyap}
    L^\gamma_C[X] = CX + XC^\top - \gamma C X C^\top .
\end{align}
\newtheorem*{repeatlem1}{Lemma 1}
\begin{repeatlem1}[ULA solution with linear score)] 
    Assuming $\operatorname{Re}(\operatorname{eig}(A)) > 0$, and ${\gamma < \frac{2}{\lambda_{max}(A)}}$, the ULA algorithm~\eqref{eq:ULA} with linear score $v(x, \theta) = -A x + b$ converges to 
     \begin{equation*}
    q = \mathcal{N}(A^{-1}b, \Sigma^\gamma_A) \quad \text{with} \quad  \Sigma^\gamma_A \eqdef (L^\gamma_A)^{-1}[2\id].
    \end{equation*}
\end{repeatlem1}
\begin{proof}
The Langevin iterates with linear score $v(x, \theta) = -A x + b$ write
\begin{equation} \label{eq:ULA_linear}
    y_0 \in \mathbb{R}^d, \quad y_{k+1} = y_k + \gamma (-A y_k + b) + \sqrt{2 \gamma}\, w_k .
\end{equation}
By linear combination of independent Gaussian random variables, for \( k \geq 1 \), each iterate \( y_k \) follows a Gaussian distribution. Taking expectations on both sides of~\eqref{eq:ULA_linear}, we obtain the recurrence relation for the mean $m_k \eqdef  \mathbb{E}[y_k]$: 
\begin{equation*}
m_{k+1} = (\mathrm{Id} - \gamma A) m_k + \gamma b.
\end{equation*}
This recurrence converges exponentially fast to
\begin{equation*}
y^* = A^{-1} b.
\end{equation*}
Now, taking the covariance on both sides, we get:
\begin{equation*}
\Sigma_{k+1} \eqdef \EE[(y_{k+1} - y^*)(y_{k+1} - y^*)^\top]  = (\mathrm{Id} - \gamma A) \Sigma_k (\mathrm{Id} - \gamma A)^\top + 2\gamma \mathrm{Id}.
\end{equation*}
At stationarity, the covariance $\Sigma_{k}$ converges to $\Sigma^\gamma_A$ such that:
\begin{equation*}
\Sigma^\gamma_A = (\mathrm{Id} - \gamma A) \Sigma^\gamma_A (\mathrm{Id} - \gamma A)^\top + 2\gamma \mathrm{Id}.
\end{equation*}
Rearranging, we obtain the discrete Lyapunov equation
\begin{equation*}
\Sigma^\gamma_A A^\top + A \Sigma^\gamma_A - \gamma A \Sigma^\gamma_A A^\top  = 2 \mathrm{Id}.
\end{equation*}
Since \( A \) is positive definite and \( \gamma < \frac{2}{\lambda_{\max}(A)} \), this equation has a unique solution, given by:
\begin{equation*}
\Sigma^\gamma_A = (L^\gamma_A)^{-1}[2\mathrm{Id}].
\end{equation*}

\end{proof}

\subsection{Proof of Theorem~\ref{thm:langevin_distance_general}} \label{app:langevin_distance_general}
Recall the form of the score~\eqref{eq:gen_inexact_score}
\begin{equation*}
    v_\sigma(x, \theta) = - A_\varepsilon x + b_\varepsilon, \quad \text{with}\quad 
    A_\varepsilon = A^* + \varepsilon \Delta, \quad b_\varepsilon = b^* + \varepsilon (\delta + \Delta \mu_\mathrm{data}),
\end{equation*}
where \(\Delta\) and \(\delta\) are uncorrelated and zero-mean random variables taking values in $\mathbb{R}^{d \times d}$ and $\mathbb{R}^{d}$.
We denote $\lambda^\sigma_i \eqdef \lambda_i + \sigma^2$ the eigenvalues of $C_\sigma = C_{\mathrm{data}} + \sigma^2 \id$. 
\newtheorem*{repeatthm2}{Theorem 2}
\begin{repeatthm2}
For a small enough stepsize $\gamma$, the ULA algorithm~\eqref{eq:ULA} with inexact linear score~\eqref{eq:gen_inexact_score} converges to an invariant Gaussian distribution $q^\varepsilon$ that satisfies 
\begin{equation*} 
\EE_{\delta, \Delta}  \left[ W_2^2(p_\mathrm{data},q_\varepsilon) \right] =  \mathcal{E}^{(0)}_{\sigma, \gamma} + \varepsilon^2 \left( \mathcal{E}_{\sigma, \gamma}^{\text{mean}}(\delta) + \mathcal{E}_{\sigma, \gamma}^{\text{cov}}(\Delta) \right) + o(\varepsilon^2)
\end{equation*}
\begin{equation*} 
\begin{split}
\text{where} &\quad 
\mathcal{E}^{(0)}_{\sigma, \gamma} = \sum_i \left(\sqrt{\lambda_i} - \frac{\lambda_i^\sigma}{\sqrt{\lambda_i^\sigma - \frac{\gamma}{2}}}\right)^2, \qquad
\mathcal{E}_{\sigma, \gamma}^{\text{mean}}(\delta) = \sum_{i=1}^d \sqrt{\lambda^\sigma_i } \, \langle \cov(\delta), u_i u_i^\top \rangle,  \\ \vspace{-0.2cm}
\mathcal{E}_{\sigma, \gamma}^{\text{cov}}(\Delta) &= \sum_{i,j=1}^d \kappa^{(1)}_{\sigma, \gamma}(\lambda_i, \lambda_j) \, \EE_{\Delta}\left[\langle \Delta, u_iu_j^\top \rangle^2\right] + \kappa^{(2)}_{\sigma, \gamma}(\lambda_i, \lambda_j) \, \EE_{\Delta}\left[\langle \Delta, u_iu_j^\top \rangle \langle \Delta, u_ju_i^\top \rangle \right] 
\end{split}
\end{equation*}
The functions \( \kappa^{(1)}_{\sigma, \gamma}, \kappa^{(2)}_{\sigma, \gamma} : \mathbb{R}^2 \to \mathbb{R} \) are kernel functions that capture interactions between the eigenvalues \( \lambda_i \) of \( C_\mathrm{data} \). Their full expressions are given equations~\eqref{eq:kappa1} and~\eqref{eq:kappa2} and their expressions when $\gamma = 0$ and $\sigma \to 0$ are given equation~\eqref{eq:kappa_simp}.
\end{repeatthm2}

\begin{proof} 

Using the score~\eqref{eq:gen_inexact_score}, applying Lemma~\ref{lem:lemma_langevin_1}, we get that for small enough stepsize, the ULA iterates converge to the invariant Gaussian distribution 
\begin{align*}
q_\varepsilon = \mathcal{N}(\mu_\varepsilon, \Sigma^\gamma_\varepsilon) \quad \text{with} \quad \mu_\varepsilon = A_\varepsilon^{-1} b_\varepsilon \quad \text{and} \quad \Sigma^\gamma_\varepsilon = (L^\gamma_{A_\varepsilon})^{-1}[2\id].
\end{align*}
First, we perform, with the following Lemma, a first-order expansion in $\varepsilon$ of the mean and a second-order expansion of the covariance. 
\begin{lem} \label{lem:Langevin_1}
With the previous notations, the mean $\mu_\varepsilon$ and the covariance $\Sigma_\varepsilon$ expand as 
\begin{align*}
\mu_\varepsilon &=  \mu + \varepsilon C_\sigma \delta   + o(\varepsilon)  \\
\text{and} \quad\Sigma^\gamma_\varepsilon &= \Sigma_0^\gamma  + \varepsilon \Sigma^\gamma_1(\Delta) + \varepsilon^2 \Sigma^\gamma_2(\Delta) + o(\varepsilon^2),
\end{align*}
with 
 \begin{align} \label{eq:Sigma_0}
\Sigma_0^\gamma &= \Sigma^\gamma_{C_\sigma^{-1}} = C_\sigma\left( \id - \frac{\gamma}{2} C_\sigma^{-1} \right)^{-1} \\
\Sigma_1^\gamma(\Delta) &= -\Bigl(L^\gamma_{C_\sigma^{-1}}\Bigr)^{-1}\Bigl[
\Delta \Sigma^\gamma_{C_\sigma^{-1}} (\id - \gamma {C_\sigma^{-1}}) + (\id - \gamma {C_\sigma^{-1}})\Sigma^\gamma_{{C_\sigma^{-1}}} \Delta^\top
\Bigr] \label{eq:Sigma_1}
\\
\Sigma_2^\gamma(\Delta) &= -\Bigl(L^\gamma_{C_\sigma^{-1}}\Bigr)^{-1}\Bigl[
\Delta \Sigma^\gamma_{{1}}(\Delta) (\id - \gamma {C_\sigma^{-1}}) + (\id - \gamma {C_\sigma^{-1}})\Sigma^\gamma_{{1}}(\Delta) \Delta^\top- \gamma   \Delta  \Sigma^\gamma_{{C_\sigma^{-1}}}  \Delta^\top
\Bigr]. \label{eq:Sigma_2}
\end{align}
\end{lem}
\begin{proof}
We start with the mean. 
 \begin{align*}
 \mu_\varepsilon &\eqdef (A_\varepsilon)^{-1}b_\varepsilon = \left( A^* + \varepsilon \Delta + o(\varepsilon) \right)^{-1} (b^* + \varepsilon\Delta\mu + \varepsilon \delta + o(\varepsilon)) \\
 &= (A^*)^{-1}b^* - \varepsilon (A^*)^{-1} \Delta (A^*)^{-1} b^* + \varepsilon (A^*)^{-1} \delta  + \varepsilon (A^*)^{-1} \Delta \mu + o(\varepsilon^2)
 \end{align*}
Using $A^* = C_\sigma^{-1}$ and $b^* = C_\sigma^{-1} \mu$, we get 
  \begin{align*}
 \mu_\varepsilon &= \mu + \varepsilon C_\sigma  \delta  + o(\varepsilon)
 \end{align*}
We now focus on the covariance.
 \begin{align*}
\Sigma^\gamma_{A} = \Bigl(L^\gamma_{A^*+\varepsilon\Delta}\Bigr)^{-1}[2I],
\end{align*}
Substitute $A = A^* + \varepsilon\Delta$ into the definition~\eqref{eq:inv_lyap} of $L^\gamma_A[Z]$:
 \begin{align*}
L^\gamma_{A^*+\varepsilon\Delta}[Z] = (A^*+\varepsilon\Delta)Z + Z(A^*+\varepsilon\Delta)^\top - \gamma  (A^*+\varepsilon\Delta)Z (A^*+\varepsilon\Delta)^\top.
 \end{align*}
We now expand each term:
 \begin{align*}
\begin{aligned}
(A^*+\varepsilon\Delta)Z &= A^* Z + \varepsilon   \Delta Z,\\
Z(A^*+\varepsilon\Delta)^\top &= Z (A^*)^\top + \varepsilon   Z \Delta^\top,\\
(A^*+\varepsilon\Delta) Z (A^*+\varepsilon\Delta)^\top 
&= A^*Z (A^*)^\top + \varepsilon\Bigl[\Delta Z (A^*)^\top + A^*Z \Delta^\top\Bigr] + \varepsilon^2   \Delta Z \Delta^\top.
\end{aligned}
 \end{align*}
Substituting these into the expression for $L^\gamma_{A^*+\varepsilon\Delta}[Z]$ yields
 \begin{align*}
L^\gamma_{A^*+\varepsilon\Delta}[Z] &= A^*Z + Z (A^*)^\top - \gamma   A^*Z (A^*)^\top\\ \nonumber
&\quad {}+ \varepsilon\Bigl[\Delta Z + Z \Delta^\top - \gamma\Bigl(\Delta Z (A^*)^\top + A^*Z \Delta^\top\Bigr)\Bigr] - \gamma  \varepsilon^2   \Delta Z \Delta^\top.
 \end{align*}
For clarity, we define the following operators:
 \begin{align*}
\mathcal{L}_0[Z] &\eqdef A^*Z + Z (A^*)^\top - \gamma   A^*Z (A^*)^\top,\\
\mathcal{L}_1[Z] &\eqdef \Delta Z + Z \Delta^\top - \gamma\Bigl(\Delta Z (A^*)^\top + A^*Z \Delta^\top\Bigr),\\
\mathcal{L}_2[Z] &\eqdef - \gamma   \Delta Z \Delta^\top.
 \end{align*}
such that
 \begin{align*}
L^\gamma_{A^*+\varepsilon\Delta}[Z] = \mathcal{L}_0[Z] + \varepsilon  \mathcal{L}_1[Z] + \varepsilon^2  \mathcal{L}_2[Z].
 \end{align*}
We now seek an expansion of the solution:
 \begin{align*}
\Sigma^\gamma_{A^*+\varepsilon\Delta} = \Sigma_0^\gamma + \varepsilon  \Sigma_1^\gamma + \varepsilon^2  \Sigma_2^\gamma + o(\varepsilon^2).
 \end{align*}
By definition,
 \begin{align*}
L^\gamma_{A^*+\varepsilon\Delta}\Bigl[\Sigma^\gamma_{A^*+\varepsilon\Delta}\Bigr] = 2I.
 \end{align*}
Substituting the expansions of both the operator and the solution, we obtain:
 \begin{align*}
\Bigl(\mathcal{L}_0 + \varepsilon  \mathcal{L}_1 + \varepsilon^2  \mathcal{L}_2\Bigr)
\Bigl[\Sigma_0^\gamma + \varepsilon  \Sigma_1^\gamma + \varepsilon^2  \Sigma_2^\gamma\Bigr] = 2I.
 \end{align*}
We now collect terms by powers of $\varepsilon$.

\paragraph{Order \(\varepsilon^0\)}
 \begin{align*}
\mathcal{L}_0\Bigl[\Sigma_0^\gamma\Bigr] = 2I.
 \end{align*}

Thus, the zeroth-order term is given by
 \begin{align*}
\Sigma_0^\gamma = \Bigl(L^\gamma_{A^*}\Bigr)^{-1}[2I].
 \end{align*}

\paragraph{Order \(\varepsilon^1\)}
Collecting the terms of order $\varepsilon$, we have:
 \begin{align*}
\mathcal{L}_0\Bigl[\Sigma_1^\gamma\Bigr] + \mathcal{L}_1\Bigl[\Sigma_0^\gamma\Bigr] = 0.
 \end{align*}
Therefore,
 \begin{align*}
\Sigma_1^\gamma = -\Bigl(L^\gamma_{A^*}\Bigr)^{-1}\Bigl[\mathcal{L}_1\Bigl[\Sigma_0^\gamma\Bigr]\Bigr],
 \end{align*}
or explicitly,
 \begin{align*}
\Sigma_1^\gamma = -\Bigl(L^\gamma_{A^*}\Bigr)^{-1}\Bigl[
  \Delta  \Sigma_0^\gamma + \Sigma_0^\gamma  \Delta^\top - \gamma\Bigl(\Delta  \Sigma_0^\gamma  (A^*)^\top + A^*  \Sigma_0^\gamma  \Delta^\top\Bigr)
\Bigr].
 \end{align*}

\paragraph{Order \(\varepsilon^2\)}
At order $\varepsilon^2$, we have:
 \begin{align*}
\mathcal{L}_0\Bigl[\Sigma_2^\gamma\Bigr] + \mathcal{L}_1\Bigl[\Sigma_1^\gamma\Bigr] + \mathcal{L}_2\Bigl[\Sigma_0^\gamma\Bigr] = 0.
 \end{align*}
This implies
 \begin{align*}
\Sigma_2^\gamma = -\Bigl(L^\gamma_{A^*}\Bigr)^{-1}\Bigl[
\mathcal{L}_1\Bigl[\Sigma_1^\gamma\Bigr] + \mathcal{L}_2\Bigl[\Sigma_0^\gamma\Bigr]
\Bigr],
 \end{align*}
or in an expanded form,
 \begin{align*}
\Sigma_2^\gamma = -\Bigl(L^\gamma_{A^*}\Bigr)^{-1}\Bigl[
  \Delta  \Sigma_1^\gamma + \Sigma_1^\gamma  \Delta^\top - \gamma\Bigl(\Delta  \Sigma_1^\gamma  (A^*)^\top + A^*  \Sigma_1^\gamma  \Delta^\top\Bigr)
- \gamma   \Delta  \Sigma_0^\gamma  \Delta^\top
\Bigr].
 \end{align*}
\end{proof}
Based on the above expansion, we derive the expansion in $\varepsilon$ of the averaged Wasserstein distance 
\begin{equation*} 
    \EE_{\delta, \Delta} \left[W_2^2(p_\mathrm{data},q_\varepsilon)\right] =   \EE_{\delta} \left[ \EE\norm{\mu - \mu_\varepsilon}^2 \right]+   \EE_{\Delta} \left[ \mathcal{B}^2(C_\mathrm{data}, \Sigma^\gamma_{\varepsilon}) \right]
\end{equation*}
We calculate each term separately.

\paragraph{$L^2$ distance between means} For the first term,
\begin{align*}
\EE_{\delta} \norm{\mu - \mu_{\epsilon}}^2 
&=  \varepsilon^2 \EE_{\delta} \norm{C_\sigma \delta}^2 + o(\varepsilon^2) \\
&= \varepsilon^2  \langle C_\sigma^2, \cov(\delta) \rangle 
+ o(\varepsilon^2) 
\end{align*}
Using that $C_\sigma = U \diag{\lambda^\sigma_i} U^\top$, 
\begin{align*}
\EE_{\delta} \norm{\mu - \mu_{\epsilon}}^2 &= \varepsilon^2 \sum_i \left(\lambda^\sigma_i \right)^2 \langle \cov(\delta), u_i u_i^\top \rangle 
+ o(\varepsilon^2) .
\end{align*}
This term corresponds to $\mathcal{E}_{\sigma, \gamma}^{\text{mean}}(\delta)$ in Theorem~\ref{thm:langevin_distance_general}.

\paragraph{Bures distance between covariances}
The Bures squared distance is defined by
\begin{align} \label{eq:Bures}
\mathcal{B}^2(C_\mathrm{data}, \Sigma^\gamma_\varepsilon) &= \tr{ C_\mathrm{data} + \Sigma^\gamma_\varepsilon - 2 (\sqrt{C_\mathrm{data}} \Sigma^\gamma_\varepsilon \sqrt{C_\mathrm{data}})^{1/2} } .
\end{align}
The authors of \cite{malago2018wasserstein} calculated the first-order decomposition in $\varepsilon$ of $\mathcal{B}^2(X, X + Z)$. Here the second term in the Bures distance~\eqref{eq:Bures} does not write as $C_\mathrm{data} + \varepsilon \Delta$ but as $ \Sigma^\gamma_0 + \varepsilon \Delta$ with $\Sigma^\gamma_0 \neq C_\mathrm{data}$~: we need to take into account the error at order $0$. 
\begin{prop}[Second order Taylor expansion of the Bures distance] \label{prop:Langevin_2}
For $\Sigma, H_0 \in \operatorname{Sym}^{++}(d)$, let $H_1 \in \operatorname{Sym}(d)$ such that ${H_0 \pm H_1 \in  \operatorname{Sym}^{++}(d)}$, then 
\begin{align*}
\mathcal{B}^2(\Sigma, H_0 + \varepsilon H_1) &= \mathcal{B}^2(\Sigma, H_0) +\varepsilon\tr{H_1} - 2\varepsilon \tr{L_{\left({\Sigma}^{1/2}H_0{\Sigma}^{1/2} \right)^{1/2}}^{-1}\left[{\Sigma}^{1/2} H_1 {\Sigma}^{1/2}\right]} \\ &+ 2\varepsilon^2\tr{L_{\left({\Sigma}^{1/2}H_0{\Sigma}^{1/2} \right)^{1/2}}^{-1}\left[\left(L_{\left({\Sigma}^{1/2}H_0{\Sigma}^{1/2} \right)^{1/2}}^{-1}\left[{\Sigma}^{1/2} H_1 {\Sigma}^{1/2}\right]\right)^2\right] } +o(\varepsilon^2) \nonumber
\end{align*} 
\end{prop}

\begin{proof} 
We have
\begin{align}
\mathcal{B}^2(\Sigma, H_0 + \varepsilon H_1) &= \tr{ \Sigma + H_0 + \varepsilon H_1 - 2 \left(\Sigma^{^{1/2}} (H_0 + \varepsilon H_1) \Sigma^{^{1/2}}\right)^{1/2} } \\
&= \tr{\Sigma +  H_0 + \varepsilon H_1 - 2 \left(\Sigma^{1/2}H_0\Sigma^{1/2}  + \varepsilon \Sigma^{1/2} H_1\Sigma^{1/3}\right)^{1/2} }. \label{eq:bures_exp_1}
\end{align}
In order to prove the result, we need to calculate the second-order expansion of the matrix square root.
\begin{lem}[Second order Taylor expansion of the matrix square root] \label{lem:taylor_square_root}
For $H_0 \in \operatorname{Sym}^{++}(d)$, let $H_1 \in \operatorname{Sym}(d)$ such that ${H_0 \pm H_1 \in  \operatorname{Sym}^{++}(d)}$, then 
\begin{align*}
(H_0 + \varepsilon H_1)^{1/2} = H_0^{1/2} + \varepsilon L_{H_0^{1/2}}^{-1}[H_1] - \varepsilon^2   L_{H_0^{1/2}}^{-1}\left[\left(L_{H_0^{1/2}}^{-1}[H_1]\right)^2\right]  + o(\varepsilon^2) .
\end{align*} 
\end{lem}
\begin{proof}
We consider the second-order development
\begin{align*}
(H_0 + \varepsilon H_1)^{1/2} &= H_0^{1/2} + \varepsilon X_1  + \varepsilon^2 X_2 + o(\varepsilon^2).
\end{align*} 
Then by taking the square on both sides,
\begin{align*}
H_0 + \varepsilon H_1 &= H_0 + \varepsilon ( H_0^{1/2} X_1 + X_1 H_0^{1/2})  + \varepsilon^2 (H_0^{1/2} X_2 +  X_2 H_0^{1/2} + X_1^2) +  o(\varepsilon^2)
\end{align*}
we thus get 
\begin{align*}
X_1 &= L_{H_0^{1/2}}^{-1}[H_1] \\
X_2 &= - L_{H_0^{1/2}}^{-1}[X_1^2]
\end{align*}
\end{proof}
Applying this result in~\eqref{eq:bures_exp_1} with $H_0 \to \Sigma^{1/2} H_0 \Sigma^{1/2}$ and $H_1 \to \Sigma^{1/2} H_1 \Sigma^{1/2}$, we get 
\begin{align*}
\mathcal{B}^2(\Sigma, H_0 + \varepsilon H_1) &= \operatorname{Tr} \left(\Sigma + H_0 + \varepsilon H_1 - 2 \left( \Sigma^{1/2} H_0 \Sigma^{1/2} + \varepsilon L_{\Sigma^{1/2} H_0 \Sigma^{1/2}}^{-1}[\Sigma^{1/2} H_1 \Sigma^{1/2}] \right. \right.\\
& \left. \left. - \varepsilon^2   L_{\Sigma^{1/2} H_0 \Sigma^{1/2}}^{-1}\left[\left(L_{\Sigma^{1/2} H_0 \Sigma^{1/2}}^{-1}[\Sigma^{1/2} H_1 \Sigma^{1/2}]\right)^2\right]  \right) \right) +o(\varepsilon^2) \nonumber .
\end{align*}
which simplifies to the desired result.
\end{proof}

The rest of the proof consists in applying the above proposition with the expansion derived in Lemma~\ref{lem:Langevin_1}: ${\Sigma_\varepsilon = \Sigma^\gamma_{C_\sigma^{-1}} + \varepsilon \Sigma^\gamma_1(\Delta) + \varepsilon^2 \Sigma^\gamma_2(\Delta)) + o(\varepsilon^2)}$. Note that from the linearity of the first-order term $\Sigma^\gamma_1(\Delta)$ in $\Delta$ and the fact that $\Delta$ has zero-mean, when taking the mean of the Bures distance in $\Delta$, the first-order term in~$\varepsilon$ disappears. We are then left to calculating 
\begin{align} \label{eq:bures_delta_main}
\EE_{\Delta}\left[\mathcal{B}^2( C_\mathrm{data}, \Sigma_\varepsilon^\gamma)\right] &= 
\underbrace{\mathcal{B}^2(C_\mathrm{data}, \Sigma^\gamma_{C_\sigma^{-1}})}_{\text{term A}} + \varepsilon^2 \underbrace{2\tr{L_{(X_0)^{1/2}}^{-1}\left[\EE\left[\left(L_{(X_0)^{1/2}}^{-1}\left[X_1(\Delta)\right]\right)^2\right]\right] }}_{\text{term B}} \\
&  +\varepsilon^2 \underbrace{\tr{\EE[\Sigma^\gamma_2(\Delta)]-2L_{(X_0)^{1/2}}^{-1}\left[\EE[X_2(\Delta)]\right]}}_{\text{term C}} +o(\varepsilon^2) \nonumber
\end{align}
where 
\begin{empheq}[left=\empheqlbrace]{align}
X_0 &\eqdef {C_\mathrm{data}}^{1/2}\Sigma^\gamma_{C_\sigma^{-1}}{C_\mathrm{data}}^{1/2} \\
X_1(\Delta) &\eqdef {C_\mathrm{data}}^{1/2}\Sigma^\gamma_{1} (\Delta) {C_\mathrm{data}}^{1/2}\\ 
X_2(\Delta) &\eqdef {C_\mathrm{data}}^{1/2}\Sigma^\gamma_{2} (\Delta) {C_\mathrm{data}}^{1/2}
\end{empheq}
\begin{remark} \label{rem:mean_second_order_0}
Note that we obtain a second-order expansion of the distance in $\varepsilon$ while we started from an approximation of the score $A_\varepsilon$ at the first-order in $\varepsilon$. This can be justified by the following fact. If we started from the second-order approximation of the score 
\begin{align*}
A_\varepsilon = C_\sigma^{-1} + \varepsilon \Delta_1 + \varepsilon^2 \Delta_2 + o(\varepsilon^2)
\end{align*}
with $\Delta_2$ a random variable with \textbf{zero-mean}, 
by linearity of $ \Sigma^\gamma_1$, we would have at the second-order in~$\varepsilon$
\begin{align*}
\Sigma_\varepsilon
&= \Sigma^\gamma_{C_\sigma^{-1}} + \varepsilon \Sigma^\gamma_1(\Delta_1) + \varepsilon^2 \left(\Sigma^\gamma_1(\Delta_2) +  \Sigma^\gamma_2(\Delta_1 + \varepsilon  \Delta_2) \right) + o(\varepsilon^2) .
\end{align*}
From the expression~\eqref{eq:Sigma_2} of $\Sigma^\gamma_2$, we have that $\Sigma^\gamma_2(\Delta_1 + \varepsilon  \Delta_2) = \Sigma^\gamma_2(\Delta_1) + O(\varepsilon)$ and thus at the second-order in $\varepsilon$, 
\begin{align*}
\Sigma_\varepsilon
&= \Sigma^\gamma_{C_\sigma^{-1}} + \varepsilon \Sigma^\gamma_1(\Delta_1) + \varepsilon^2 \left(\Sigma^\gamma_1(\Delta_2) +  \Sigma^\gamma_2(\Delta_1) \right) + o(\varepsilon^2) .
\end{align*}
The second-order term then becomes $\Sigma^\gamma_2(\Delta) \to \Sigma^\gamma_1(\Delta_2) +  \Sigma^\gamma_2(\Delta_1)$. However, by linearity of the Bures distance's expansion \eqref{eq:bures_delta_main} in $\Sigma^\gamma_2(\Delta)$, when taking the mean in $\Delta$, the additional term is $0$.
\end{remark}

The three terms $A$, $B$ and $C$ in~\eqref{eq:bures_delta_main} are now calculated in detail by decomposing each term in the eigenbasis of the data.
\paragraph{Term A.} Both covariances $C_\mathrm{data}$ and $\Sigma^\gamma_{C_\sigma^{-1}} = C_\sigma\left( \id - \frac{\gamma}{2} C_\sigma^{-1} \right)^{-1} $ co-diagonalize and thus the Bures distance between them is simply  
\begin{align*}
\mathcal{B}^2(C_\mathrm{data}, \Sigma^\gamma_{C_\sigma^{-1}}) &= \sum_i \left(\sqrt{\lambda_i} - \sqrt{\frac{(\lambda_i^\sigma)^2}{\lambda_i^\sigma - \frac{\gamma}{2}}} \right)^2 = \sum_i \left(\sqrt{\lambda_i} - \frac{\lambda_i^\sigma}{\sqrt{\lambda_i^\sigma - \frac{\gamma}{2}}}\right)^2 .
\end{align*}
It corresponds to the first term $\mathcal{E}^{(0)}_{\sigma, \gamma}$ in Theorem~\ref{thm:langevin_distance_general}.
\paragraph{Term B.} First, let's decompose $X_0$ and $X_1(\Delta)$ in the eigenbasis of $C_\mathrm{data}$.
\begin{align*}
    X_0 &= {C_\mathrm{data}}^{1/2} \Sigma^\gamma_{C_\sigma^{-1}}{C_\mathrm{data}}^{1/2} =  U \operatorname{Diag}\left(\frac{\lambda_i (\lambda^\sigma_i)^2}{\lambda^\sigma_i - \frac{\gamma}{2}} \right) U^\top =  U \operatorname{Diag} \left(\eta^{\sigma, \gamma}_i \right) U^\top,
\end{align*}
where we denote
\begin{align*} 
\eta^{\sigma, \gamma}_i \eqdef \frac{\lambda_i (\lambda^\sigma_i)^2}{\lambda^\sigma_i - \frac{\gamma}{2}}.
\end{align*}
Now for $X_1(\Delta)$. Recall from Lemma~\ref{lem:Langevin_1} that 
\begin{align} \label{eq:sigma_1}
\Sigma_1^\gamma(\Delta) &= -\Bigl(L^\gamma_{C_\sigma^{-1}}\Bigr)^{-1}\Bigl[
\Delta \Sigma^\gamma_{{C_\sigma^{-1}}} (\id - \gamma {C_\sigma^{-1}}) + (\id - \gamma {C_\sigma^{-1}})\Sigma^\gamma_{{C_\sigma^{-1}}} \Delta^\top
\Bigr].
\end{align}
We write $\Sigma_1^\gamma(\Delta)$ in the eigenbasis of the data. First, we have
\begin{align*}
\Sigma^\gamma_{{C_\sigma^{-1}}}(\id - \gamma {C_\sigma^{-1}}) &= ({{C_\sigma^{-1}}})^{-1}\left(\id - \frac{\gamma}{2} {{C_\sigma^{-1}}}\right)^{-1} (\id - \gamma {C_\sigma^{-1}}) \\ &= U \operatorname{Diag} \left( \frac{\lambda^\sigma_i (1 - \gamma (\lambda^\sigma_i)^{-1})}{1 - \frac{\gamma}{2 \lambda^\sigma_i}}\right) U^\top \\
&= U \operatorname{Diag} \left( \frac{\lambda^\sigma_i (\lambda^\sigma_i - \gamma)}{\lambda^\sigma_i - \frac{\gamma}{2}}\right) U^\top \\
&= U \Lambda_{\sigma,\gamma} U^\top
\end{align*}
where 
\begin{align} \label{eq:l_s_g_i}
\Lambda^{\sigma,\gamma} \eqdef \operatorname{Diag}(\lambda^{\sigma,\gamma}_i) \eqdef \operatorname{Diag} \left( \frac{\lambda^\sigma_i (\lambda^\sigma_i - \gamma)}{\lambda^\sigma_i - \frac{\gamma}{2}}\right).
\end{align}
Using the decomposition of the inverse Lyapunov operator$\Bigl(L^\gamma_{C_\sigma^{-1}}\Bigr)^{-1}$ detailed in Lemma~\ref{lem:SGD_3}, we deduce the expression of $X_1(\Delta)$
\begin{align*}
X_1(\Delta) &\eqdef{C_\mathrm{data}}^{1/2} \Sigma^\gamma_{1} (\Delta) {C_\mathrm{data}}^{1/2} \\ &= - U \left( \left( U^{\top} \Delta U \Lambda^{\sigma, \gamma} + \Lambda^{\sigma, \gamma} U^{\top} \Delta^\top U \right) \odot \left( K^{\sigma, \gamma}_{ij} \sqrt{\lambda_i \lambda_j} \right)_{ij} \right) U^\top
\end{align*}
where, for $1 \leq i,j \leq d$,
\begin{align} \label{eq:K_ij}
    K^{\sigma, \gamma}_{ij} \eqdef \frac{\lambda^\sigma_i \lambda^\sigma_j}{\lambda^\sigma_i + \lambda^\sigma_j - \gamma}
\end{align}
Then, using the decomposition of the inverse Lyapunov operator $L_{(X_0)^{1/2}}^{-1}$ given by Lemma~\ref{lem:SGD_3},
\begin{align*}
    L_{(X_0)^{1/2}}^{-1}[X_1(\Delta)] 
    &= U \left( (U^\top X_1(\Delta) U) \odot \left(\frac{1}{ \sqrt{\eta^{\sigma, \gamma}_i} + \sqrt{\eta^{\sigma, \gamma}_j} }\right)_{ij}\right) U^\top \\
    &= - U \left( \left(  U^{\top} \Delta U \Lambda^{\sigma, \gamma} + \Lambda^{\sigma, \gamma} U^{\top} \Delta^\top U \right) 
    \odot \left(\frac{K^{\sigma, \gamma}_{ij} \sqrt{\lambda_i \lambda_j}}{ \sqrt{\eta^{\sigma, \gamma}_i} + \sqrt{\eta^{\sigma, \gamma}_j} }\right)_{ij}\right) U^\top \\
    &= - U \left( \left(  U^{\top} \Delta U \Lambda^{\sigma, \gamma} + \Lambda^{\sigma, \gamma} U^{\top} \Delta^\top U \right) 
    \odot M^{\sigma, \gamma} \right) U^\top 
\end{align*}
where 
\begin{align*}
    M^{\sigma, \gamma}_{ij} &\eqdef \frac{K^{\sigma, \gamma}_{ij} \sqrt{\lambda_i \lambda_j}}{ \sqrt{\eta^{\sigma, \gamma}_i} + \sqrt{\eta^{\sigma, \gamma}_j} } .
\end{align*}
Now, we can calculate the full second term, again using the decomposition of $L_{(X_0)^{1/2}}^{-1}$:
\begin{align*}
\text{Term B} &=  2 \tr{L_{(X_0)^{1/2}}^{-1}\left[ \EE \left[\left(L_{(X_0)^{1/2}}^{-1}[X^1(\Delta)]\right)^2 \right]\right]} \\
&= \sum_{i=1}^d  \left( U^\top \EE \left[ \left(L_{(X^0_A)^{1/2}}^{-1}[X^1(\Delta)]\right)^2 \right] U \right)_{ii} \frac{1}{\sqrt{\eta^{\sigma, \gamma}_i}} \\
&= \sum_{i=1}^d \sum_{j=1}^d \EE\left[ \left(  \langle \Delta, u_iu_j^\top \rangle \lambda^{\sigma, \gamma}_j  + \langle \Delta, u_ju_i^\top \rangle \lambda^{\sigma, \gamma}_i \right)^2 \right]  \frac{(M^{\sigma, \gamma}_{ij})^2}{ \sqrt{\eta^{\sigma, \gamma}_i}} .
\end{align*}
Denoting $G^{\sigma, \gamma}_{ij} = G^{\sigma, \gamma}(\lambda_i, \lambda_j)$ defined by
\begin{align} \label{eq:G_ij}
G^{\sigma, \gamma}(\lambda, \lambda_j) =\frac{M^{\sigma, \gamma}_{ij}}{ (\eta^{\sigma, \gamma}_i)^{1/4}} 
    &= \frac{\lambda^\sigma_i \lambda^\sigma_j}{\lambda^\sigma_i + \lambda^\sigma_j - \gamma} \frac{1}{ \left(\frac{\lambda_i (\lambda^\sigma_i)^2}{\lambda^\sigma_i - \frac{\gamma}{2}}\right)^{1/4}} \frac{ \sqrt{\lambda_i \lambda_j}}{ \sqrt{\frac{\lambda_i (\lambda^\sigma_i)^2}{\lambda^\sigma_i - \frac{\gamma}{2}}} + \sqrt{\frac{\lambda_i (\lambda^\sigma_j)^2}{\lambda^\sigma_j - \frac{\gamma}{2}}}}
\end{align}

The term B further develops as
\begin{align*}
\text{Term B} &= \sum_{ij}   \left( G^{\sigma, \gamma}_{ij} \right)^2 (\lambda^{\sigma, \gamma}_j)^2 \EE_{\Delta}\left[\langle \Delta, u_i u_j^\top \rangle^2\right]  + \left( G^{\sigma, \gamma}_{ij} \right)^2 \left(\lambda^{\sigma, \gamma}_i\right)^2 \EE_{\Delta}\left[\langle \Delta, u_j u_i^\top \rangle\right] \\
&\nonumber +   \sum_{ij}  2\left( G^{\sigma, \gamma}_{ij} \right)^2 (\lambda^{\sigma, \gamma}_j) \left(\lambda^{\sigma, \gamma}_i\right) \EE_{\Delta}\left[\langle \Delta, u_i u_j^\top \rangle \langle \Delta, u_j u_i^\top \rangle \right] .
\end{align*}
Using the fact that $M_{ij}^{\sigma, \gamma}$ is symmetric, when swapping the indices $i$ and $j$ in the first term, we get the second one. Then both are equal and
\begin{align*}
\text{Term B} &= \sum_{ij}  2 \left( G^{\sigma, \gamma}_{ij} \right)^2 (\lambda^{\sigma, \gamma}_j)^2 \EE_{\Delta}\left[\langle \Delta, u_i u_j^\top \rangle^2\right] \\ &+ \nonumber  \sum_{ij} 2 \left( G^{\sigma, \gamma}_{ij} \right)^2 (\lambda^{\sigma, \gamma}_j) \left(\lambda^{\sigma, \gamma}_i\right) \EE_{\Delta}\left[\langle \Delta, u_i u_j^\top \rangle \langle \Delta, u_j u_i^\top \rangle \right]  .
\end{align*}

\paragraph{Term C.} First, in the same way as for the term B, using the decomposition of $L_{(X_0)^{1/2}}^{-1}$:
\begin{align*}
    2 \tr{L_{(X_0)^{1/2}}^{-1}\left[\EE[X_2(\Delta)]\right]} &= 2 \sum_i \left( U^\top \EE \left[ X_2(\Delta) \right] U \right)_{ii} \frac{1}{2 \sqrt{\eta^{\sigma,\gamma}_i}} \\
    &= \sum_i \left( U^\top \EE \left[ X_2(\Delta) \right] U \right)_{ii} \frac{1}{\sqrt{\eta^{\sigma,\gamma}_i}} .
\end{align*}
Recall that $X_2(\Delta) = {\Sigma}^{1/2}\Sigma^\gamma_{2} (\Delta) {\Sigma}^{1/2}$. Let us calculate $\Sigma^\gamma_{2} (\Delta)$ in the eigenbasis of the data. Recall the expression of $\Sigma^\gamma_2(\Delta)$ derived in Lemma~\ref{lem:Langevin_1}:
\begin{align*}
\Sigma_2^\gamma(\Delta) &= -\Bigl(L^\gamma_{{C_\sigma^{-1}}}\Bigr)^{-1}\Bigl[
\Delta \Sigma^\gamma_{{1}}(\Delta) (\id - \gamma {{C_\sigma^{-1}}}) + (\id - \gamma {{C_\sigma^{-1}}})\Sigma^\gamma_{{1}}(\Delta) \Delta^\top- \gamma   \Delta  \Sigma^\gamma_{{{C_\sigma^{-1}}}}  \Delta^\top
\Bigr] \\
&= \underbrace{-\Bigl(L^\gamma_{{C_\sigma^{-1}}}\Bigr)^{-1}\Bigl[
\Delta \Sigma^\gamma_{{1}}(\Delta) (\id - \gamma {{C_\sigma^{-1}}}) \Bigr]}_{\text{term } C_1} \underbrace{-\Bigl(L^\gamma_{{C_\sigma^{-1}}}\Bigr)^{-1}\Bigl[ (\id - \gamma {{C_\sigma^{-1}}})\Sigma^\gamma_{{1}}(\Delta) \Delta^\top \Bigr]}_{\text{term } C_2} \\ & \nonumber +  \underbrace{\gamma \Bigl(L^\gamma_{{C_\sigma^{-1}}}\Bigr)^{-1}\Bigl[\Delta  \Sigma^\gamma_{{{C_\sigma^{-1}}}}  \Delta^\top
\Bigr]}_{\text{term } C_3}.
\end{align*}
We look at the first term $C_1$. Using the calculation of $\Sigma^\gamma_{{1}}(\Delta)$ derived below~\eqref{eq:sigma_1},
\begin{align*}
\text{term } C_1 &= -\Bigl(L^\gamma_{{C_\sigma^{-1}}}\Bigr)^{-1}\Bigl[ \Delta \Sigma_1^\gamma(\Delta)(\id - \gamma {{C_\sigma^{-1}}}) \Bigr] \\
&= \Bigl(L^\gamma_{{C_\sigma^{-1}}}\Bigr)^{-1}\Bigl[  U (U^\top \Delta U) (U^\top \Sigma_1^\gamma(\Delta) U)\diag{\frac{\lambda^\sigma_i - \gamma}{\lambda^\sigma_i }}
U^\top  \Bigr]
\\
&= U \left( \Bigl[ ( U^{\top} \Delta U) \left(  (  U^{\top} \Delta U \Lambda^{\sigma, \gamma} + \Lambda^{\sigma, \gamma} U^{\top} \Delta^\top U) \odot  K^{\sigma, \gamma}\right) \diag{\frac{\lambda^\sigma_i - \gamma}{\lambda^\sigma_i }}
\Bigr] \odot K^{\sigma, \gamma} \right)U^{\top}  \\
&= U ( T_1(\Delta) + T_2(\Delta) ) U^\top.
\end{align*}
where, denoting $ \mathcal{U}(\Delta) \eqdef  U^{\top} \Delta U$ and dropping the $\sigma,\gamma$ indexes for readability, we have
\begin{align*}
T_1(\Delta)_{ij} &\eqdef \left(\mathcal{U}(\Delta)\left( \left(\mathcal{U}(\Delta) \Lambda \right) \odot K \right) \diag{\frac{\lambda^\sigma_i - \gamma}{\lambda^\sigma_i }} \right)_{ij}  K_{ij} \\
&= K_{ij} \sum_{k} \mathcal{U}(\Delta)_{ik} (\mathcal{U}(\Delta) \Lambda)_{kj} K_{kj} \frac{\lambda^\sigma_j - \gamma}{\lambda^\sigma_j } \\
&=  K_{ij} \sum_{k,l} \mathcal{U}(\Delta)_{ik} \mathcal{U}(\Delta)_{kl} \Lambda_{lj} K_{kj} \frac{\lambda^\sigma_j - \gamma}{\lambda^\sigma_j } .
\end{align*}
Using that $\Lambda$ is diagonal, 
\begin{align*}
T_1(\Delta)_{ij} =  K_{ij} \sum_{k} \Lambda_{jj} K_{kj} \frac{\lambda^\sigma_j - \gamma}{\lambda^\sigma_j } \mathcal{U}(\Delta)_{ik} \mathcal{U}(\Delta)_{kj}  .
\end{align*}
Similarly, 
\begin{align*}
T_2(\Delta)_{ij} &\eqdef  \left(\mathcal{U}(\Delta)( \Lambda \mathcal{U}(\Delta)^\top \odot K) \diag{\frac{\lambda^\sigma_i - \gamma}{\lambda^\sigma_i }} \right)_{ij}  K_{ij} \\
&=  K_{ij}\sum_{k} \mathcal{U}(\Delta)_{ik} ( \Lambda \mathcal{U}(\Delta)^\top)_{kj} K_{kj}\frac{\lambda^\sigma_j - \gamma}{\lambda^\sigma_j } \\
&=  K_{ij}\sum_{k,l} \mathcal{U}(\Delta)_{ik} \Lambda_{kl} \mathcal{U}(\Delta)_{jl}  K_{kj}\frac{\lambda^\sigma_j - \gamma}{\lambda^\sigma_j } \\
&=  K_{ij}\sum_{k} \Lambda_{kk}  K_{kj} \frac{\lambda^\sigma_j - \gamma}{\lambda^\sigma_j }\mathcal{U}(\Delta)_{ik} \mathcal{U}(\Delta)_{jk} .
\end{align*}
The term $C_2$ is the transpose of the term $C_1$, thus
\begin{align*}
\text{term } C_2 &= U ( T_1(\Delta)^\top + T_2(\Delta)^\top ) U^\top.
\end{align*}
And finally for the third term 
\begin{align*}
\text{term } C_3 &= U T_3(\Delta) U^\top
\end{align*}
with
\begin{align*}
    T_3(\Delta)_{ij} &\eqdef \gamma  \left(U^\top \Delta \Sigma^\gamma_{C_\sigma^{-1}} \Delta^\top U \right)_{ij}  K_{ij}\\
    &= \gamma   \left( \mathcal{U}(\Delta)  C_\sigma\left(\id - \frac{\gamma}{2}{{C_\sigma^{-1}}}\right)^{-1}  \mathcal{U}(\Delta)^\top \right)_{ij} K_{ij} \\
    &=  \left( \mathcal{U}(\Delta) \diag{\frac{ \gamma (\lambda^\sigma_i)^2}{\lambda^\sigma_i - \frac{\gamma}{2}}}  \mathcal{U}(\Delta)^\top \right)_{ij}  K_{ij} \\
    &=    K_{ij} \sum_{k} \frac{\gamma (\lambda^\sigma_k)^2}{\lambda^\sigma_k - \frac{\gamma}{2}} \mathcal{U}(\Delta)_{ik} \mathcal{U}(\Delta)_{jk}  .
\end{align*}
Overall
\begin{align*}
\Sigma_2^\gamma(\Delta) &= U \left(  T_1(\Delta) + T_2(\Delta) + T_1(\Delta)^\top + T_2(\Delta)^\top + T_3(\Delta) \right) U^\top .
\end{align*}
Thus, using $X_2(\Delta) \eqdef {C_\mathrm{data}}^{1/2}\Sigma^\gamma_{2} (\Delta) {C_\mathrm{data}}^{1/2}$,
\begin{align*}
 \text{term C}  &= \tr{\EE[\Sigma^\gamma_2(\Delta)]-2L_{(X_0)^{1/2}}^{-1}\left[\EE[X_2(\Delta)]\right]} \\ 
 &= \sum_i \left( \EE_{\Delta}\left[T_1(\Delta) + T_2(\Delta) + T_1(\Delta)^\top + T_2(\Delta)^\top + T_3(\Delta)  \right] \right)_{ii} \left(1 - \frac{\lambda_i}{ \sqrt{\eta^{\sigma,\gamma}_i}} \right)\\
 &= \sum_i \left( 2\EE_{\Delta}[T_1(\Delta)]  + 2\EE_{\Delta}[T_2(\Delta)] + \EE_{\Delta}[T_3(\Delta)] \right)_{ii} \left(1 - \frac{\lambda_i}{ \sqrt{\eta^{\sigma,\gamma}_i}} \right)
 \\
 &= \sum_{i,j} \, \hat g^{\sigma, \gamma}_i \, K^{\sigma, \gamma}_{ii} \,\left( 
2 \,  \lambda^{\sigma, \gamma}_{i}\, K^{\sigma, \gamma}_{i,j} \, \frac{\lambda^\sigma_i - \gamma}{\lambda^\sigma_i } \EE_{\Delta}\left[\mathcal{U}(\Delta)_{ij} \mathcal{U}(\Delta)_{ji}\right] \right. \\
&+ 2 \,  \lambda^{\sigma, \gamma}_{j} \, K^{\sigma, \gamma}_{i,j} \,\frac{\lambda^\sigma_i - \gamma}{\lambda^\sigma_i } \EE_{\Delta}\left[ \mathcal{U}(\Delta)_{ij}^2\right] + \left. \gamma 
 \frac{(\lambda^\sigma_j)^2}{\lambda^\sigma_i - \frac{\gamma}{2}} \EE_{\Delta}\left[ \mathcal{U}(\Delta)_{ij}^2\right]  \right) \nonumber
\end{align*}
where 
\begin{align} \label{eq:g_ij}
\hat g^{\sigma,\gamma}_i \eqdef 1 - \frac{\lambda_i}{\sqrt{\eta^{\sigma,\gamma}_i}}  &= 1 - \frac{\sqrt{\lambda_i}\sqrt{ \lambda_i^\sigma - \frac{\gamma}{2}}}{\lambda_i^\sigma}.
\end{align}
Overall, we can decompose the averaged squared Bures distance~\eqref{eq:bures_delta_main} as 
\begin{equation*}
\begin{split}
\EE_{\Delta}\left[\mathcal{B}^2( C_\mathrm{data}, \Sigma_\varepsilon^\gamma)\right] &=  \mathcal{E}^{(0)}_{\sigma, \gamma} +
\sum_{i,j=1}^d \kappa^{(1)}_{\sigma, \gamma}(\lambda_i, \lambda_j) \, \EE_{\Delta}\left[\langle \Delta, u_iu_j^\top \rangle^2\right] \\
&+ \sum_{i,j=1}^d  \kappa^{(2)}_{\sigma, \gamma}(\lambda_i, \lambda_j) \, \EE_{\Delta}\left[\langle \Delta, u_iu_j^\top \rangle \langle \Delta, u_ju_i^\top \rangle \right]  + o(\varepsilon^2),
\end{split}
\end{equation*}
where $\kappa^{(1)}_{\sigma, \gamma}, \kappa^{(2)}_{\sigma, \gamma} : \RR^4\to  \RR$ are kernels which quantify the interactions between the eigenvalues~$\lambda_i$ of~$C_\mathrm{data}$. We give here their analytical form with respect to $G^{\sigma, \gamma}_{ij}$ (equation~\ref{eq:G_ij}), $K^{\sigma, \gamma}_{ij}$ (equation~\ref{eq:K_ij}), $\hat g^{\sigma, \gamma}_{ij}$(equation~\ref{eq:g_ij}), and $\lambda^{\sigma, \gamma}_i$ (equation~\ref{eq:l_s_g_i}), which all depend on $\lambda_i, \lambda_j$:
\begin{align} \label{eq:kappa1}
    \kappa^{(1)}_{\sigma, \gamma}(\lambda_i, \lambda_j) &=  2 ( G^{\sigma, \gamma}_{ij})^2  (\lambda^{\sigma, \gamma}_j)^2 
    + 4 \hat g^{\sigma, \gamma}_i K^{\sigma, \gamma}_{ii} \lambda^{\sigma, \gamma}_{j} \, K^{\sigma, \gamma}_{ij} \,\frac{\lambda^\sigma_i - \gamma}{\lambda^\sigma_i } 
    + 2 \gamma \hat g^{\sigma, \gamma}_i K^{\sigma, \gamma}_{ii}
 \frac{(\lambda^\sigma_j)^2}{\lambda^\sigma_j - \frac{\gamma}{2}}
\end{align}

\begin{align} \label{eq:kappa2}
 \kappa^{(2)}_{\sigma, \gamma}(\lambda_i, \lambda_j)  &= 2  (G^{\sigma, \gamma}_{ij})^2 (\lambda^{\sigma, \gamma}_j) \left(\lambda^{\sigma, \gamma}_i\right) +  4 \hat g^{\sigma, \gamma}_i K^{\sigma, \gamma}_{ii} \lambda^{\sigma, \gamma}_{i} \, K^{\sigma, \gamma}_{ij} \,\frac{\lambda^\sigma_i - \gamma}{\lambda^\sigma_i } 
\end{align}

\end{proof}
\paragraph{Simplifications for $\gamma=0$ and $\sigma \to 0$.} 
In the limit of a zero stepsize $\gamma = 0$ and small $\sigma \to 0$, we have $\hat g^{0, 0}_{ij} = 0$ and the above kernels simplify to 
\begin{align} \label{eq:kappa_simp}
   \kappa^{(1)}_{0, 0}(\lambda_i, \lambda_j) &= 2 ( G^{0, 0}_{ij})^2  (\lambda^{0, 0}_j)^2 = 2 \frac{ (\lambda_i \lambda_j)^3}{(\lambda_i + \lambda_j)^4} \frac{\lambda_j^2}{\lambda_i} = 2 \frac{ \lambda_i^2 \lambda_j^5}{(\lambda_i + \lambda_j)^4}  \\
   \kappa^{(2)}_{0, 0}(\lambda_i, \lambda_j) &=  2 ( G^{0, 0}_{ij})^2  (\lambda^{0, 0}_j) (\lambda^{0, 0}_i) = 2 \frac{ (\lambda_i \lambda_j)^3}{(\lambda_i + \lambda_j)^4} \lambda_j =
   2 \frac{ (\lambda_i)^3 (\lambda_j)^4}{(\lambda_i + \lambda_j)^4} .
\end{align}

\subsection{Proof of Corollary~\ref{cor:langevin_final}} \label{app:langevin_final}
\newtheorem*{repeatcor1}{Corollary 1}
\begin{repeatcor1}[ULA sampling error with linear score trained by SGD]
Under assumption~\ref{ass:gaussian}, consider a linear score $v_\sigma(x, \theta) = - A x + b$ with parameters \( \theta=(A,b) \) trained via the SGD algorithm~\eqref{eq:score_matching_SGD_emp}, with \( N \) data samples and constant learning rate \( \tau > 0 \). Then, for small enough stepsize $\gamma$, the ULA algorithm~\eqref{eq:ULA} samples an invariant Gaussian distribution $q$ that satisfies \vspace{-0.1cm}
\begin{equation*}
 \! \! \mathbb{E}_{\theta}\left[  W_2\left(p, q\right)^2 \right] \! =  
 \mathcal{E}^{(0)}_{\sigma, \gamma} \! +  \! \!  \sum_{i,j = 1}^d  \! \left( \! \frac{\tau}{\sigma^2} k_{\sigma, \gamma}^\tau (\lambda_i, \lambda_j)  \! +  \!\frac{\tau}{N \sigma^2} k_{\sigma, \gamma}^{\tau, N}(\lambda_i, \lambda_j)  \!+  \! \frac{1}{N} k_{\sigma, \gamma}^N(\lambda_i, \lambda_j) \! \right)  \! + \! o\left(\! \tau, \frac{1}{N} \! \right).
 \end{equation*} 
$k_{\sigma, \gamma}^\tau, k_{\sigma, \gamma}^N, k_{\sigma, \gamma}^{\tau, N} : \RR^d \times \RR^d \to \RR$ are kernel functions capturing interactions between the eigenvalues of the $C_\mathrm{data}$. Their expressions, with respect to the kernels $\kappa^{(1)}_{\sigma, \gamma}$ and $\kappa^{(2)}_{\sigma, \gamma}$ defined in Theorem~\ref{thm:langevin_distance_general} are:
\begin{align*}
k_{\sigma, \gamma}^\tau (\lambda_i, \lambda_j) &= \frac{1}{2}\frac{\lambda_j}{\lambda^\sigma_j}\kappa^{(1)}_{\sigma, \gamma}(\lambda_i, \lambda_j)+  \mathbf{1}_{i=j} \frac{1}{2}\frac{\lambda_i}{\lambda^\sigma_i}\left(\left(\lambda^\sigma_i\right)^2  + \kappa^{(2)}_{\sigma, \gamma}(\lambda_i, \lambda_i)\right) \\
k_{\sigma, \gamma}^N (\lambda_i, \lambda_j) &= \left(\kappa^{(1)}_{\sigma, \gamma}(\lambda_i, \lambda_j) + \kappa^{(2)}_{\sigma, \gamma}(\lambda_i, \lambda_j) \right) \frac{\lambda_i \lambda_j}{(\lambda^\sigma_i \lambda^\sigma_j)^2} \\
&+ \nonumber \mathbf{1}_{i=j}\left(  \lambda_i +  \left(\kappa^{(1)}_{\sigma, \gamma}(\lambda_i, \lambda_i) + \kappa^{(2)}_{\sigma, \gamma}(\lambda_i, \lambda_i) \right) \frac{\lambda_i^2}{(\lambda^\sigma_i)^4}\right)
\\
k_{\sigma, \gamma}^{\tau, N} (\lambda_i, \lambda_j) &= 
- \left( \frac{\lambda_j}{\lambda^\sigma_j}\right)^2  \kappa^{(1)}_{\sigma, \gamma}(\lambda_i, \lambda_j) \\ 
&+ \nonumber \mathbf{1}_{i=j} \left(\frac{1}{2}(\lambda^\sigma_i)^2 \sum_k \frac{\lambda_k^2}{\lambda^\sigma_k} - \left( \frac{\lambda_i}{\lambda^\sigma_i}\right)^2\left((\lambda^\sigma_i)^2  + \kappa^{(2)}_{\sigma, \gamma}(\lambda_i, \lambda_i)\right) \right)
\end{align*}
Their simplified expressions for $\gamma = 0$ and $\sigma \to 0$ are given equations~\eqref{eq:ktau_simp},~\eqref{eq:kN_simp} and~\eqref{eq:ktauN_simp}. 
\end{repeatcor1}
\begin{remark}
The above kernels are not written in symmetric form but can be symmetrized by the operation 
\begin{align} \label{eq:sym}
    \hat k_{\sigma, \gamma} (\lambda_i, \lambda_j)  = \frac{1}{2} \left(k_{\sigma, \gamma} (\lambda_i, \lambda_j)  + k_{\sigma, \gamma} (\lambda_j, \lambda_i)\right)
\end{align}
without changing the value of the $W_2$ error.
\end{remark}
\begin{proof}
Applying Theorem~\ref{thm:SGD_error}, the score $v_\sigma(x, \theta) = -A x + b$ satisfies
\begin{equation*}
\begin{split}
    b = C_\sigma^{-1}\mu_\mathrm{data} + \delta + \Delta \mu_\mathrm{data} \quad &\text{and} \quad
    A = C_\sigma^{-1} + \Delta, \\ 
    \quad\text{where}\quad \delta \sim \mathcal{N}(0, \cov(\kappa)) \quad &\text{and}\quad \Delta \sim \mathcal{N}(0, \cov(A)),
\end{split}
\end{equation*}
with (denoting $P_\sigma \eqdef C_\sigma^{-1} C_\mathrm{data}$ and $Q_\sigma \eqdef C_\sigma^{-2}C_\mathrm{data}$)
\begin{equation*}
\begin{split}
\cov(\kappa) &= \frac{\tau}{2 \sigma^2} P_\sigma +  \frac{1}{N} Q_\sigma + \frac{\tau}{N \sigma^2} \left( \frac12 \langle P_\sigma, C_\mathrm{data} \rangle \id -  P_\sigma^2 \right) 
+ o \left(\tau, \frac{1}{N}\right) 
\end{split}
\end{equation*}
\begin{equation*}
\begin{split}
\cov(A)  = \frac{\tau}{2\sigma^2} P_\sigma \otimes \id + \frac{1}{N} \left( Q_\sigma \otimes Q_\sigma \right) (I_{d^2} + P_{trans})  -  \frac{\tau}{N \sigma^2} P_\sigma^2 \otimes \id  + o\left(\tau, \frac{1}{N}\right) .
\end{split}
\end{equation*}
Applying Theorem~\ref{thm:langevin_distance_general}, the ULA algorithm with score $v_\sigma$ converges to an invariant Gaussian distribution $q_\varepsilon$ that satisfies
\begin{equation} \label{eq:292}
\EE_{\delta, \Delta}  \left[ W_2^2(p_\mathrm{data},q_\varepsilon) \right] =  \mathcal{E}^{(0)}_{\sigma, \gamma} + \left( \mathcal{E}_{\sigma, \gamma}^{\text{mean}}(\delta) + \mathcal{E}_{\sigma, \gamma}^{\text{cov}}(\Delta) \right) + o\left(\norm{\delta}, \norm{\Delta}\right)
\end{equation}
\begin{equation} 
\begin{split}
\text{where} &\quad 
\mathcal{E}^{(0)}_{\sigma, \gamma} = \sum_i \left(\sqrt{\lambda_i} - \frac{\lambda_i^\sigma}{\sqrt{\lambda_i^\sigma - \frac{\gamma}{2}}}\right)^2, \qquad
\mathcal{E}_{\sigma, \gamma}^{\text{mean}}(\delta) = \sum_{i=1}^d \sqrt{\lambda^\sigma_i } \, \langle \cov(\delta), u_i u_i^\top \rangle, \label{eq:eps_0_bis} \\ \vspace{-0.2cm}
\mathcal{E}_{\sigma, \gamma}^{\text{cov}}(\Delta) &= \sum_{i,j=1}^d \kappa^{(1)}_{\sigma, \gamma}(\lambda_i, \lambda_j) \, \EE_{\Delta}\left[\langle \Delta, u_iu_j^\top \rangle^2\right] + \kappa^{(2)}_{\sigma, \gamma}(\lambda_i, \lambda_j) \, \EE_{\Delta}\left[\langle \Delta, u_iu_j^\top \rangle \langle \Delta, u_ju_i^\top \rangle \right] 
\end{split}
\end{equation}
We calculate, in this order, $\langle \cov(\delta), u_i u_i^\top \rangle $, $\EE[\langle \Delta, u_i u_j^\top \rangle ^2]$ and $\EE_{\Delta}\left[ \langle \Delta, u_i u_j^\top \rangle \langle \Delta, u_j u_i^\top \rangle \right]$:
\begin{align*}
    \langle \cov(\delta), u_i u_i^\top \rangle &=  \langle \cov(\kappa), u_i u_i^\top \rangle \\
    &= \frac{\tau}{2\sigma^2} \langle P_\sigma, u_i u_i^\top \rangle 
    +  \frac{\tau}{N \sigma^2} \left( 
    \left\langle \frac12 \langle P_\sigma, C_\mathrm{data} \rangle \id -  P_\sigma^2, u_i u_i^\top \right\rangle \right) \\ \nonumber
    &+ \frac{1}{N} \langle Q_\sigma, u_i u_i^\top \rangle 
    +  o \left(\tau, \frac{1}{N}\right) \\
    &= \frac{\tau}{2\sigma^2} \frac{\lambda_i}{\lambda^\sigma_i}
    +  \frac{\tau}{N \sigma^2} \left( \frac{1}{2} \sum_k \frac{\lambda_k^2}{\lambda^\sigma_k}- \left(\frac{\lambda_i}{\lambda^\sigma_i}\right)^2 \right) 
    + \frac{1}{N} \frac{\lambda_i}{(\lambda^\sigma_i)^2}
    +  o \left(\tau, \frac{1}{N}\right) 
\end{align*}


\begin{align} 
\EE[\langle \Delta, u_i u_j^\top \rangle ^2]   &=\text{vec}(u_i u_j^\top)^\top \cov(A) \text{vec}(u_i u_j^\top) \\
    &= \text{vec}(u_j \otimes u_i)^\top \cov(A) vec(u_j \otimes u_i) \\
    &=  \label{eq:ps_delta} \tau  \underbrace{ \frac{1}{2\sigma^2} \text{vec}(u_j \otimes u_i)^\top (P_\sigma \otimes \id) vec(u_j \otimes u_i)}_{\text{term 1}} \\
    &- \nonumber \frac{\tau}{N}  \underbrace{\frac{1}{\sigma^2} \text{vec}(u_j \otimes u_i)^\top (P_\sigma^2 \otimes \id) vec(u_j \otimes u_i)}_{\text{term 2}} \\ \nonumber &+ \frac{1}{N} \underbrace{\text{vec}(u_j \otimes u_i)^\top \left( Q_\sigma \otimes Q_\sigma \right) (I_{d^2} + P_{trans}) vec(u_j \otimes u_i)}_{\text{term 3}} + o \left(\tau, \frac{1}{N}\right) 
    \end{align}
For the first term of~\eqref{eq:ps_delta}, using that $P_\sigma = C_\sigma^{-1} C_\mathrm{data}$,
 \begin{align*}
 \text{term 1} 
 &= \frac{1}{2\sigma^2}\text{vec}(u_j \otimes u_i)^\top (\left( C_\sigma^{-1} C_\mathrm{data} \right) \otimes \id) (u_j \otimes u_i) \\
    &= \frac{1}{2\sigma^2} u_j^\top \left( C_\sigma^{-1} C_\mathrm{data} \right) u_j \\
    &=  \frac{1}{2\sigma^2} \frac{\lambda_j}{\lambda^\sigma_j}
\end{align*}
Similarly, for the second term of~\eqref{eq:ps_delta}, we get 
\begin{align*}
    \text{term 2} &=  \frac{1}{\sigma^2}\left( \frac{\lambda_j}{\lambda^\sigma_j}\right)^2.
\end{align*}
For the last term of~\eqref{eq:ps_delta}, we use that $ P_{trans} \text{vec}(u_i u_j^\top) =  P_{trans} \text{vec}(u_j u_i^\top)$, and thus 
\begin{align*}
      \text{term 3}  &= \text{vec}(u_i u_j^\top)^\top (C_\sigma^{-2} C_\mathrm{data} \otimes C_\sigma^{-2} C_\mathrm{data})(I_{d^2} + P_{trans}) \text{vec}(u_i u_j^\top) \\
    &= \text{vec}(u_i u_j^\top)^\top (C_\sigma^{-2} C_\mathrm{data} \otimes C_\sigma^{-2} C_\mathrm{data}) \text{vec}(u_i u_j^\top) \\
    &+ \nonumber \text{vec}(u_i u_j^\top)^\top (C_\sigma^{-2} C_\mathrm{data} \otimes C_\sigma^{-2} C_\mathrm{data}) P_{trans} \text{vec}(u_i u_j^\top) \\
     &= (u_j \otimes u_i)^\top (C_\sigma^{-2} C_\mathrm{data} \otimes C_\sigma^{-2} C_\mathrm{data})  (u_j \otimes u_i) \\
     &+ \nonumber (u_j \otimes u_i)^\top (C_\sigma^{-2} C_\mathrm{data} \otimes C_\sigma^{-2} C_\mathrm{data})  (u_i \otimes u_j) \\
    &= (u_j^\top C_\sigma^{-2} C_\mathrm{data} u_j) (u_i^\top C_\sigma^{-2} C_\mathrm{data}  u_i) + (u_j^\top C_\sigma^{-2} C_\mathrm{data} u_i) (u_i^\top  C_\sigma^{-2} C_\mathrm{data} u_j) \\
    &= \frac{\lambda_i \lambda_j}{(\lambda^\sigma_i \lambda^\sigma_j)^2}( 1  + \mathbf{1}_{i=j}).
    \end{align*}
And overall,~\eqref{eq:ps_delta} simplifies to
\begin{align*}
     \EE[\langle \Delta, u_i u_j^\top \rangle ^2]
      &=  \frac{\tau}{2\sigma^2} \frac{\lambda_j}{\lambda^\sigma_j} - \frac{\tau}{N \sigma^2 } \left( \frac{\lambda_j}{\lambda^\sigma_j}\right)^2 + \frac{1}{N}\frac{\lambda_i \lambda_j}{(\lambda^\sigma_i \lambda^\sigma_j)^2}( 1  + \mathbf{1}_{i=j})  + o \left(\tau, \frac{1}{N}\right) 
    \end{align*}
Finally, we calculate $\EE[\langle \Delta, u_i u_j^\top \rangle \langle \Delta, u_j u_i^\top \rangle ] $ in a similar way:
\begin{align} 
\EE[\langle \Delta, u_i u_j^\top \rangle \langle \Delta, u_j u_i^\top \rangle ]  
      &=\text{vec}(u_i u_j^\top)^\top \cov(A) \text{vec}(u_j u_i^\top) \\
    &= \text{vec}(u_j \otimes u_i)^\top \cov(A) vec(u_i \otimes u_j) \\
    &= \tau  \underbrace{ \frac{1}{2\sigma^2} \text{vec}(u_j \otimes u_i)^\top (P_\sigma \otimes \id) vec(u_i \otimes u_j)}_{\text{term 1}}  \label{eq:ps_delta_2} \\
    &- \nonumber \frac{\tau}{N}  \underbrace{\frac{1}{\sigma^2} \text{vec}(u_j \otimes u_i)^\top (P_\sigma^2 \otimes \id) vec(u_i \otimes u_j)}_{\text{term 2}} \\ \nonumber &+ \frac{1}{N} \underbrace{\text{vec}(u_j \otimes u_i)^\top \left( Q_\sigma \otimes Q_\sigma \right) (I_{d^2} + P_{trans}) vec(u_i \otimes u_j)}_{\text{term 3}} + o \left(\tau, \frac{1}{N}\right) .
    \end{align}
For the term 1 of~\eqref{eq:ps_delta_2}, 
 \begin{align*}
 \text{term 1} 
 &= \frac{1}{2\sigma^2}\text{vec}(u_j \otimes u_i)^\top (\left( C_\sigma^{-1} C_\mathrm{data} \right) \otimes \id) (u_i \otimes u_j) \\
    &= \frac{1}{2\sigma^2} \mathbf{1}_{i=j} u_j^\top \left( C_\sigma^{-1} C_\mathrm{data} \right) u_i  \\
    &=  \frac{1}{2\sigma^2} \mathbf{1}_{i=j} \frac{\lambda_i}{\lambda^\sigma_i}
\end{align*}
Similarly, for the term 2 of~\eqref{eq:ps_delta_2}, we get 
\begin{align*}
    \text{term 2} &= - \frac{1}{\sigma^2} \mathbf{1}_{i=j}\left( \frac{\lambda_i}{\lambda^\sigma_i}\right)^2.
\end{align*}
and for the term 3 of~\eqref{eq:ps_delta_2}, using that $ P_{trans} \text{vec}(u_j u_i^\top) =  P_{trans} \text{vec}(u_i u_j^\top)$,
\begin{align*}
      \text{term 3}  &= \text{vec}(u_i u_j^\top)^\top (C_\sigma^{-2} C_\mathrm{data} \otimes C_\sigma^{-2} C_\mathrm{data})(I + P_{trans}) \text{vec}(u_j u_i^\top) \\
    &= (u_j^\top C_\sigma^{-2} C_\mathrm{data} u_i) (u_i^\top C_\sigma^{-2} C_\mathrm{data}  u_j) + (u_j^\top C_\sigma^{-2} C_\mathrm{data} u_j) (u_i^\top  C_\sigma^{-2} C_\mathrm{data} u_i) \\
    &= \frac{\lambda_i \lambda_j}{(\lambda^\sigma_i \lambda^\sigma_j)^2}( 1  + \mathbf{1}_{i=j}) .
    \end{align*}
Overall we get for~\eqref{eq:ps_delta_2} 
\begin{equation*}
\begin{split}
     \EE[\langle \Delta, u_i u_j^\top \rangle \langle \Delta, u_j u_i^\top \rangle ]  
      &=  \frac{\tau}{2\sigma^2} \mathbf{1}_{i=j} \frac{\lambda_i}{\lambda^\sigma_i} - \frac{\tau}{N \sigma^2 } \mathbf{1}_{i=j} \left( \frac{\lambda_i}{\lambda^\sigma_i}\right)^2 \\
      &+  \frac{1}{N}\frac{\lambda_i \lambda_j}{(\lambda^\sigma_i \lambda^\sigma_j)^2}( 1  + \mathbf{1}_{i=j})  + o \left(\tau, \frac{1}{N}\right) .
\end{split}
 \end{equation*}

With the above calculations, the term $ \mathcal{E}_{\sigma, \gamma}^{\text{mean}}(\delta) + \mathcal{E}_{\sigma, \gamma}^{\text{cov}}(\Delta)$ in~\eqref{eq:292} can be written as the sum of three kernel norms over the eigenvalues $\lambda_i$ of $C_{\mathrm{data}}$, as defined in the corollary.


\end{proof}

\paragraph{Simplifications for $\gamma=0$ and $\sigma \to 0$.} Using the calculations from~\eqref{eq:kappa_simp}, in the limit of zero stepsize $\gamma=0$ (continuous limit) and zero noise level $\sigma \to 0$, the three kernels simplify to 
\begin{align*}
    k_{0, 0}^\tau (\lambda_i, \lambda_j) &= \frac{ \lambda_i^2 (\lambda_j)^5}{(\lambda_i + \lambda_j)^4} +  \mathbf{1}_{i=j} \frac{1}{2} \left( \lambda_i^2  + \frac{1}{8} \lambda_i^3\right)\\ 
    k_{0, 0}^N (\lambda_i, \lambda_j) &=  2 \frac{ \lambda_i(\lambda_j)^4 + \lambda_i^2 (\lambda_j)^3}{(\lambda_i + \lambda_j)^4} + \mathbf{1}_{i=j}\left(  \lambda_i + \frac{1}{2} \lambda_i \right) \\
    k_{0, 0}^{\tau, N} (\lambda_i, \lambda_j) &= 
- 2 \frac{ \lambda_i^2 (\lambda_j)^5}{(\lambda_i + \lambda_j)^4} + \mathbf{1}_{i=j} \left(\frac{1}{2}(\lambda_i)^2 \left( \sum_k \lambda_k - 2 \right)  - \frac{1}{8} \lambda_i^3\right) 
\end{align*}
Their symmetrized form~\eqref{eq:sym} is 
\begin{align} \label{eq:ktau_simp}
    \hat k_{0, 0}^\tau (\lambda_i, \lambda_j) &= \frac{1}{2}(\lambda_i \lambda_j)^2 \frac{ \lambda_i^3 + \lambda_j^3}{(\lambda_i + \lambda_j)^4} +  \mathbf{1}_{i=j} \frac{1}{2} \left( \left(\lambda_i\right)^2  + \frac{1}{8} \lambda_i^3\right)\\ \label{eq:kN_simp}
    \hat k_{0, 0}^N (\lambda_i, \lambda_j) &= \lambda_i \lambda_j \frac{\lambda_i^3 + \lambda_j^3 + \lambda_i \lambda_j(\lambda_i + \lambda_j)}{(\lambda_i + \lambda_j)^4} + \mathbf{1}_{i=j}\left(  \lambda_i + \frac{1}{2} \lambda_i \right) \\ \label{eq:ktauN_simp}
    \hat k_{0, 0}^{\tau, N} (\lambda_i, \lambda_j) &= 
- (\lambda_i \lambda_j)^2 \frac{ \lambda_i^3 + \lambda_j^3}{(\lambda_i + \lambda_j)^4} + \mathbf{1}_{i=j} \left(\frac{1}{2}(\lambda_i)^2 \left( \sum_k \lambda_k - 2 \right)  - \frac{1}{8} \lambda_i^3\right) 
\end{align}
\subsection{Trade-off in the choice of the optimal $\sigma$ parameter}
\label{app:analysis_Langevin}

Recall that, from Corollary~\ref{cor:langevin_final}, the $W_2$ error of Langevin sampling writes
\begin{equation} \label{eq:langevin_W2_bis}
 \! \! \mathbb{E}_{\theta}\left[  W_2\left(p, q\right)^2 \right] \! =  
 \mathcal{E}^{(0)}_{\sigma, \gamma} \! +  \! \!  \sum_{i,j = 1}^d  \! \left( \! \frac{\tau}{\sigma^2} k_{\sigma, \gamma}^\tau (\lambda_i, \lambda_j)  \! +  \!\frac{\tau}{N \sigma^2} k_{\sigma, \gamma}^{\tau, N}(\lambda_i, \lambda_j)  \!+  \! \frac{1}{N} k_{\sigma, \gamma}^N(\lambda_i, \lambda_j) \! \right)  \! + \! o\left(\! \tau, \frac{1}{N} \! \right). 
 \end{equation} 
As shown in Figure~\ref{fig:langevin}, the evolution of this error with respect to $\sigma$ is bell-shaped. Indeed, the first term~$\mathcal{E}^{(0)}$, defined in~\eqref{eq:eps_0_bis} tends to $0$ when $\sigma \to 0$ while the second and third terms diverge for small~\(\sigma\).

\paragraph{Error analysis for small $\sigma$}  To explain this behavior, we calculate the expansion of~\eqref{eq:langevin_W2_bis} for small~$\sigma$. For simplicity, we consider the continuous limit case~$\gamma = 0$. The zeroth-order term then expands as
\begin{align*}
    \mathcal{E}^{(0)}_{\sigma, 0} = \sum_i \left(\sqrt{\lambda_i} - \sqrt{\lambda_i^\sigma}\right)^2 &= \sum_i  \lambda_i \left( 1 - \sqrt{1 + \frac{\sigma^2}{\lambda_i}}\right)^2 \\
    &= \sum_i  \frac{\sigma^4}{4 \lambda_i} + o(\sigma^4) \\
    &= \left( \sum_i\frac{1}{4 \lambda_i} \right)\sigma^4 + o(\sigma^4)
\end{align*}
On the other hand, we saw equations~\eqref{eq:ktau_simp} and~\eqref{eq:ktauN_simp} that $k^\tau_{\sigma, 0} \to k^\tau_{0, 0}$, $k^{\tau,N}_{\sigma, 0} \to k^{\tau,N}_{0, 0}$ and $k^{N}_{\sigma, 0} \to k^{N}_{0, 0}$ i.e. have finite limits when $\sigma \to 0$. Therefore, for small $\sigma$,
\begin{align} \label{eq:langevin_W2_sigma_small}
 \! \! \mathbb{E}_{\theta}\left[  W_2\left(p, q\right)^2 \right] \! &\approx \left( \sum_i\frac{1}{4 \lambda_i} \right) \sigma^4 + \frac{\tau}{\sigma^2} \sum_{i,j}  \left(  k^\tau_{0, 0}(\lambda_i, \lambda_j) + \frac{1}{N} k^{\tau,N}_{0, 0}(\lambda_i, \lambda_j) \right) \\
&\approx \left( \sum_i\frac{1}{4 \lambda_i} \right) \sigma^4 + \frac{\tau}{\sigma^2} \sum_{i,j}  k^\tau_{0, 0}(\lambda_i, \lambda_j)  
\end{align} 
 which is minimized for
 \begin{equation} \label{eq:approx_sigma_star}
 \sigma^*(\tau, N) = \left( 2 \frac{ \sum_{i,j} k^\tau_{0, 0}(\lambda_i, \lambda_j)}{\left( \sum_i\frac{1}{\lambda_i} \right)} \right)^{1/6} \tau^{1/6} = C(\lambda) \tau^{1/6}
  \end{equation}
Using the expression~\eqref{eq:ktau_simp} of $k^\tau_{0, 0}$, we can calculate explicitly the constant $C$ with respect to the power spectrum $\lambda = (\lambda_i)_i$.

\begin{figure}[h]
    \centering
    \begin{minipage}[b]{0.45\linewidth}
        \centering
        \includegraphics[width=0.7\linewidth]{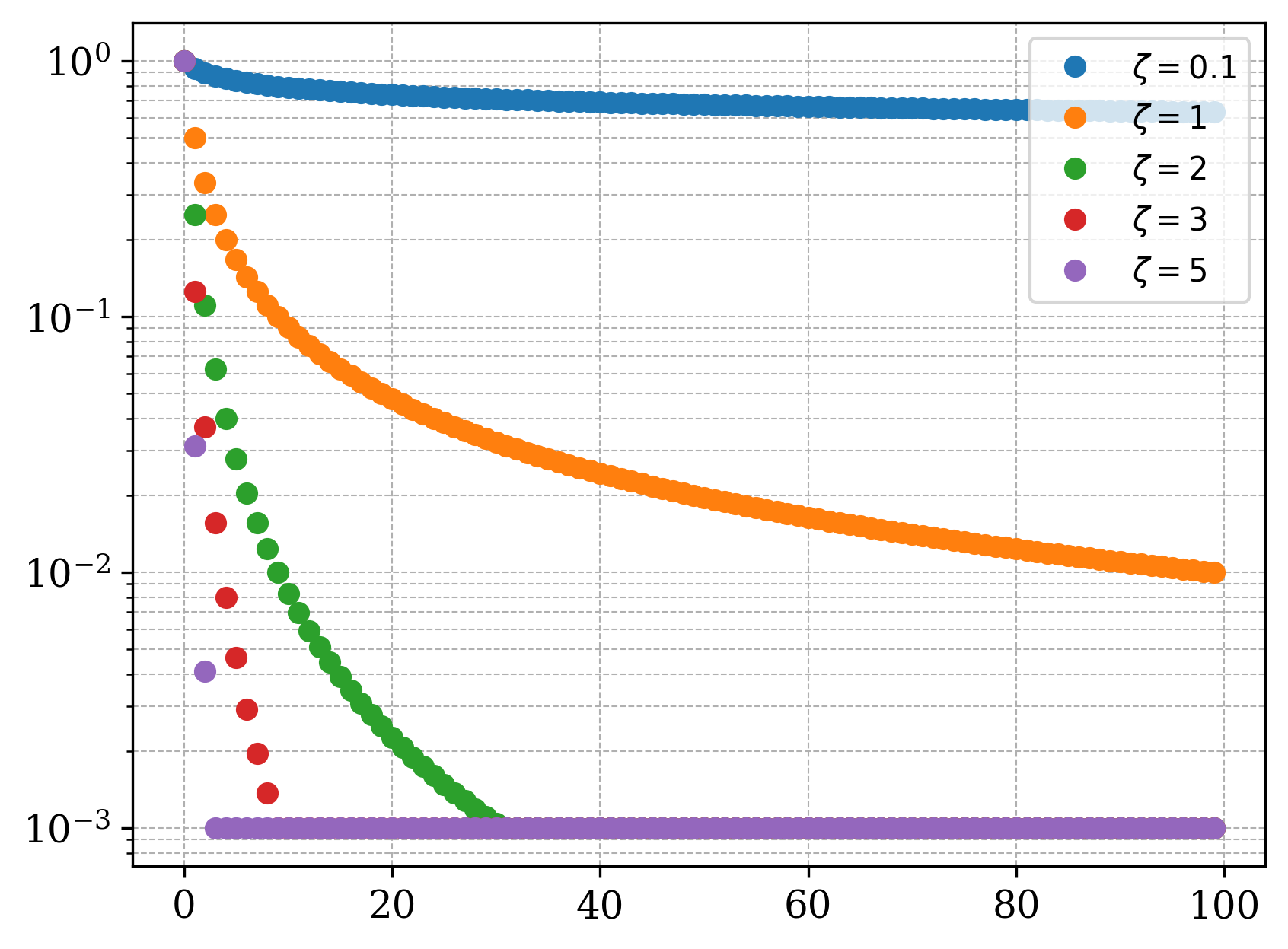}
        \caption{Power spectrum of $C_{\mathrm{data}}$ for different values of the power law coefficient $\zeta$.}
        \label{fig:power_spectrum}
    \end{minipage}
    \hfill
    \begin{minipage}[b]{0.45\linewidth}
        \centering
        \includegraphics[width=0.8\linewidth]{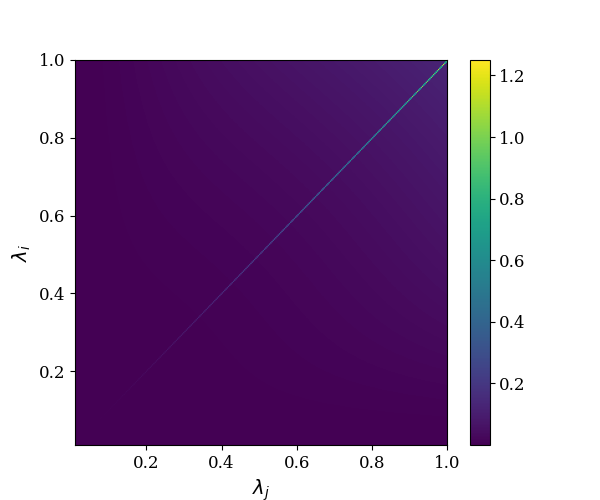}
        \caption{$k^\tau_{0,0}(\lambda_i, \lambda_j)$ for $\lambda_i, \lambda_j \in [0,1]$}
        \label{fig:heatmap}
    \end{minipage}
\end{figure}

\begin{figure}[h]
    \centering
    \begin{subfigure}[b]{0.3\textwidth}
    \centering
     \includegraphics[width=\linewidth]{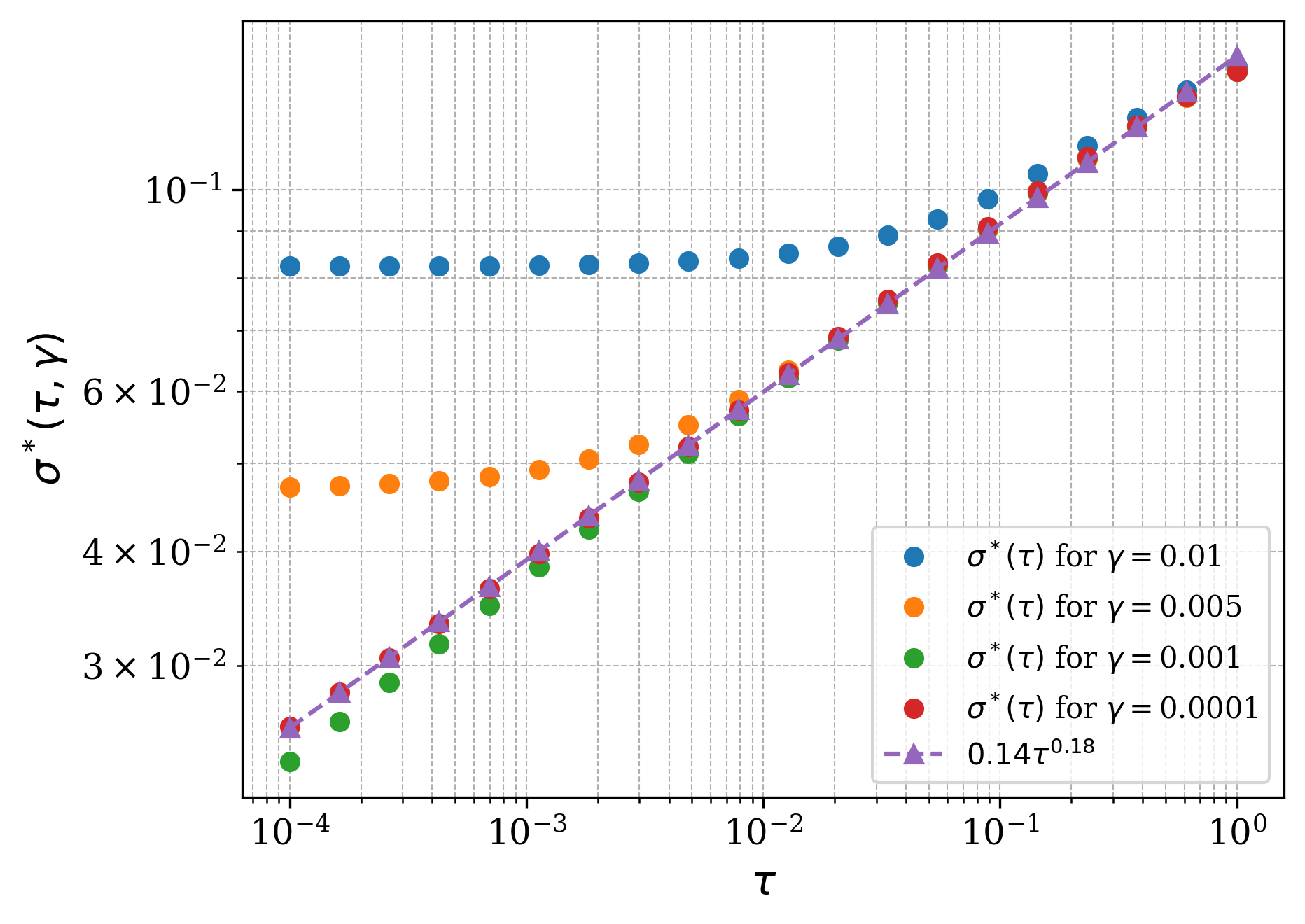}
     \caption{$\sigma^*(\tau)$ for various $\gamma$.}
     \label{fig:sigma_opt_tau}
    \end{subfigure}
    \begin{subfigure}[b]{0.3\textwidth}
    \centering
    \includegraphics[width=\linewidth]{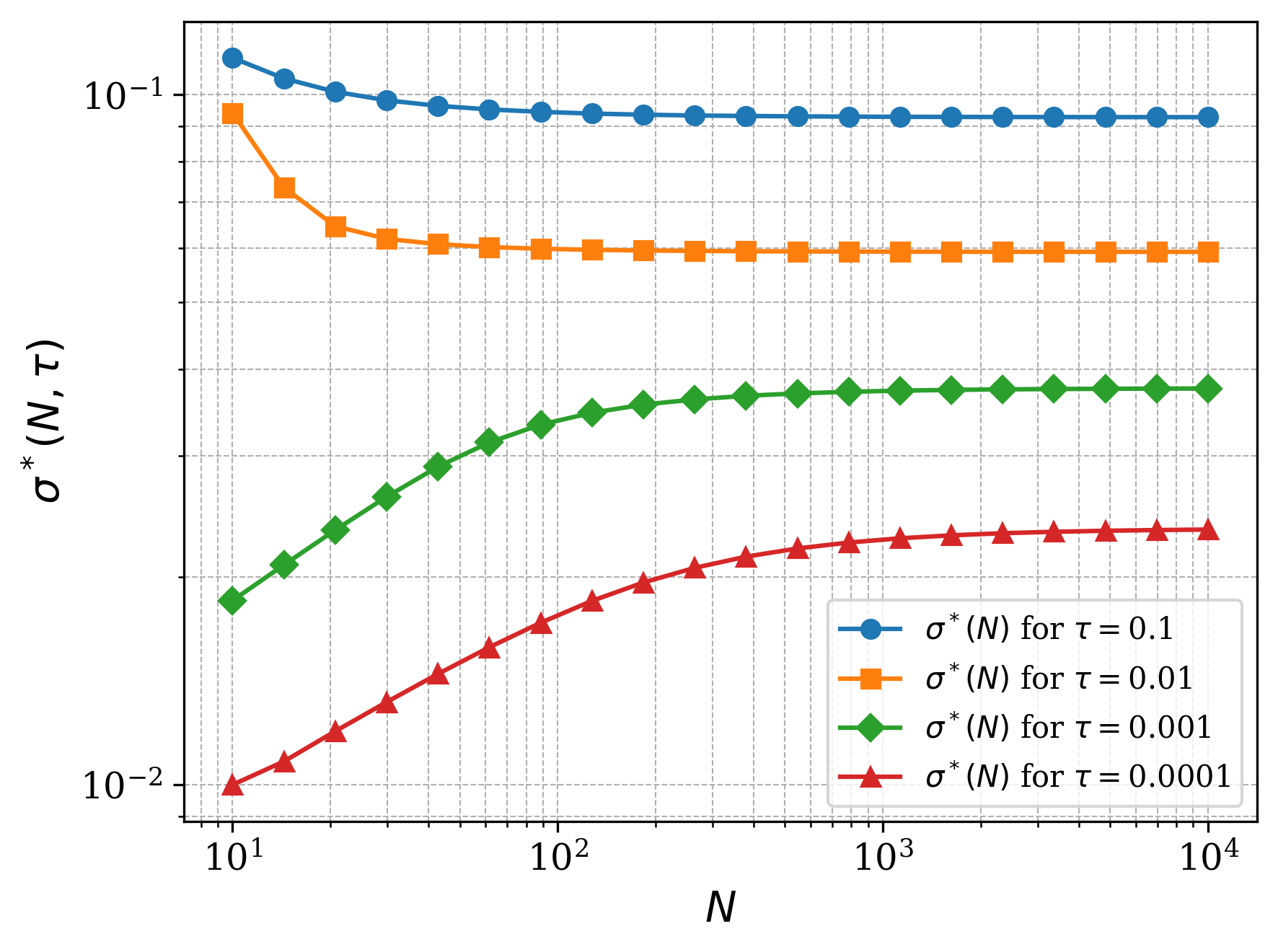}
     \caption{$\sigma^*(N)$ for various $\tau$}
      \label{fig:sigma_opt_N}
    \end{subfigure}
    \begin{subfigure}[b]{0.3\textwidth}
    \centering
    \includegraphics[width=\linewidth]{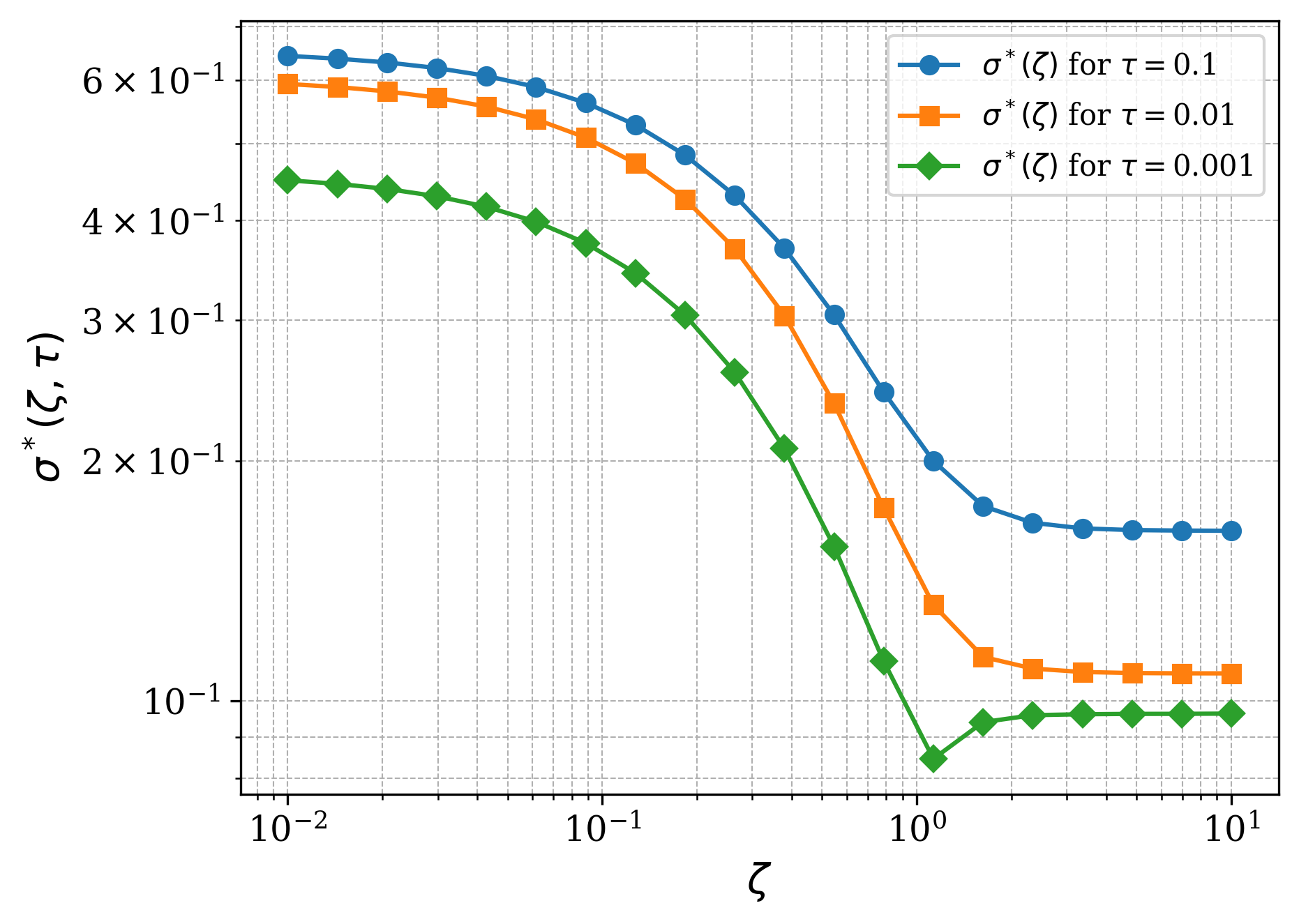}
     \caption{$\sigma^*(\zeta)$ for various $\tau$.}
      \label{fig:sigma_opt_zeta}
    \end{subfigure}
    \caption{Evolution of the optimal noise level parameter $\sigma$ with respect to the SGD parameters~$\tau$ (learning rate) and~$N$ (number of training samples) and the data power law coefficient $\zeta$.}
    \label{fig:analysis_Langevin}
\end{figure}

\paragraph{Numerical calculation of $\sigma^*$}  For different values of $\tau, N, \gamma, \lambda$, we calculate numerically the value of the  $\sigma^*$ that minimizes the theoretical $W_2$ error~\eqref{eq:langevin_W2_bis} (using scipy's $minimize\_scalar$ function, based on Brent's method).  We plot in Figure~\ref{fig:analysis_Langevin} the evolution of $\sigma^*$ with respect to $\tau$ and $\gamma$ (Figure~\ref{fig:sigma_opt_tau}), $N$ (Figure~\ref{fig:sigma_opt_N}) and the power law of the spectrum of $C_{\mathrm{data}}$ (Figure~\ref{fig:sigma_opt_zeta}). We consider, in dimension $d=100$, eigenvalues \( \lambda_i \) decreasing in the power-law $\left(\lambda_i = i^{-\zeta}\right)_{1 \leq i \leq d}$ with coefficient $\zeta>0$. The eigenvalues are clamped to $\lambda_i > 10^{-3}$ for numerical stability. The evolution of the power spectrum for different choices of $\zeta$ is illustrated in Figure~\ref{fig:power_spectrum}. Figures~\ref{fig:sigma_opt_tau} and~\ref{fig:sigma_opt_N} are obtained with $\zeta=1$.

First, we verify in Figure~\ref{fig:sigma_opt_tau} through linear regression that as $\gamma \to 0$, $\sigma^*(\tau)$ approximately scales as~$\tau^{1/6}$. Second, consistent with~\eqref{eq:langevin_W2_sigma_small}, Figure~\ref{fig:sigma_opt_N} shows that for small $\tau$, $\sigma^*$ initially increases with~$N$, but eventually becomes independent of~$N$ as~$N$ grows large.

Finally, we observe in Figure~\ref{fig:sigma_opt_zeta} that the optimal noise parameter decreases as the power-law exponent~$\zeta$ increases in the spectrum defined by $\lambda_i = i^{-\zeta}$. This behavior is explained by Figure~\ref{fig:heatmap}, which shows that $k^\tau_{0,0}(\lambda_i, \lambda_j)$ is dominated by diagonal contributions where $\lambda_i = \lambda_j$ takes large values. When the spectrum decays more slowly (i.e., for smaller $\zeta$), such dominant diagonal terms are more numerous, leading to a larger value of $\sigma^*$ as predicted by the approximation in~\eqref{eq:approx_sigma_star}.


\section{Diffusion model error}

\subsection{Backward diffusion SDE}
\label{app:reverse}
Let \( (x_t)_{t \geq 0} \) be the stochastic process in \( \mathbb{R}^d \) defined as the solution to the Itô stochastic differential equation:
\begin{align*}
    dx_t = -\beta_t(x_t)\,dt + \sqrt{2 \xi_t}\,dw_t,
\end{align*}
where \( \beta_t : \mathbb{R}^d \to \mathbb{R}^d \) is a measurable drift field,
and \( \xi_t > 0 \) is a time-dependent diffusion coefficient. The law \( p_t(x) \) of \( x_t \) admits a smooth density that evolves according to the Fokker-Planck equation:
\begin{align*}
\partial_t p_t(x) = \operatorname{div} \left( \beta_t(x)\, p_t(x) \right) + \xi_t \Delta p_t(x),
\end{align*}
Equivalently, using the identity
\begin{align*}
\Delta p_t(x) = \operatorname{div} \left( \nabla \log p_t(x) \cdot p_t(x) \right),
\end{align*}
this equation can be rewritten in conservative form:
\begin{align*}
\partial_t p_t(x) = -\operatorname{div} \left( f_t(x, p_t)\, p_t(x) \right),
\end{align*}
where the effective velocity field \( f_t \) is given by:
\begin{align*}
f_t(x, p_t) = -\beta_t(x) - \xi_t \nabla \log p_t(x).
\end{align*}
Consider the reverse $q_t = p_{T-t}$, we have 
$ \partial_t q_t = - \partial_t p_{T-t} $ and the following Fokker-Planck equation for~$q_t$
\begin{align*}
\partial_t q_t(x) = -\operatorname{div}(\beta_{T-t}(x)q_t(x)) - \xi_{T-t}\Delta q_t 
\end{align*}
As the diffusion term is negative this is unstable. We can make it positive by using for $\alpha \geq 0$
\begin{align*}
- \Delta q_t = \alpha \Delta q_t - (1+\alpha)\operatorname{div}( q_t \nabla \log q_t) \end{align*}
to get
\begin{align*}\partial_t q_t(x) = -\operatorname{div} \left(\left((1+\alpha)\xi_{T-t}\nabla \log q_t + \beta_{T-t}(x)\right)q_t(x)\right) + \alpha \xi_{T-t} \Delta q_t
\end{align*}
for which the corresponding SDE is 
\begin{align*}dy_t = [(1+\alpha)\xi_{T-t} \nabla \log q_t(y_t) + \beta_{T-t}(y_t)]dt + \sqrt{2\alpha \xi_{T-t}} dw_t.
\end{align*}

\subsection{Proof of Lemma~\ref{prop:diffusion_discretization_1}}
\label{app:diffusion_discretization_1}
\begin{proof}
The Euler-Maruyama discretized diffusion process~\eqref{eq:euler_disc} with linear score 
\begin{align} \label{eq:linear_score_gauss}
v_t(x) = \nabla \log p_t(x) = - \Sigma_t^{-1} x + \Sigma_t^{-1} \mu_t
\end{align}
can be rewritten as
\begin{equation*}
    y_0 \sim q_0, \quad  \text{for} \  k \in \llbracket 0, K-1 \rrbracket \quad 
    \left\{
    \begin{aligned}
        w_k &\sim \mathcal{N}(0, \id) \\
        y_{k+1} &= (1 + \gamma H_{t_k})y_k + \gamma r_{t_k}  + \sqrt{2 \alpha \gamma \xi_{T-t_k}} \cdot w_k
    \end{aligned}
    \right.
\end{equation*}
where we defined for $t \leq T$
\begin{align*} 
H_t &\eqdef \beta_{T-t}\id - (1 + \alpha) \xi_{T-t} \Sigma_{T-t}^{-1}\\
 r_t &\eqdef (1 + \alpha) \xi_{T-t} \Sigma_{T-t}^{-1} \mu_{T-t}.
\end{align*}
We use the following notations, for $t\in [0,T]$ and $k \in \llbracket 0, K \rrbracket$: we define the mean and covariance of the non-discretized process~\eqref{eq:backward_SDE} with linear Gaussian score~\eqref{eq:linear_score_gauss} $m_t \eqdef  \EE[y_t] = \mu_{T-t}$, $C_t \eqdef  \cov[y_t] = \Sigma_{T-t}$. And the corresponding mean and covariance of the discretized process~\eqref{eq:with_Ht}, $\hat m_k = \EE[y_k]$ and  $\hat C_k = \cov[y_k]$. Taking mean and covariance of the two stochastic processes, we get that these means and covariances follow the independent equations: 
  \begin{align*} 
    \dot m_t &= H_t m_t + r_t, \\
    \hat m_{k+1}
    &=
    \hat m_k+\gamma(H_{t_k}\hat m_k+r_{t_k}) \\
    \dot C_t &= H_t C_t + C_t H_t^\top + 2 a_t \id \\
    \hat C_{k+1}
    &=
    (I+\gamma H_{t_k}) \hat C_k(I+\gamma H_{t_k})^\top
    +2\gamma a_{t_k} \id.
  \end{align*}
    Moreover, from~\eqref{eq:H_t}, $H_s$ co-diagonalizes with $\Sigma_{\mathrm{data}} = U \operatorname{Diag}(\lambda_i) U^\top$ and, from~\eqref{eq:r_t},  $r_t$ takes the form $r_t = P_t \mu_{\mathrm{data}}$ with $P_t$ that co-diagonalizes with $\Sigma_{\mathrm{data}}$.
    We thus get, subtracting the continuous and discrete equations, the following decomposition at time $t_k = k \gamma$:
\begin{align*}
\hat m_k-m_{t_k}
  &=  U \operatorname{Diag}\Big( \Lambda_{T- t_k}^d(\lambda_i)\Big) U^\top \mu_{\mathrm{data}},\\
\hat C_k - C_{t_k}
 &= U \operatorname{Diag}\Big( \Lambda_{T- t_k}^D(\lambda_i)\Big) U^\top .
\end{align*}
The vectors $\Lambda_t^d(\lambda_i),\Lambda_t^D(\lambda_i)\in\mathbb R^d$ are first defined on the grid
$\{T-t_k:0\leq k\leq K\}$ by these identities and then extended arbitrarily to
all $t\in[0,T]$. Setting
\[
d_t:=U\operatorname{Diag}(\Lambda_t^d(\lambda_i))U^\top,
\qquad
D_t:=U\operatorname{Diag}(\Lambda_t^D(\lambda_i))U^\top,
\]
gives time-dependent matrices co-diagonalizing with $\Sigma_{\mathrm{data}}$.
\end{proof}

\subsection{Proof of Theorem~\ref{thm:distance_general_diff}}
\label{app:distance_general_diff}

\newtheorem*{repeatthm3}{Theorem 3}
\begin{repeatthm3}[Diffusion sampling error with inexact linear score] 
Under Assumption~\ref{ass:gaussian}, the discretized diffusion process~\eqref{eq:euler_disc} with score~\eqref{eq:diff_inexact_score} samples a Gaussian distribution ${q^\varepsilon_k = \mathcal{N}(\EE[y_{k}], \cov(y_k))}$ that satisfies, 
\begin{equation*}  
\EE_{\delta, \Delta}  \left[ W_2^2(p_\mathrm{data},q_k^\varepsilon) \right] =  \mathcal{E}^{(0)}_{T-t_k} + \gamma \mathcal{E}^{\text{disc}}_{T-t_k} + \varepsilon^2 \gamma \left( \mathcal{E}_{T-t_k}^{\text{mean}}(\delta) + \mathcal{E}_{T-t_k}^{\text{cov}}(\Delta) \right) + o(\gamma, \varepsilon^2)
\end{equation*}
where, for $r = T-t \in [0,T]$,
\begin{align*}
\mathcal{E}^{(0)}_{r} &=  \sum_{i=1}^d \left(\sqrt{\lambda_i} - \sqrt{\lambda^r_i}\right)^2 +  (1- s_{r})^2\norm{\mu}^2, \\
\mathcal{E}^{\text{disc}}_{r} &=  \sum_{i=1}^d \Lambda^{D}_{r}(\lambda_i) \left( 1 - \sqrt{\frac{\lambda_i}{\lambda^{r}_i}} \right)  + \sum_{i=1}^d 2 (s_{r}-1) \Lambda^{d}_{r}(\lambda_i)(U^\top \mu)^2_i\\
\mathcal{E}_{r}^{\text{mean}}(\delta)  &= \sum_{i=1}^d \int_{r}^{T} h_{r,s}^2 \left(\frac{\lambda_i^{r}}{\lambda_i^{s}}\right)^{\alpha+1} \EE_{\delta} \left[\langle \cov \delta_{s}, u_i u_i^\top \rangle\right]ds \\
\mathcal{E}_{r}^{\text{cov}}(\Delta)  &=  \sum_{i,j=1}^d  \int_{r}^{T} \! \!\left( \kappa^{(1)}_{r,s}(\lambda_i, \lambda_j)  \EE_{\Delta}\!\left[\langle \Delta_s, u_iu_j^\top \rangle^2\right]  + \!\kappa^{(2)}_{r,s}(\lambda_i, \lambda_j) \, \EE_{\Delta}\!\left[\langle \Delta_s, u_iu_j^\top \rangle \langle \Delta_s, u_ju_i^\top \rangle \right] \right) ds
\end{align*}
where we denoted $\left(\lambda^r_i = s_r^2 (\lambda_i + \sigma_r^2)\right)_i$; the eigenvalues of $\Sigma_r$, $\Lambda^{d}_{r}(\lambda_i)$ the eigenvalues of~$d_r$ and $\Lambda^{D}_{r}(\lambda_i)$ the eigenvalues of~$D_r$ (with $d_r$ and $D_r$ defined in Lemma~\ref{prop:diffusion_discretization_1}), 
and $h_{r,s} \eqdef (1+\alpha) e^{\alpha (B_{r}-B_s)} \xi_{s}$.
Finally,  \( \kappa^{(1)}_{r,s}, \kappa^{(2)}_{r,s} : \mathbb{R}^2 \to \mathbb{R} \) are kernel functions on the eigenvalues \( \lambda_i \) of \( C_\mathrm{data} \). Their full expression are given equations~\eqref{eq:kappa1_diff} and~\eqref{eq:kappa2_diff}.
\end{repeatthm3} 

\begin{proof}

In order to prove this result, we start by deriving the expansion in $\varepsilon$ of the Gaussian distribution sampled by the diffusion process at step $k$. Then, we calculate the $W_2$ error based on this expansion. 

\subsubsection{Explicit solution of the diffusion~\eqref{eq:euler_disc} with inexact score~\eqref{eq:diff_inexact_score}.}

We study the Euler-Maruyama discretized diffusion process~\eqref{eq:euler_disc} with linear score $v_t(x) = \nabla \log p_t(x) = - \Sigma_t^{-1} x + \Sigma_t^{-1} \mu_t$. The exact-score algorithm~\eqref{eq:euler_disc} can be rewritten with this score as
\begin{equation} \label{eq:with_Ht}
    y_0 \sim q_0, \quad  \text{for} \  k \in \llbracket 0, K-1 \rrbracket \quad 
    \left\{
    \begin{aligned}
        w_k &\sim \mathcal{N}(0, \id) \\
        y_{k+1} &= (1 + \gamma H_{t_k})y_k + \gamma r_{t_k}  + \sqrt{2 \alpha \gamma \xi_{T-t_k}} \cdot w_k
    \end{aligned}
    \right.
\end{equation}
where we defined for $t \leq T$
\begin{align} 
H_t &\eqdef \beta_{T-t}\id - (1 + \alpha) \xi_{T-t} \Sigma_{T-t}^{-1} \label{eq:H_t}\\
 r_t &\eqdef (1 + \alpha) \xi_{T-t} \Sigma_{T-t}^{-1} \mu_{T-t}. \label{eq:r_t}
\end{align}

For inexact score, $H_t$ and $r_t$ are replaced by the random quantities:
\begin{align*}
    r^\varepsilon_{t} 
    &= r_{t} + \varepsilon (1+\alpha)\xi_{T-{t}} (\delta_{T-t} + \Delta_{T-t}\mu_{T-{t}}) \\
    H^\varepsilon_{t_k} 
    & = H_{t} - \varepsilon (1 + \alpha) \xi_{T-t}\Delta_{T-t}
\end{align*}
such that the discretized diffusion process~\eqref{eq:euler_disc} with score~\eqref{eq:diff_inexact_score} can be rewritten as
\begin{equation} \label{eq:euler_disc_linear_pert_app}
    y_0 \sim p_T, \quad  \text{for} \  k \in \llbracket 1, K-1 \rrbracket \quad 
    \left\{
    \begin{aligned}
        w_k &\sim \mathcal{N}(0, \id) \\
        y_{k+1} &= (1 + \gamma H^{\varepsilon}_{t_k})y_k + \gamma r^{\varepsilon}_{t_k}  + \sqrt{2 \alpha \gamma \xi_{T-t_k}} \cdot w_k
    \end{aligned}
    \right.
\end{equation}

First, with the following Proposition, we start by deriving the second-order expansion, in $\varepsilon$, of the mean and covariance of the solution of this process.
\begin{prop}
\label{prop:pertubed_diffusion}
Under the Gaussian assumption~\ref{ass:gaussian}, the stochastic process~\eqref{eq:euler_disc_linear_pert_app} has solution $y_k$ which follows a Gaussian distribution $\hat q^\varepsilon_k = \mathcal{N}(\hat \mu^\varepsilon_k, \hat \Sigma^\varepsilon_k)$, with mean and covariance expanding as
\begin{align*}
\hat \mu_k^\varepsilon &= \EE[y_{k}] = \mu_{k}^{(0)} + \varepsilon \mu_k^{(1)}(\delta) + o(\varepsilon^2)  \\
\hat\Sigma_k^\varepsilon &= \cov[y_{k}] = \Sigma_{k}^{(0)} + \varepsilon \Sigma_k^{(1)}(\Delta) + \varepsilon^2 \Sigma_k^{(2)}(\Delta) + o(\varepsilon^2) .
\end{align*}
Denoting $r_k \eqdef T-t_k$, and defining for $0 \leq t,s \leq T$
\begin{align*}
h_{t,s} &\eqdef (1+\alpha) e^{\alpha (B_{t}-B_s)} \xi_{s} \\
F_{t,s} &\eqdef (\Sigma_{t} \Sigma_{s}^{-1})^{\frac{\alpha+1}{2}} \quad \text{and} \quad \hat F^\gamma_{t,s} \eqdef F_{t,s} + \mathcal{O}(\gamma) \\
E_{t,s} &\eqdef \Sigma_{t} \left(\Sigma_{t}\Sigma_{s}^{-1}\right)^{(\alpha-1)/2} \quad \text{and} \quad \hat E^\gamma_{t,s} \eqdef E_{t,s} + \mathcal{O}(\gamma),
\end{align*}
The different terms in the expansions are given by 
\begin{empheq}[left=\empheqlbrace]{align}
    \mu_{k}^{(0)} &= \mu_{r_k} + \gamma d_{r_k} \cdot \mu + \mathcal{O}(\gamma^2) \\
 \mu_k^{(1)}(\delta) &=  \sum_{\ell=0}^{k-1} \gamma \, h_{r_k,r_\ell} \hat F^\gamma_{r_k, r_\ell} (\delta_{r_\ell} - \gamma \Delta_{r_\ell}d_{r_k} \cdot \mu) + \mathcal{O}(\gamma^2)  
\end{empheq} 
and
\begin{empheq}[left=\empheqlbrace]{align}
  \Sigma_{k}^{(0)} &= \Sigma_{r_k} + \gamma D_{r_k} + \mathcal{O}(\gamma^2) \\
 \Sigma_{k}^{(1)}(\Delta) &= - \sum_{\ell=0}^{k-1} \gamma \, h_{r_k,r_\ell} e^{\alpha (B_{r_k}-B_{r_\ell})} \left( \hat F^\gamma_{r_k,r_\ell} \Delta_{r_\ell} \hat E^\gamma_{r_k,r_\ell} +   \hat E^\gamma_{r_k,r_\ell} \Delta_{r_\ell}^\top  \hat F^\gamma_{r_k,r_\ell} \right)+ \mathcal{O}(\gamma^2)\\
\Sigma_{k}^{(2)}(\Delta) &=  \sum_{\ell=0}^{k-1} \sum_{i=0}^{k-1} \gamma^2 \,
 h_{r_k,r_{\ell}} h_{r_k,r_{i}} \cdot \\ \nonumber
 &\Biggl( e^{\alpha(B_{r_k}- B_{r_i})} \left(\hat F^\gamma_{r_k,s_{\ell}} \Delta_{r_\ell} \hat F^\gamma_{r_\ell,r_i} \Delta_{r_i} \hat E^\gamma_{r_k,r_i}
 + \hat E^\gamma_{r_k,r_i} \Delta_{r_i}^\top \hat F^\gamma_{r_\ell,r_i} \Delta_{r_\ell}^\top \hat F^\gamma_{r_k,r_\ell} \right)  \\ \nonumber  &+  e^{\alpha(B_{r_i}- B_{r_\ell})} \hat F^\gamma_{r_k,r_\ell} \Delta_{r_\ell} \hat E^\gamma_{r_\ell, r_i} \Delta_{r_i}^\top \hat F^\gamma_{r_k,r_i} \Biggl) + \mathcal{O}(\gamma^2)
\end{empheq} 

\end{prop}
\begin{proof}
In the proof, we use the following lemma, which makes explicit the first-order error when discretizing a general Riemann integral with the left-endpoint rectangle method. 

\begin{lem}[Left-endpoint rectangle discretization error for vector-valued integrals] \label{lem:disc}
Let \( {f \in \mathcal{C}^2([a,b]; \mathbb{R}^m)} \) and \( \{ t_i \}_{i=\ell}^k \) be a time grid, where \(a < t_\ell < t_{\ell+1} < \cdots < t_k < b\). Define \( {\gamma_i := t_{i+1} - t_i } \), and \( {\gamma := \max_{i \in \llbracket \ell, k-1 \rrbracket} \gamma_i} \). Then:
\begin{align*}
\sum_{i=\ell}^{k-1} \gamma_i f(t_i) = \int_{t_\ell}^{t_k} f(s) \, ds - \sum_{i=\ell}^{k-1} \frac{\gamma_i^2}{2} f'(t_i) + O\left(\gamma^2 \right),
\end{align*}

In the case of a uniform step size \( \gamma_i = \gamma \) for all \( i \in \llbracket \ell, k-1 \rrbracket \), we have:
\begin{align*}
\gamma \sum_{i=\ell}^{k-1} f(t_i) = \int_{t_\ell}^{t_k} f(s) \, ds - \frac{\gamma}{2} \left( f(t_k) - f(t_\ell) \right) + O\left( \gamma^2 \right).
\end{align*}
\end{lem}

\begin{proof}
We expand the sum using the fundamental theorem of calculus:
\begin{align*}
\sum_{i=\ell}^{k-1} \gamma_i f(t_i) = \sum_{i=\ell}^{k-1} \int_{t_i}^{t_{i+1}} f(s) \, ds + \sum_{i=\ell}^{k-1} \int_{t_i}^{t_{i+1}} \left( f(t_i) - f(s) \right) \, ds.
\end{align*}
Apply a first-order Taylor expansion of \( f \) around \( t_i \), componentwise:
\begin{align*}
f(s) = f(t_i) + (s - t_i) f'(t_i) + O((s - t_i)^2),
\end{align*}
so that:
\begin{align*}
\int_{t_i}^{t_{i+1}} \left( f(s) - f(t_i) \right) \, ds 
&= \int_{t_i}^{t_{i+1}} (s - t_i) f'(t_i) \, ds + O(\gamma_i^3) \\
&= \frac{\gamma_i^2}{2} f'(t_i) + O(\gamma_i^3) .
\end{align*}
Summing over \( i \) gives:
\begin{align*}
\sum_{i=\ell}^{k-1} \gamma_i f(t_i) 
= \int_{t_\ell}^{t_k} f(s) \, ds - \sum_{i=\ell}^{k-1} \frac{\gamma_i^2}{2} f'(t_i) + O\left( \sum_{i=\ell}^{k-1} \gamma_i^3 \right).
\end{align*}
Since \( \sum_{i=\ell}^{k-1} \gamma_i^3 \leq \gamma^2 \sum_{i=\ell}^{k-1} \gamma_i = \gamma^2 (t_k - t_\ell) \), we conclude:
\begin{align*}
\sum_{i=\ell}^{k-1} \gamma_i f(t_i) 
= \int_{t_\ell}^{t_k} f(s) \, ds - \sum_{i=\ell}^{k-1} \frac{\gamma_i^2}{2} f'(t_i) + O\left( \gamma^2 \right).
\end{align*}

For the uniform step size case \( \gamma_i = \gamma \), apply the same result to \( f' \) and observe:
\begin{align*}
\sum_{i=\ell}^{k-1} f'(t_i) = \frac{1}{\gamma} \sum_{i=\ell}^{k-1} (f(t_{i+1}) - f(t_i)) = \frac{1}{\gamma} (f(t_k) - f(t_\ell)).
\end{align*}
So:
\begin{align*}
\sum_{i=\ell}^{k-1} \frac{\gamma^2}{2} f'(t_i) = \frac{\gamma}{2}(f(t_k) - f(t_\ell)),
\end{align*}
which gives:
\begin{align*}
\gamma \sum_{i=\ell}^{k-1} f(t_i) 
= \int_{t_\ell}^{t_k} f(s) \, ds - \frac{\gamma}{2}(f(t_k) - f(t_\ell)) + O\left( \gamma^2 \right).
\end{align*}
\end{proof}
We are now ready for the proof of Proposition~\ref{prop:pertubed_diffusion}. We depart from a second-order expansion in $\varepsilon$ of the solution at step $k$
\begin{equation*}
    y_k = y^{(0)}_k + \varepsilon y^{(1)}_k + \varepsilon^2 y^{(2)}_k + o(\varepsilon^2)
\end{equation*}
and write the SDE verified by each term:
\begin{empheq}[left=\empheqlbrace]{align}
        y^{(0)}_0 = p_T, \quad &y_{k+1}^{(0)} = y_k^{(0)} + \gamma (H_{t_k} y^{(0)}_k + r_{t_k}) + \sqrt{2\alpha \gamma \xi_{T-t_k}} w_k \label{eq:y0_k}\\ \label{eq:y1_k}
        y^{(1)}_0 = 0, \quad &y^{(1)}_{k+1} = y^{(1)}_k \\ &+ \nonumber \gamma \left(H_{t_k} y^{(1)}_k 
        - (1 + \alpha) \xi_{T-t_k} \left(\Delta_{T-t_k} y^{(0)}_k - (\delta_{T-t_k} + \Delta_{T-t_k}\mu_{T-t_k}) \right) \right) \\
        y^{(2)}_0 = 0, \quad  &y^{(2)}_{k+1} = y^{(2)}_{k} + \gamma \left(H_{t_k} y^{(2)}_k - (1 + \alpha) \xi_{T-t_k}\Delta_{T-t_k} y^{(1)}_k\right) \label{eq:y2_k} .
\end{empheq}
Then we solve sequentially the equations \eqref{eq:y0_k}, \eqref{eq:y1_k} and \eqref{eq:y2_k}. For the former, $y^{(0)}_k$ is given by the non-perturbed solution given in Lemma~\ref{prop:diffusion_discretization_1}. For solving $y^{(1)}_k$ and $y^{(2)}_k$, we make explicit the accumulation of iterates. 
\paragraph{Order \(\varepsilon^1\):} For the first-order term $y^{(1)}_t$:
\begin{align*}
y^{(1)}_{k+1} 
&= (\id + \gamma H_{t_k}) y^{(1)}_k + \gamma g_{t_k}
\end{align*}
where 
\begin{align*}
g_{t_k} 
&\eqdef (1 + \alpha) \xi_{T-t_k}( - \Delta_{T-t_k} (y^{(0)}_k - \mu_{T-t_k}) +  \delta_{T-t_k})
\end{align*}
and we remind that 
\begin{align*}
     H_{t_k} = \beta_{T-t_k}\id - (1 + \alpha) \xi_{T-t_k} \Sigma_{T-t_k}^{-1}.
\end{align*}
Using  $P_k = \id + H_{t_k} \gamma$, and unrolling the recursion, we obtain:
\begin{align*}
    y^{(1)}_k &= \Psi_{k,0} y^{(1)}_0 + \gamma \sum_{\ell=0}^{k-1} \Psi_{k,\ell} g_{t_\ell} = \gamma \sum_{\ell=0}^{k-1} \Psi_{k,\ell} g_{t_\ell}
\end{align*}
where 
\begin{align*}
\Psi_{k,\ell} := \prod_{i=\ell}^k P_i = \prod_{i=\ell}^k (\id + \gamma H_{t_i}) =  e^{\gamma \sum_{i=\ell}^k H_{t_i}} + \mathcal{O}(\gamma^2) 
\end{align*}
We now expand $\Psi_{k,\ell}$ at first-order in $\gamma$.
\begin{lem}
$\Psi_{k,\ell}$ is the discretization at time $t = t_k$ and $s = t_\ell$ of 
\begin{align*}
\hat \Psi_{t,s} &= e^{\alpha (Q_{s} - Q_{t})} \left( \Sigma_{T-t} \Sigma_{T-s}^{-1} \right)^{\frac{\alpha + 1}{2}}(\id - \frac{\gamma}{2}(H_{t} - H_{s})) + \mathcal{O}(\gamma^2) .
\end{align*}
\end{lem}
\begin{proof}
    The term $\gamma \sum_{i=\ell}^k H_{t_i}$ corresponds to the left-endpoint rectangle discretization, with intervals $\gamma$,  of the integral $\int_{t_\ell}^{t_k} H_s ds$. It satisfies, by Lemma~\ref{lem:disc},
\begin{align*}
\gamma \sum_{i=\ell}^k H_{t_i} = \int_{t_\ell}^{t_k} H_s ds - \frac{\gamma}{2}(H_{t_k} - H_{t_\ell}) + \mathcal{O}(\gamma^2) 
.
\end{align*}
Thus, by commutativity between the $(H_{t_i})_i$, and first-order expansion of the exponential,
\begin{align*}
\Psi_{k,\ell} &= e^{\int_{t_\ell}^{t_k} H_s ds}e^{- \frac{\gamma}{2}(H_{t_k} - H_{t_\ell}) + \mathcal{O}(\gamma^2)} + \mathcal{O}(\gamma^2)\\ 
&= e^{\int_{t_\ell}^{t_k} H_s ds}(\id - \frac{\gamma}{2}(H_{t_k} - H_{t_\ell})) + \mathcal{O}(\gamma^2) \\
&= e^{\int_{t_\ell}^{t_k} H_s ds} +  \frac{\gamma}{2} e^{\int_{t_\ell}^{t_k} H_s ds}(H_{t_\ell}-H_{t_k}) + \mathcal{O}(\gamma^2).
\end{align*}
To calculate the integral $\int_{t_\ell}^{t_k} H_s ds$ explicitly, we saw in Section~\ref{sec:background} that with Gaussian data, the distribution $p_t \sim \mathcal{N} (\mu_t, \Sigma_t)$, with
\begin{align*}
    \Sigma_t = s_t^2(C_{\mathrm{data}} + \sigma_t^2\id) = e^{-2 B_t} \left( C_{\mathrm{data}} +  \left(2 \int_0^t \xi_s e^{2B_s} \, ds \right) \id \right)
\end{align*}
where $B_t = \int_0^t \beta_s ds$. In particular $\Sigma_t$ satisfies 
\begin{align*}
\frac{d}{dt}\Sigma_t = - 2\beta_{t} \Sigma_t + 2\xi_t \id.
\end{align*}
Using the property
\begin{align*}
d[\Sigma_t^{-p}] = - p  \Sigma_t^{-p} d[\Sigma_t] \Sigma_t^{-1}
\end{align*}
we get 
\begin{align*}
\frac{d}{dt}[\Sigma_{T-t}^{-\frac{\alpha + 1}{2}}] &= \frac{\alpha + 1}{2} \Sigma_{T-t}^{-\frac{\alpha + 1}{2}} \frac{d}{dt}[\Sigma_{T-t}]  \Sigma_{T-t}^{-1} \\
& =  (\alpha + 1) \Sigma_{T-t}^{-\frac{\alpha + 1}{2}}(- \beta_{T-t} \Sigma_{T-t} + \xi_{T-t} \id)  \Sigma_{T-t}^{-1} \\
&= (\alpha + 1) \Sigma_{T-t}^{-\frac{\alpha + 1}{2}}( - \beta_{T-t} \id + \xi_{T-t} \Sigma_{T-t}^{-1}).
\end{align*}
That is to say
\begin{align*}
\Sigma_{T-t}^{\frac{\alpha + 1}{2}} \frac{d}{dt} \left[\Sigma_{T-t}^{-\frac{\alpha + 1}{2}}\right] &= - (1+\alpha) \beta_{T-t} \id + (1+\alpha) \xi_{T-t} \Sigma_{T-t}^{-1} = - H_t - \alpha  \beta_{T-t} \id, 
\end{align*}
or equivalently, using the log of a positive definite matrix, 
\begin{align*}
 H_t &= -\frac{d}{dt} \log \Sigma_{T-t}^{-\frac{\alpha + 1}{2}}  + \alpha \frac{d}{dt} B_{T-t} \\
 &= \frac{d}{dt}  \log \left( e^{\alpha B_{T-t}} \Sigma_{T-t}^{\frac{\alpha + 1}{2}} \right).
\end{align*}
We can then calculate the integral $\int_{t_\ell}^{t_k} H_s ds$ explicitly. Passing to the matrix exponential, and denoting $Q_t \eqdef B_{T} - B_{T-t}$,
\begin{align*}
e^{\int_{t_\ell}^{t_k} H_s ds} &= e^{\alpha (B_{T-t_k} - B_{T-t_\ell})} \Sigma_{T-t_k}^{\frac{\alpha + 1}{2}} \Sigma_{T-t_\ell}^{-\frac{\alpha + 1}{2}} \\
&= e^{\alpha (Q_{t_\ell} - Q_{t_k})} \left( \Sigma_{T-t_k} \Sigma_{T-t_\ell}^{-1} \right)^{\frac{\alpha + 1}{2}}.
\end{align*}
We thus get that $\Psi_{k,\ell}$ is the discretization at time $t = t_k$ and $s = t_\ell$ of 
\begin{align*}
\hat \Psi_{t,s} &= e^{\alpha (Q_{s} - Q_{t})} \left( \Sigma_{T-t} \Sigma_{T-s}^{-1} \right)^{\frac{\alpha + 1}{2}}(\id - \frac{\gamma}{2}(H_{t} - H_{s})) + \mathcal{O}(\gamma^2) .
\end{align*}
\end{proof}
With this lemma, we get:
\begin{align*}
    y^{(1)}_k &= \gamma \sum_{\ell=0}^{k-1} e^{\alpha (Q_{t_\ell} - Q_{t_k})} \left( \Sigma_{T-t_k} \Sigma_{T-t_\ell}^{-1} \right)^{\frac{\alpha+1}{2}} \left(\id - \frac{\gamma}{2}(H_{t_{k}} - H_{t_\ell})\right)g_{t_\ell} + O(k\gamma^3) \\
    &= \gamma \sum_{\ell=0}^{k-1} K_{t_k, t_\ell}\left(\id - \frac{\gamma}{2}(H_{t_{k}} - H_{t_\ell})\right) \left( \delta_{T-t_\ell} - \Delta_{T-t_\ell} (y^{(0)}_\ell - \mu_{T-t_\ell}) \right) + \mathcal{O}(\gamma^2)
\end{align*}
where we denoted, for $0 \leq t,s \leq T$,
\begin{align*}
K_{t,s}&= h_{T-t, T-s} F_{T-t, T-s}
\end{align*}
with 
\begin{align*}
h_{t,s} =  (1+\alpha) \xi_{s}e^{ \alpha (B_t-B_s)} \quad \text{and} \quad F_{t,s} \eqdef  ( \Sigma_{t} \Sigma_{s}^{-1})^{\frac{\alpha+1}{2}}.
\end{align*}
We also define
\begin{align*}
\hat K^\gamma_{t,s} \eqdef K_{t,s}\left(\id - \frac{\gamma}{2}(H_{t} - H_{s})\right)
\end{align*}
to write
\begin{align*}
y^{(1)}_k &= \gamma \sum_{\ell=0}^{k-1} \hat K^\gamma_{t_k, t_\ell} \left( \delta_{T-t_\ell} - \Delta_{T-t_\ell} (y^{(0)}_\ell - \mu_{T-t_\ell}) \right) + \mathcal{O}(\gamma^2).
\end{align*}
For convenience, we also define $\hat F^\gamma_{t,s} \eqdef  F^\gamma_{t,s}\left(\id - \frac{\gamma}{2}(H_{t} - H_{s})\right)$ in order to have also
\begin{align*}
\hat K^\gamma_{t,s}&= h_{T-t, T-s} \hat F^\gamma _{T-t, T-s}.
\end{align*}
The mean of $y^{(1)}_k$ then satisfies
\begin{align*}
\EE\left[y^{(1)}_k\right] &=  \gamma \sum_{\ell=0}^{k-1} \hat K^\gamma_{t_k, t_\ell} \left( \delta_{T-t_\ell} - \Delta_{T-t_\ell} (\EE\left[y^{(0)}_\ell\right] - \mu_{T-t_\ell}) \right) + \mathcal{O}(\gamma^2) .
\end{align*}
Using Lemma~\ref{prop:diffusion_discretization_1}, assuming perfect mean initialization $\EE[y_0] = \mu_T$, 
\begin{align*}
\EE\left[y^{(0)}_\ell\right] = \mu_{T-t_\ell} + \gamma d_{T-t_\ell} \cdot \mu  + \mathcal{O}(\gamma^2)
\end{align*}
such that, at the first-order in $\gamma$,
\begin{align*}
\EE \left[y^{(1)}_\ell\right] &=  \gamma \sum_{\ell=0}^{k-1}\hat K^\gamma_{t_k, t_\ell} \left( \delta_{T-t_\ell} - \gamma \Delta_{T-t_\ell} d_{T-t_\ell} \cdot \mu \right) + \mathcal{O}(\gamma^2) \\
&=  \gamma \sum_{\ell=0}^{k-1} K_{t_k, t_\ell} \delta_{T-t_\ell}  \\
&+ \nonumber \gamma^2 \sum_{\ell=0}^{k-1} K_{t_k, t_\ell} \left( (H_{t_\ell}- H_{t_k})\delta_{T-t_\ell} -  \Delta_{T-t_\ell} d_{T-t_\ell} \cdot \mu \right)
+ \mathcal{O}(\gamma^2).
\end{align*}

\paragraph{Order \(\varepsilon^2\):} The second-order term $y^{(2)}_k$ satisfies
\begin{align*}
   y^{(2)}_0 &= 0, \quad  y^{(2)}_{k+1} = y^{(2)}_{k} + \gamma \left(H_{t_k} y^{(2)}_k - (1 + \alpha) \xi_{T-t_k}\Delta_{T-t_k} y^{(1)}_k\right).
\end{align*}
With the same unrolling than for $y^{(1)}$, we get
\begin{align*}
    y^{(2)}_k &= - \gamma \sum_{\ell=0}^{k-1} \Psi_{k,\ell} \left((1 + \alpha) \xi_{T-t_\ell}\Delta_{T-t_\ell} y^{(1)}_\ell \right) \\
    &= - \gamma \sum_{\ell=0}^{k-1} \hat K^\gamma_{t_k,t_\ell} \Delta_{T-t_\ell} y^{(1)}_\ell  + \mathcal{O}(\gamma^2).
\end{align*}
\paragraph{Covariance of $y_k$.} The second-order decomposition of the covariance is
\begin{align*}
\cov(y_k) &= \cov(y^{(0)}_k) + \varepsilon \left( \cov(y^{(0)}_k, y^{(1)}_k) + \cov(y^{(1)}_k, y^{(0)}_k) \right) \\
&+ \varepsilon^2 \left( \cov(y^{(0)}_k, y^{(2)}_k)  + \cov(y^{(2)}_k, y^{(0)}_k) + \cov(y^{(1)}_k) \right)  + o(\varepsilon^2).
\end{align*}
We calculate each term individually. 
\paragraph{First $\cov(y^{(0)}_k, y^{(1)}_k)$:}
We have
\begin{align} \label{eq:y_1-mean}
y^{(1)}_k - \EE[y^{(1)}_k] &= - \gamma \sum_{\ell=0}^{k-1} \hat K^\gamma _{t_k,t_\ell} \Delta_{T-t_\ell} \left(y^{(0)}_\ell - \EE[y^{(0)}_\ell] \right) + \mathcal{O}(\gamma^2)
\end{align}
such that
\begin{align*}
\cov(y^{(0)}_k, y^{(1)}_k) &= \EE\left[(y^{(0)}_k- \EE[y^{(0)}_k])(y^{(1)}_k- \EE[y^{(1)}_k])^\top \right]  \\ &= - \EE\left[ \left(y^{(0)}_k - \EE[y^{(0)}_k] \right) \gamma \sum_{\ell=0}^{k-1}  \left(y^{(0)}_\ell - \EE[y^{(0)}_\ell] \right)^\top \Delta_{T-t_\ell}^\top  \hat K^\gamma_{t_k,t_\ell}  \right] + \mathcal{O}(\gamma^2) \\
&= - \gamma \sum_{\ell=0}^{k-1} \cov(y^{(0)}_k, y^{(0)}_\ell)  \Delta_{T-t_\ell}^\top  \hat K^\gamma_{t_k,t_\ell} + \mathcal{O}(\gamma^2).
\end{align*}
\paragraph{Second $\cov(y^{(0)}_k, y^{(2)}_k)$:}
We have
\begin{align*} 
y^{(2)}_k - \EE[y^{(2)}_k] &= - \gamma \sum_{\ell=0}^{k-1} \hat K^\gamma _{t_k,t_\ell} \Delta_{T-t_\ell} \left(y^{(1)}_\ell - \EE[y^{(1)}_\ell] \right) + \mathcal{O}(\gamma^2)
\end{align*}
such that
\begin{align*}
\cov(y^{(0)}_k, y^{(2)}_k) &= \EE\left[(y^{(0)}_k- \EE[y^{(0)}_k])(y^{(2)}_k- \EE[y^{(2)}_k])^\top \right]  \\
&= - \EE\left[ \left(y^{(0)}_k - \EE[y^{(0)}_k] \right) \gamma \sum_{\ell=0}^{k-1} \left(y^{(1)}_{t_\ell} - \EE[y^{(1)}_{t_\ell}] \right)^\top \Delta_{T-t_\ell}^\top  \hat K^\gamma_{t_k,t_\ell}  \right] + \mathcal{O}(\gamma^2) \\
&=  - \gamma \sum_{\ell=0}^{k-1} \cov(y^{(0)}_k, y^{(1)}_\ell)  \Delta_{T-t_\ell}^\top  \hat K^\gamma_{t_k,t_\ell} + \mathcal{O}(\gamma^2)
\end{align*}
From the expression~\eqref{eq:y_1-mean} of $y^{(1)}_\ell$, we have 
\begin{align*}
\cov(y^{(0)}_k, y^{(1)}_\ell) &=  - \gamma \sum_{i=0}^{\ell-1}  \cov(y^{(0)}_k, y^{(0)}_i) \Delta_{T-t_i}^\top \hat K^\gamma_{t_\ell,t_i} + \mathcal{O}(\gamma^2)
\end{align*}
and thus
\begin{align*}
\cov(y^{(0)}_k, y^{(2)}_k) &= \gamma^2 \sum_{\ell=0}^{k-1} \sum_{i=0}^{\ell-1}\cov(y^{(0)}_k, y^{(0)}_i)  \Delta_{T-t_i}^\top \hat K^\gamma_{t_\ell,t_i} \Delta_{T-t_\ell}^\top \hat K^\gamma_{t_k,t_\ell} + \mathcal{O}(\gamma^2).
\end{align*}
\paragraph{Third $\cov(y^{(1)}_k)$:}
\begin{align*}
\cov(y^{(1)}_k) &= \EE\left[(y^{(1)}_k- \EE[y^{(1)}_k])(y^{(1)}_k- \EE[y^{(1)}_k])^\top \right] \\
&= \gamma^2 \sum_{\ell=0}^{k-1} \sum_{i=0}^{k-1} \hat K^\gamma_{t_k,s_1} \Delta_{T-t_\ell}   \cov(y^{(0)}_{t_\ell}, y^{(0)}_{t_i})  \Delta_{T-t_i}^\top  \hat K^\gamma_{t_k,t_i} + \mathcal{O}(\gamma^2)
\end{align*}
All terms considered, we get
\begin{align} \label{eq:cov_entire}
\cov(y_k)  &=  \cov(y^{(0)}_k) + \varepsilon \Sigma^{(1)}_k(\Delta) +  \varepsilon^2 \Sigma^{(2)}_k(\Delta) + o(\varepsilon^2) 
\end{align}
where 
\begin{align*} 
\Sigma^{(1)}_k(\Delta) &= - \gamma \sum_{\ell=0}^{k-1} \left( \hat K^\gamma_{t_k,t_\ell} \Delta_{T-t_\ell} \cov(y^{(0)}_k, y^{(0)}_\ell) + \cov(y^{(0)}_k, y^{(0)}_\ell)  \Delta_{T-t_\ell}^\top  \hat K^\gamma_{t_k,t_\ell} \right) + \mathcal{O}(\gamma^2)
\end{align*}
\begin{equation*} 
\begin{split}
\Sigma^{(2)}_k(\Delta) =  \gamma^2 \sum_{\ell=0}^{k-1} \sum_{i=0}^{j-1} \Biggl( &\cov(y^{(0)}_k, y^{(0)}_i)  \Delta_{T-t_i}^\top \hat K^\gamma_{t_\ell,t_i} \Delta_{T-t_\ell}^\top \hat K^\gamma_{t_k,t_\ell} \\
&+ \hat K^\gamma_{t_k,t_\ell} \Delta_{T-t_\ell}  \hat K^\gamma_{t_k,t_i}  \Delta_{T-t_i}  \cov(y^{(0)}_{t_\ell}, y^{(0)}_{t_i})  \\
&+  \hat K^\gamma_{t_k,t_\ell} \Delta_{T-t_\ell} \cov(y^{(0)}_k, y^{(0)}_i) \Delta_{T-t_i}^\top  \hat K^\gamma_{t_\ell,t_i}  \Biggr)  + \mathcal{O}(\gamma^2)
\end{split}
\end{equation*}
The above expression depends on $\cov(y^{(0)}_k, y^{(0)}_\ell)$. We now turn to its derivation. 

\paragraph{Calculation of $\cov(y^{(0)}_{k}, y^{(0)}_{\ell})$:}
Reminding that for $k \in \llbracket 1, n \rrbracket$, $P_{t_k} = \id + \gamma_k H_{t_k} $, the discretized stochastic process writes:
\begin{equation*}
    y_{k+1} = P_{t_k} y_k + r_k \gamma_k + \sqrt{2\alpha \xi_{T - t_k} \gamma_k} \cdot w_k, \quad w_k \sim \mathcal{N}(0, I).
\end{equation*}
By unrolling the recursion, we obtain:
\begin{align*}
    y_k &= \Psi_{k,0} y_0 + \gamma \sum_{\ell=0}^{k-1} \Psi_{k,\ell} r_\ell  + \sum_{\ell=0}^{k-1}\Psi_{k,\ell} \sqrt{2\alpha \xi_{T - t_\ell} \gamma_k} \cdot w_\ell, \\
    &= \Psi_{k,0} y_0 + \phi_k  + \eta_k
\end{align*}
with $\phi_k$ deterministic and 
\begin{align*}
\eta_k \eqdef \sum_{i=0}^{k-1}\Psi_{k,i} \sqrt{2\alpha \xi_{T - t_i} \gamma} \cdot w_i
\end{align*}
which follows a Gaussian distribution with zero-mean, independent of $y_0$. 
Thus
\begin{align*}
\cov(y^{(0)}_k, y^{(0)}_\ell) &= \Psi_{k,0} \cov(y_0) \Psi_{\ell,0}
+ \EE[\eta_k \eta_{\ell}^\top] .
\end{align*}
Using independence between the $(w_i)$,
\begin{align*}
     \EE[\eta_k \eta_{\ell}^\top]  &=  2\alpha \gamma \sum_{i=0}^{k-1} \sum_{l=0}^{j-1} \Psi_{k,i} \Psi_{\ell,l}  \sqrt{ \xi_{T - t_i} \xi_{T - t_l}} \EE[w_i w_l] \\
     &= 2\alpha \gamma \sum_{i=0}^{j-1}  \Psi_{k,i} \Psi_{\ell,i} \xi_{T - t_i} \\
    &= 2\alpha \gamma \sum_{i=0}^{j-1}  \hat \Psi_{t_k,t_i} \hat\Psi_{t_\ell,t_i} \xi_{T - t_i}
\end{align*}
where we remind that
\begin{align*}
\hat \Psi_{t,s} &= e^{\alpha (Q_{s} - Q_{t})} \left( \Sigma_{T-t} \Sigma_{T-s}^{-1} \right)^{\frac{\alpha+1}{2}} \left(\id - \frac{\gamma}{2}(H_t - H_s)\right) + \mathcal{O}(\gamma) .
\end{align*}
$ \EE[\eta_k \eta_{\ell}^\top]$ is thus the left-endpoint rectangle discretization, with interval $\gamma$, of the integral:
\begin{align*}
2 \alpha \int_0^{t_\ell} \hat \Psi_{t_k,s} \hat \Psi_{t_\ell,s} \xi_{T - s}ds
\end{align*}
and we have, from Lemma~\ref{lem:disc}, at zeroth-order in $\gamma$,
\begin{align}
\EE[\eta_k \eta_{\ell}^\top] &= 2\alpha  \int_0^{t_\ell} \hat \Psi_{t_k,s} \hat \Psi_{t_\ell,s} \xi_{T - s}ds + \mathcal{O}(\gamma)\\
&= 2\alpha \left( \Sigma_{T-t_k} \Sigma_{T-t_\ell} \right)^{\frac{\alpha+1}{2}} e^{- \alpha( Q_{t_k} + Q_{t_\ell})}\int_0^{t_\ell} e^{2 \alpha Q_{s}}  \xi_{T - s} \Sigma_{T-s}^{-(\alpha + 1)} ds + \mathcal{O}(\gamma) \label{eq:493}.
\end{align}
To solve the first integral, we use that
\begin{align*}
d[e^{2 \alpha Q_t} \Sigma_{T-t}^{-\alpha}] &= e^{2 \alpha Q_t} d[\Sigma_{T-t}^{-\alpha}] - 2 \alpha \beta_{T-t} e^{2 \alpha Q_t} \Sigma_{T-t}^{-\alpha} \\
&=  2 \alpha e^{2 \alpha Q_t} \Sigma_{T-t}^{-\alpha}(\xi_{T-t}\Sigma_{T-t}^{-1} + \beta_{T-t}\text{Id}) - 2 \alpha \beta_{T-t} e^{2 \alpha t} \Sigma_{T-t}^{-\alpha} \\
&= 2 \alpha \xi_{T-t} e^{2 \alpha Q_t} \Sigma_{T-t}^{-(\alpha+1)}
\end{align*}
to get
\begin{align*}
2 \alpha \int_0^{t_k} \xi_{T-s} e^{2 \alpha Q_s} \Sigma_{T-s}^{-(\alpha + 1)} ds  = \int_0^{t_k} d[e^{2 \alpha Q_s} \Sigma_{T-s}^{- \alpha}] = e^{2\alpha Q_t}\Sigma_{T-t}^{-\alpha} - \Sigma_{T}^{-\alpha} .
\end{align*}
And thus~\eqref{eq:493} simplifies to
\begin{align*}
     \EE[\eta_k \eta_{\ell}^\top] &=   e^{-\alpha (Q_{t_k}+Q_{t_\ell})} \left( \Sigma_{T-{t_k}} \Sigma_{T-{t_\ell}}\right)^{\frac{\alpha+1}{2}}  \left( e^{2\alpha Q_{t_\ell}}\Sigma_{T-{t_\ell}}^{-\alpha} - \Sigma_{T}^{-\alpha} \right) + \mathcal{O}(\gamma).
\end{align*}
The cross covariance $\cov(y^{(0)}_k, y^{(0)}_\ell)$ then writes
\begin{align} \label{eq:cross_cov_2}
\cov(y^{(0)}_k, y^{(0)}_\ell) &=  \Psi_{k,0} \cov(y_0) \Psi_{\ell,0}^\top + \EE[\eta_k \eta_{\ell}^\top] \\
&=  e^{\alpha (B_{T-t_k}-B_{T-t_\ell})} E_{T-t_k, T-t_\ell} + \mathcal{O}(\gamma),
\end{align}
where we define, for $0 \leq t,s \leq T$
\begin{equation*}
\begin{split}
    E_{t,s} &= e^{-2\alpha (B_T - B_s)}\Sigma_{t}^{\frac{\alpha+1}{2}} \Sigma_T^{-\frac{\alpha+1}{2}} \cov(y_0) \Sigma_T^{-\frac{\alpha+1}{2}}
\Sigma_{s}^{\frac{\alpha+1}{2}} \\
&+ (\Sigma_{t}\Sigma_{s})^{\frac{\alpha+1}{2}}  \left( \Sigma_{s}^{-\alpha} - e^{-2\alpha (B_T - B_s)}\Sigma_{T}^{-\alpha} \right).
\end{split}
\end{equation*}
For perfect initialization of the covariance $\cov(y_0) = \Sigma_T$, $E_{t,s}$ simplifies to
\begin{align*}
E_{t,s} &= \Sigma_{t} \left(\Sigma_{t}\Sigma_{s}^{-1}\right)^{(\alpha-1)/2}
\end{align*}
In other words,
\begin{align*}
\cov(y^{(0)}_k, y^{(0)}_\ell) &= e^{\alpha (B_{r_k}-B_{r_\ell})} \hat E^\gamma_{r_k, r_\ell} + \mathcal{O}(\gamma^2)
\end{align*}
where, 
\begin{align*}
\hat E^\gamma_{r_k, r_\ell} = E_{r_k, r_\ell} + \mathcal{O}(\gamma)
\end{align*}

Plugging~\eqref{eq:cross_cov_2} into~\eqref{eq:cov_entire}, we get the expected second-order expansion of the covariance. 
\end{proof}

\subsubsection{Wasserstein distance to the data distribution}
Using the above explicit solution $q^\varepsilon_k = \mathcal{N}(\mu^\varepsilon_k, \Sigma^\varepsilon_k)$, we can derive the expansion of the averaged Wasserstein distance 
\begin{equation*} 
    \EE_{\delta, \Delta} \left[W_2^2(p_\mathrm{data},q_k^\varepsilon)\right] =   \EE_{\delta, \Delta} \left[ \EE\norm{\mu - \mu_k^\varepsilon}^2 \right]+   \EE_{\delta, \Delta} \left[ \mathcal{B}^2(C_\mathrm{data}, \Sigma_k^{\varepsilon}) \right].
\end{equation*}
First, we remind that 
\begin{align*}
    \mu_t &= s_t \mu \\
    \Sigma_t &= s_t^2 ( C_{\mathrm{data}} + \sigma_t^2 \id).
\end{align*}
From the eigendecomposition of the covariance of the data $C_{\mathrm{data}} = U \operatorname{Diag}(\lambda_i) U^\top$, we can calculate the eigendecomposition of $\Sigma_t$
\begin{align*}
\Sigma_t \eqdef  s_t^2 ( C_{\mathrm{data}} + \sigma_t^2 \id) = U \operatorname{Diag}\left(\lambda^t_i\right) U^T
\end{align*}
where for $i \in \llbracket 1,d \rrbracket$
\begin{align*}
 \lambda^t_i \eqdef s_t^2 (\lambda_i + \sigma_t^2) .
\end{align*}

\subsubsection*{$L^2$ distance between means}
By the fact that $\EE_{\delta}\left[\mu_k^{(1)}(\delta)\right] = 0$ 
\begin{align}
\EE_{\delta} \norm{\mu - \EE[y_k]}^2 &= \EE_{\Delta, \delta}\norm{\mu - \hat \mu^\varepsilon_{k}}^2 \\
&= \EE_{\Delta, \delta}\norm{\mu - \left(\mu_k^{(0)} + \varepsilon \mu_k^{(1)}(\delta) + o(\varepsilon^2) \right)}  \\
&= \norm{\mu - \mu_k^{(0)}}^2 + \varepsilon^2 \EE_{\delta, \Delta} \norm{\mu_k^{(1)}(\delta)}^2 + o(\varepsilon^2)  \label{eq:509}.
\end{align}
For the first term, from Proposition~\ref{prop:pertubed_diffusion}, using the notation $r_k = T-t_k$.
\begin{align*}
\mu_k^{(0)} = \mu_{r_k} + \gamma d_{r_k} \cdot \mu + \mathcal{O}(\gamma^2)
\end{align*} 
such that
\begin{align} \label{eq:512}
\norm{\mu - \mu_k^{(0)}}^2 &= \norm{\mu - \mu_{r_k}}^2 -  2 \gamma \langle\mu - \mu_{r_k}, d_{r_k} \cdot \mu\rangle + \mathcal{O}(\gamma^2) \\
&= (1- s_{r_k})^2 \norm{\mu}^2 +  2 \gamma (s_{r_k}-1) \langle \mu, d_{r_k} \cdot \mu \rangle  +  \mathcal{O}(\gamma^2) .
\end{align} 
From Lemma~\ref{prop:diffusion_discretization_1}, for any $r \in [0,T]$, $d_r$ decomposes in the eigenbasis of the data as ${d_r = U \Lambda^{d}_r(\lambda_i) U^\top}$.
The zeroth-order distance between means~\eqref{eq:512} then simplifies to 
\begin{align*}
\norm{\mu - \mu_k^{(0)}}^2 &= (1- s_{r_k})^2 \norm{\mu}^2 + \gamma \sum_{i=1}^d 2 (s_{r_k}-1) \Lambda^{d}_{r_k}(\lambda_i)(U^\top \mu)^2_i. 
\end{align*} 
For the second term of~\eqref{eq:509}, from Proposition~\ref{prop:pertubed_diffusion}
\begin{align*}
    \mu_k^{(1)}(\delta) &=  \sum_{\ell=0}^{k-1} \gamma \, h_{r_k,r_\ell} \hat F^\gamma_{r_k, r_\ell} (\delta_{r_\ell} - \gamma \Delta_{r_\ell}d_{r_k} \cdot \mu) + \mathcal{O}(\gamma^2)  .
\end{align*}
By uncorrelation across time between the $(\delta_t)_{t\geq 0}$ and the $(\Delta_t)_{t\geq 0}$, and uncorrelation between~$\delta_t$ and~$\Delta_t$ at fixed~$t$, at order $1$ in $\gamma$ we have 
\begin{align*}
\EE_{\delta, \Delta} \norm{\mu_k^{(1)}(\delta)}^2 
&= \EE_{\delta, \Delta} \norm{ \sum_{\ell=0}^{k-1} \gamma \, h_{r_k,r_\ell} \hat F^\gamma_{r_k, r_\ell} (\delta_{r_\ell} - \gamma \Delta_{r_\ell}d_{r_k} \cdot \mu) + \mathcal{O}(\gamma^2)  }^2  \\
&= \sum_{\ell=0}^{k-1} \gamma^2  h_{r_k,r_\ell}^2 \EE_{\delta} \norm{ \hat F^\gamma_{r_k, r_\ell} (\delta_{r_\ell} - \gamma \Delta_{r_\ell}d_{r_k} \cdot \mu)}^2 + \mathcal{O}(\gamma^2) \\
&= \sum_{\ell=0}^{k-1} \gamma^2  h_{r_k,r_\ell}^2 \EE_{\delta} \norm{ \hat F^\gamma_{r_k, r_\ell} \delta_{r_\ell}}^2 + \mathcal{O}(\gamma^2) \\
&= \gamma \sum_{\ell=0}^{k-1} \gamma h_{r_k,r_\ell}^2 \left\langle \cov \delta_{r_\ell},  \left(\hat F^\gamma_{r_k,r_\ell}\right)^2 \right\rangle + \mathcal{O}(\gamma^2).
\end{align*}
Using that $\hat F^\gamma_{r_k,r_\ell} = F_{r_k,r_\ell} + \mathcal{O}(\gamma)$,
\begin{align*}
\EE_{\delta, \Delta} \norm{\mu_k^{(1)}(\delta)}^2  &= \gamma \sum_{\ell=0}^{k-1} \gamma h_{r_k,r_\ell}^2 \left\langle \cov \delta_{r_\ell},   F_{r_k,r_\ell}^2 + \mathcal{O}(\gamma) \right\rangle + \mathcal{O}(\gamma^2) \\
&= \gamma \sum_{\ell=0}^{k-1} \gamma h_{r_k,r_\ell}^2 \left\langle \cov \delta_{r_\ell},   F_{r_k,r_\ell}^2 \right\rangle + \mathcal{O}(\gamma^2).
\end{align*}
Using the eigendecomposition $F_{t,s} = U \diag{\left(\frac{\lambda_i^t}{\lambda_i^s}\right)^{\frac{\alpha+1}{2}}} U^\top $, 
\begin{align*}
\langle \cov \delta_{r_\ell}, F_{r_k,r_\ell}^2 \rangle
&= \sum_{i=1}^d\left(\frac{\lambda_i^{r_k}}{\lambda_i^{r_\ell}}\right)^{\alpha+1} \EE_{\delta} \left[\langle \cov \delta_{r_\ell}, u_i u_i^\top \rangle\right] .
\end{align*}

The overall distance between means is thus 
\begin{equation} \label{eq:L2_means}
\begin{split}
    \EE_{\delta, \Delta} \left[\norm{\mu - \mu_k^\varepsilon }^2\right]  
    &= (1- s_{r_k})^2\norm{\mu}^2  + \gamma \sum_{i=1}^d 2 (s_{r_k}-1) \Lambda^{d}_{r_k}(\lambda_i)(U^\top \mu)^2_i  \\
    &+ \gamma \varepsilon^2 \underbrace{\sum_{i=1}^d \sum_{\ell=0}^{k-1} \gamma h_{r_k,r_\ell}^2 \left(\frac{\lambda_i^{r_k}}{\lambda_i^{r_\ell}}\right)^{\alpha+1} \EE_{\delta} \left[\langle \cov \delta_{r_\ell}, u_i u_i^\top \rangle\right]}_{\mathcal{E}_{T-t_k}^{\text{mean}}(\delta)} + \mathcal{O}(\gamma^2).
\end{split}
\end{equation}
The term $\mathcal{E}_{T-t_k}^{\text{mean}}(\delta)$ is the left-endpoint rectangle approximation of an integral $\int_0^{T-t_k}$. Using Lemma~\ref{lem:disc}, it can be replaced, at zeroth-order in $\gamma$,  by its corresponding integral : 
\begin{equation*} 
\mathcal{E}_{T-t_k}^{\text{mean}}(\delta) = \sum_{i=1}^d \int_0^{T-t_k} h_{T-t_k,T-s}^2 \left(\frac{\lambda_i^{T-t_k}}{\lambda_i^{T-s}}\right)^{\alpha+1} \EE_{\delta} \left[\langle \cov \delta_{T-s}, u_i u_i^\top \rangle\right]ds + \mathcal{O}(\gamma).
\end{equation*}
With the change of variable $s \to T-s$, for any $t \in [0,T]$, $\mathcal{E}_{r}^{\text{mean}}(\delta)$ writes
\begin{equation*} 
\mathcal{E}_{r}^{\text{mean}}(\delta) = \sum_{i=1}^d \int_r^{T} h_{r,s}^2 \left(\frac{\lambda_i^{r}}{\lambda_i^{s}}\right)^{\alpha+1} \EE_{\delta} \left[\langle \cov \delta_{s}, u_i u_i^\top \rangle\right]ds + \mathcal{O}(\gamma).
\end{equation*}

The first two terms of~\eqref{eq:L2_means} will make part of the  $\mathcal{E}_{r}^{(0)}$ and $\mathcal{E}_{r}^{disc}$ of the Theorem, along with terms arising from the Bures distance, which is now calculated.

\subsubsection*{Bures distance between covariances} Similarly to the proof of Theorem~\ref{thm:langevin_distance_general}, we use the second-order Taylor expansion of the Bures distance of Proposition~\ref{prop:Langevin_2}. Again, from the linearity of the first-order term $\Sigma_k^{(1)}(\Delta)$ in the $\Delta_{r_k}$ and the fact that the $\Delta_{r_k}$ has zero-mean, when taking the mean of the Bures distance in $\Delta$, the first-order term in $\varepsilon$ disappears. We are then left to calculating 
\begin{align*}
\EE_{\Delta} \left[ \mathcal{B}^2(C_\mathrm{data}, \cov(y_{k}))\right] &= \underbrace{\mathcal{B}^2(C_\mathrm{data}, \Sigma^{(0)}_k)}_{\text{term A}} \\
& \nonumber+ \varepsilon^2 \underbrace{2\tr{L_{\left(X^{(0)}_{k}\right)^{1/2}}^{-1}\left[\EE_{\Delta}\left[\left(L_{\left(X^{(0)}_{k}\right)^{1/2}}^{-1}\left[X^{(1)}_{k}(\Delta)\right]\right)^2\right]\right] }}_{\text{term B}} \\ \nonumber
&+ \varepsilon^2 \underbrace{\tr{ \EE_{\Delta}[\Sigma^{(2)}_{k}(\Delta)]
- 2 L_{\left(X^{(0)}_{k}\right)^{1/2}}^{-1}\left[\EE_{\Delta}[X^{(2)}_{k}(\Delta)]\right]}}_{\text{term C}}  +o(\varepsilon^2)
\end{align*}
where
\begin{empheq}[left=\empheqlbrace]{align}
        X^{(0)}_k &\eqdef {C_\mathrm{data}}^{1/2}\Sigma_k{C_\mathrm{data}}^{1/2} \\
        X^{(1)}_k (\Delta) &\eqdef {C_\mathrm{data}}^{1/2}\Sigma^{(1)}_k (\Delta) {C_\mathrm{data}}^{1/2} \\
        X^{(2)}_k(\Delta) &\eqdef {C_\mathrm{data}}^{1/2}\Sigma^{(2)}_k (\Delta) {C_\mathrm{data}}^{1/2}.
\end{empheq}
\paragraph{Term A}
First, let's calculate the term A. We have $\Sigma^{(0)}_k = \Sigma_{r_k} + \gamma D_{r_k} + \mathcal{O}(\gamma)$, 
where $D_{r_k}$, diagonalizes in the eigenbasis of the data as $D_{r_k} = U \Lambda^{D}_{r_k}(\lambda_i)U^\top$.
Using successively the first-order expansion $(a + \gamma b)^{\theta} = a^{\theta} + \gamma\theta a^{\theta-1} b + \mathcal{O}(\gamma^2)$ for $\theta = 1/2$ and $\theta =2$, the term A then simply writes 
\begin{align*}
\mathcal{B}^2(C_\mathrm{data}, \Sigma^{(0)}_k) &= \sum_{i=1}^d \left( \left(\lambda^{r_k}_i + \gamma \,\Lambda^{D}_{r_k}(\lambda_i) + \mathcal{O}(\gamma^2) \right)^{1/2} - (\lambda_i)^{1/2} \right)^2 \\
&=  \sum_{i=1}^d \left( \left(\lambda^{r_k}_i\right)^{1/2} + \gamma  \frac{1}{2} \frac{\Lambda^{D}_{r_k}(\lambda_i)}{(\lambda^{r_k}_i)^{1/2}} + \mathcal{O}(\gamma^2) - (\lambda_i)^{1/2} \right)^2 \\
&= \sum_{i=1}^d \left( \left(\lambda^{r_k}_i\right)^{1/2} - (\lambda_i)^{1/2}  \right)^2 + \gamma \sum_{i=1}^d  \left((\lambda^{r_k}_i)^{1/2}  - (\lambda_i)^{1/2} \right) \frac{\Lambda^{D}_{r_k}(\lambda_i)}{(\lambda^{r_k}_i)^{1/2}} + \mathcal{O}(\gamma^2)\\
&= \sum_{i=1}^d \left(\sqrt{\lambda_i} - \sqrt{\lambda^{r_k}_i} \right)^2 + \gamma \sum_{i=1}^d\Lambda^{D}_{r_k}(\lambda_i) \left( 1 - \sqrt{\frac{\lambda_i}{\lambda^{r_k}_i}} \right) + \mathcal{O}(\gamma^2)
\end{align*}
\paragraph{Term B.} Denoting, for $0 \leq t \leq T$, $\eta_i^t \eqdef \lambda_i \lambda^t_i$ and $ Z_k \eqdef L_{\left(X_k^{(0)}\right)^{1/2}}^{-1}[X_k^{(1)}(\Delta)] $, 
we have, using the decomposition of the inverse Lyapunov operator given in Lemma~\ref{lem:SGD_3},
\begin{align*}
L_{(X^{(0)}_k)^{1/2}}^{-1}\left[Z_k^2\right] &=  U \left( U^\top Z_k^2 U \odot \left(\frac{1}{ (\eta^{r_k}_i)^{1/2} + (\eta^{r_k}_j)^{1/2} }\right)_{ij} \right) U^\top 
\end{align*}
such that
\begin{align*}
\EE_{\Delta} \tr{L_{(X^{(0)}_k)^{1/2}}^{-1}\left[Z_k^2\right]} &= \sum_i  \left( U^\top \EE_{\Delta} \left[  Z_k^2\right] U \right)_{ii} \frac{1}{2 (\eta^{r_k}_i)^{1/2}} .
\end{align*}
We have, again using the decomposition of the inverse Lyapunov operator
\begin{align*}
    \left(U^\top \EE_{\Delta}\left[Z_k^2\right] U \right)_{ii} &=   
    \left( (U^\top X^{(1)}_{k} (\Delta) U) \odot \left(\frac{1}{ (\eta^{r_k}_i)^{1/2} + (\eta^{r_k}_j)^{1/2} }\right)_{ij} \right)^2_{ii} \\
    &=
    \sum_{j} \left(\EE_{\Delta}\left[(U^\top X^{(1)}_k(\Delta) U)^2 \right]\right)_{ij} \left(\frac{1}{ (\eta^{r_k}_i)^{1/2} + (\eta^{r_k}_j)^{1/2} }\right)^2 .
\end{align*}
Such that, finally,
\begin{align} \label{eq:termB_1}
\text{term B} &=  \sum_{ij} \left(\EE_{\Delta}\left[(U^\top X^{(1)}_k(\Delta) U)^2 \right]\right)_{ij} \left(\frac{1}{ (\eta^{r_k}_i)^{1/2} + (\eta^{r_k}_j)^{1/2} }\right)^2 \frac{1}{(\eta^{r_k}_i)^{1/2}}
\end{align}
We got in Proposition~\ref{prop:pertubed_diffusion}, using $g_{t,s} = e^{\alpha(B_{t} - B_{s})} h_{t,s}$, 
\begin{align*}
 \Sigma_{t}^{(1)}(\Delta) 
 &= - \sum_{\ell=0}^{k-1} \gamma \, g_{r_k,r_\ell} \left( \hat F^\gamma_{r_j,r_\ell} \Delta_{r_\ell} \hat E^\gamma_{r_k,r_\ell} +   \hat E^\gamma_{r_k,r_\ell} \Delta_{r_\ell}^\top \hat F^\gamma_{r_k,r_\ell} \right) + \mathcal{O}(\gamma^2)
\end{align*}
where, $\hat E^\gamma_{t,s} = E_{t,s}+\mathcal{O}(\gamma)$ and $\hat F^\gamma_{t,s} = F_{t,s}+\mathcal{O}(\gamma)$, and
\begin{align}
E_{t,s} \label{eq:LE}
&= \Sigma_{t} (\Sigma_{t}\Sigma_{s}^{-1})^{(\alpha-1)/2} = U \diag{\lambda^E_{t,s}}_i U^\top \quad \text{with}  \quad 
(\lambda^E_{t,s})_i \eqdef \lambda^t_i \left(\frac{\lambda^t_i}{\lambda^s_i}\right)^{(\alpha-1)/2} \\ \label{eq:LF}
F_{t,s} &= (\Sigma_{t} \Sigma_{s}^{-1})^{\frac{\alpha+1}{2}} = U \diag{\lambda^F_{t,s}}_i U^\top \quad \text{with}  \quad 
(\lambda^F_{t,s})_i \eqdef \left(\frac{\lambda^t_i}{\lambda^s_i}\right)^{\frac{\alpha+1}{2}}
\end{align}
Therefore $\Sigma_{t}^{(1)}(\Delta)$  writes in the eigenbasis of $C_\mathrm{data}$ as  
\begin{equation*}
\begin{split}
 \left(U^\top \Sigma_{t}^{(1)}(\Delta) U\right)_{ij}
 = -  \sum_{\ell=0}^{k-1} \gamma \,  &g_{r_k,r_\ell} \Biggl( \left\langle \Delta_{r_\ell}, u_i u_j^\top \right\rangle \biggl( (\lambda^F_{r_k,r_\ell})_i (\lambda^E_{r_k,r_\ell})_j + \gamma C_{i,j, l} \biggr)\\
 &+ \left\langle \Delta_{r_\ell}, u_j u_i^\top \right\rangle \biggl( (\lambda^E_{r_k,r_\ell})_i  (\lambda^F_{r_k,r_\ell})_j + \gamma C'_{i,j,l} \biggr) \Biggr) + \mathcal{O}(\gamma^2)
\end{split}
\end{equation*}
for some constants $C_{i,j,l}$ and $C'_{i,j,l}$.

We deduce that $ X^{(1)}_k (\Delta) = {C_\mathrm{data}}^{1/2}\Sigma^{(1)}_k (\Delta) {C_\mathrm{data}}^{1/2}$ writes in the eigenbasis of $C_\mathrm{data}$ as 
\begin{equation*}
\begin{split}
 (U^\top X_{t}^{(1)}(\Delta) U)_{ij} = -  \sqrt{\lambda_i \lambda_j}  &\sum_{\ell=0}^{k-1} \gamma \,  g_{r_k,r_\ell} \Biggl( \langle \Delta_{r_\ell}, u_i u_j^\top \rangle \biggl( (\lambda^F_{r_k,r_\ell})_i (\lambda^E_{r_k,r_\ell})_j + \gamma C_{i,j, l} \biggr) \\
 &+ \langle \Delta_{r_\ell}, u_j u_i^\top \rangle \biggl( (\lambda^E_{r_k,r_\ell})_i  (\lambda^F_{r_k,r_\ell})_j + \gamma C'_{i,j,l} \biggr) \Biggr)   + \mathcal{O}(\gamma)
\end{split}
\end{equation*}
Using the independence across time between the $(\Delta_t)_{t \geq 0}$, the expectation of the square of the above:
\begin{equation*}
\begin{split}
\EE_\Delta[ (U^\top X_{t}^1(\Delta) U)^2]_{ij}
 = \lambda_i \lambda_j  \sum_{\ell=0}^{k-1} \gamma^2 \,  g_{r_k,r_\ell}^2 \EE_\Delta\biggl[\biggl( &\langle \Delta_{r_\ell}, u_i u_j^\top \rangle (\lambda^F_{r_k,r_\ell})_i (\lambda^E_{r_k,r_\ell})_j \\
 &+ \langle \Delta_{r_\ell}, u_j u_i^\top \rangle (\lambda^E_{r_k,r_\ell})_i (\lambda^F_{r_k,r_\ell})_j \biggr)^2\biggr] + \mathcal{O}(\gamma^2) .
\end{split}
\end{equation*}
Note that the constants $C_{i,j,l}$ and $C'_{i,j,l}$ were absorbed in the $\mathcal{O}(\gamma^2)$.
Denoting, for $0 \leq t,s \leq T$, the kernel
\begin{align} \label{eq:Gts}
    G_{t,s}(\lambda_i, \lambda_j) &\eqdef g_{t,s} (\lambda_i \lambda_j)^{1/2} \left(\frac{1}{ (\eta^t_i)^{1/2} + (\eta^t_j)^{1/2} }\right)  \frac{1}{(\eta^t_i)^{1/4}}  \\
    &= e^{\alpha(B_t-B_s)}h_{t,s}  \left(\frac{(\lambda_i \lambda_j)^{1/2}}{ (\lambda_i \lambda^t_i)^{1/2} + (\lambda_j \lambda^t_j)^{1/2} }\right)  \frac{1}{(\lambda_i \lambda^t_i)^{1/4}}
\end{align}
and for simplicity $(G_{t,s})_{ij} =  G_{t,s}(\lambda_i, \lambda_j)$
which is a symmetric matrix,~\eqref{eq:termB_1} simplifies to
\begin{align*}
\text{term B} &=  \sum_{ij} \left(\EE_{\Delta}\left[(U^\top X^{(1)}_k(\Delta) U)^2 \right]\right)_{ij} \left(\frac{1}{ (\eta^t_i)^{1/2} + (\eta^{r_k}_j)^{1/2} }\right)^2 \frac{1}{(\eta^{r_k}_i)^{1/2}} \\
&= \sum_{ij} \sum_{\ell=0}^{k-1} \gamma^2 \,  G_{r_k,r_\ell}^2 \Biggl( \EE_{\Delta}\left[\langle\Delta_{r_\ell}, u_i u_j^\top \rangle^2\right] (\lambda^F_{r_k,r_\ell})^2_i (\lambda^E_{r_k,r_\ell})^2_j \\
&+  \EE_{\Delta}\left[\langle\Delta_{r_\ell}, u_j u_i^\top \rangle^2\right] (\lambda^F_{r_k,r_\ell})^2_j (\lambda^E_{r_k,r_\ell})^2_i \nonumber \\
&+ 2 \EE_{\Delta}\left[\langle\Delta_{r_\ell}, u_i u_j^\top \rangle\langle\Delta_{r_\ell}, u_j u_i^\top \rangle\right]  (\lambda^F_{r_k,r_\ell})_i (\lambda^F_{r_k,r_\ell})_j (\lambda^E_{r_k,r_\ell})_i (\lambda^E_{r_k,r_\ell})_j \Biggr) \nonumber \\
&= \sum_{ij} \sum_{\ell=0}^{k-1} 2 \gamma^2 \,  G_{r_k,r_\ell}^2 \Biggl( \EE_{\Delta}\left[\langle\Delta_{r_\ell}, u_i u_j^\top \rangle^2\right] (\lambda^F_{r_k,r_\ell})^2_i (\lambda^E_{r_k,r_\ell})^2_j \\
&+ \EE_{\Delta}\left[\langle\Delta_{r_\ell}, u_i u_j^\top \rangle\langle\Delta_{r_\ell}, u_j u_i^\top \rangle\right] (\lambda^F_{r_k,r_\ell})_i (\lambda^F_{r_k,r_\ell})_j (\lambda^E_{r_k,r_\ell})_i (\lambda^E_{r_k,r_\ell})_j \Biggr) \nonumber
\end{align*}

\paragraph{Term C.} First, in the same way as for the term B,
\begin{align*}
   \text{term C} &=   \tr{ \EE_{\Delta}[\Sigma^{(2)}_{k}(\Delta)]} - 2 \tr{L_{\left(X^{(0)}_{k}\right)^{1/2}}^{-1}\left[\EE_{\Delta}[X^{(2)}_{k}(\Delta)]\right]} \\
   &= \sum_i \left( U^\top \EE \left[ \Sigma^{(2)}_{k}(\Delta) \right] U \right)_{ii}  - 2 \sum_i \left( U^\top \EE \left[ X^{(2)}_{k}(\Delta) \right] U \right)_{ii} \frac{1}{2 (\eta^{r_k}_i)^{1/2}} \\
    &= \sum_i \left( U^\top \EE \left[ \Sigma^{(2)}_{k}(\Delta) \right] U \right)_{ii}  - \sum_i \left( U^\top \EE \left[ X^{(2)}_{k}(\Delta) \right] U \right)_{ii} \frac{1}{ (\eta^{r_k}_i)^{1/2}} 
\end{align*}
Recall that $X^{(2)}_{k}(\Delta) = {\Sigma}^{1/2}\Sigma^{(2)}_{k}(\Delta) {\Sigma}^{1/2}$. Therefore, 
\begin{align*}
   \text{term C} &=  
     \sum_i \left( U^\top \EE \left[ \Sigma^{(2)}_{k}(\Delta) \right] U \right)_{ii}  - \sum_i \left( U^\top \EE \left[ \Sigma^{(2)}_{k}(\Delta) \right] U \right)_{ii} \frac{\lambda_i}{ (\eta^{r_k}_i)^{1/2}} \\
     &=  \sum_i \left( U^\top \EE \left[ \Sigma^{(2)}_{k}(\Delta) \right] U \right)_{ii}\left( 1 - \sqrt{\frac{\lambda_i}{\lambda^{r_k}_i }} \right)
\end{align*}
We calculate the expectation $\EE \left[ \Sigma^{(2)}_{k}(\Delta) \right]$ in the eigenbasis of the data. We established in  Proposition~\ref{prop:pertubed_diffusion}:   
\begin{align*}
\Sigma_{k}^{(2)}(\Delta) &=  \sum_{\ell=0}^{k-1} \sum_{i=0}^{k-1} \gamma^2 \,
 h_{r_k,r_{\ell}} h_{r_k,r_{i}} \cdot \\ \nonumber
 &\Biggl( e^{\alpha(B_{r_k}- B_{r_i})} \left(\hat F^\gamma_{r_k,s_{\ell}} \Delta_{r_\ell} \hat F^\gamma_{r_\ell,r_i} \Delta_{r_i} \hat E^\gamma_{r_k,r_i}
 + \hat E^\gamma_{r_k,r_i} \Delta_{r_i}^\top \hat F^\gamma_{r_\ell,r_i} \Delta_{r_\ell}^\top \hat F^\gamma_{r_k,r_\ell} \right)  \\ \nonumber  &+  e^{\alpha(B_{r_i}- B_{r_\ell})} \hat F^\gamma_{r_k,r_\ell} \Delta_{r_\ell} \hat E^\gamma_{r_\ell, r_i} \Delta_{r_i}^\top \hat F^\gamma_{r_k,r_i} \Biggl) + \mathcal{O}(\gamma^2)
 \end{align*}
Taking the expectation with respect to the $(\Delta_{t_\ell})_{\ell}$, by uncorrelation across time between the  $(\Delta_{t_\ell})_{\ell}$, we get, using that $F_{t,t} = \id$ and $E_{t,t} = \Sigma_t$,
\begin{equation*}
\begin{split}
 \Sigma_{k}^{(2)}(\Delta) =  \sum_{\ell=0}^{k-1} \gamma^2 \,
 h_{r_k,r_{\ell}}^2 \biggl( & e^{\alpha(B_{r_k}- B_{r_\ell})} \left(F_{r_k,s_{\ell}} \Delta_{r_\ell}^2 E_{r_k,r_\ell} 
 + E_{r_k,r_\ell} (\Delta_{r_i}^\top)^2 F_{r_k,r_\ell} \right) \\ 
 &+ F_{r_k,r_\ell} \Delta_{r_\ell} \Sigma_{r_\ell} \Delta_{r_\ell}^\top F_{r_k,{r_\ell}} \biggr) + \mathcal{O}(\gamma^2) 
 \end{split}
 \end{equation*}
Its projection in the eigenbasis of $C_{\mathrm{data}}$ satisfies, at position $(i,i)$,
\begin{equation*}
\begin{split}
\left(U^\top \EE \left[ \Sigma^{(2)}_{t}(\Delta) \right] U \right)_{ii} =  \sum_{\ell=0}^{k-1} \gamma^2 h_{r_k,r_{\ell}}^2 \bigg( &2  e^{\alpha(B_{r_k}- B_{r_\ell})}(\lambda^F_{r_k,r_{\ell}})_i  (\lambda^E_{r_k,r_{\ell}})_i \EE[\langle \Delta_{r_l}^2, u_i u_i^\top \rangle ] \\ &+  (\lambda^F_{r_k,r_{\ell}})^2_i  \EE[\langle \Delta_{r_\ell}  \Sigma_{r_\ell}  \Delta_{r_\ell}^\top, u_i u_i^\top \rangle ]   \biggr) + \mathcal{O}(\gamma^2) 
\end{split}
\end{equation*}
The above terms can be written with respect to the projections of the perturbations in the outer product space of the data:
\begin{align*}
\EE_{\Delta}[\langle \Delta_{r_\ell}^2, u_i u_i^\top \rangle ] &= \sum_j \EE_{\Delta}\left[\langle \Delta_{r_\ell}, u_i u_j^\top \rangle \langle \Delta_{r_\ell}, u_j u_i^\top \rangle \right] 
\end{align*}
and 
\begin{align*}
 \EE[\langle \Delta_{r_\ell}  \Sigma_{r_\ell}  \Delta_{r_\ell}^\top, u_i u_i^\top \rangle ]  &= \sum_j \lambda^{r_\ell}_j  \EE_{\Delta}\left[\langle \Delta_{r_\ell}, u_i u_j^\top \rangle^2 \right] 
\end{align*}
Finally, we can develop the term C as
\begin{align*}
   \text{term C} 
     &=  \sum_i \left( U^\top \EE \left[ \Sigma^{(2)}_{k}(\Delta) \right] U \right)_{ii}\left( 1 - \sqrt{\frac{\lambda_i}{\lambda^{r_k}_i }} \right) \\
     &= \sum_{i,j}  \sum_{\ell=0}^{k-1} \gamma^2 h_{r_k,r_{\ell}}^2 \left( 1 - \sqrt{\frac{\lambda_i}{\lambda^{r_k}_i }}\right) \Biggl((\lambda^F_{r_k,r_{\ell}})^2_i \, \lambda^{r_\ell}_j \, \EE_{\Delta}\left[\langle \Delta_{r_\ell}, u_i u_j^\top \rangle^2 \right] \\
     &+ 2 e^{\alpha(B_{r_k}- B_{r_\ell})}  (\lambda^F_{r_k,r_{\ell}})_i  (\lambda^E_{r_k,r_{\ell}})_i \EE_{\Delta}\left[\langle \Delta_{r_\ell}, u_i u_j^\top \rangle \langle \Delta_{r_\ell}, u_j u_i^\top \rangle \right] \Biggr) + \mathcal{O}(\gamma^2) \nonumber
\end{align*}
\paragraph{Overall Bures distance expression} Grouping the previous developments of the term A, B and C, we get
\begin{equation} \label{eq:expectation_W2_diff}
\begin{split}
\EE_{\Delta}[\mathcal{B}_2^2(\Sigma_{k}, C_\mathrm{data})] &=  \sum_{i=1}^d \left(\sqrt{\lambda_i} - \sqrt{\lambda^{r_k}_i} \right)^2 + \gamma \sum_{i=1}^d \Lambda^{D}_{r_k}(\lambda_i) \left( 1 - \sqrt{\frac{\lambda_i}{\lambda^{r_k}_i}} \right)\\
& + \varepsilon^2 \gamma \left( \sum_{i,j=1}^d   \sum_{\ell=0}^{k-1} \gamma\kappa^{(1)}_{r_k, r_l} (\lambda_i, \lambda_j) \, \EE_{\Delta}[\langle \Delta_{r_\ell}, u_iu_j^\top \rangle^2] \right. \\
&+ \left. \sum_{i,j=1}^d  \sum_{\ell=0}^{k-1} \gamma \kappa^{(2)}_{r_k, r_l}(\lambda_i,\lambda_j) \, \EE_{\Delta}[\langle \Delta_{r_\ell}, u_iu_j^\top \rangle \langle \Delta_{r_\ell}, u_ju_i^\top \rangle ] \right) + o(\varepsilon^2, \gamma^2) 
\end{split}
\end{equation}
where $\kappa^{(1)}_{r_k, r_l}, \kappa^{(2)}_{r_k, r_l} : \RR^2 \to \RR$ are kernel functions on the eigenvalues $\lambda_i$ of $C_\mathrm{data}$. We give here their expression with respect to $G_{t,s}$~\eqref{eq:Gts}, $\lambda^E_{t,s}$~\eqref{eq:LE} and $\lambda^F_{t,s}$~\eqref{eq:LF}. For $0 \leq t,s \leq T$, 
\begin{align} \label{eq:kappa1_diff}
\kappa^{(1)}_{t,s}(\lambda_i,\lambda_j) &=  2 G_{t,s}(\lambda_i,\lambda_j)^2 (\lambda^F_{t,s})^2_i (\lambda^E_{t,s})^2_j +  h_{t,s}^2 \left( 1 - \left( \frac{\lambda_i}{\lambda^t_i }\right)^{1/2}\right) (\lambda^F_{t,s})^2_i \lambda^s_j 
\end{align}
\begin{equation} \label{eq:kappa2_diff}
\begin{split}
    \kappa^{(2)}_{t,s}(\lambda_i,\lambda_j) &= 2 G_{t,s}(\lambda_i,\lambda_j)^2 (\lambda^F_{t,s})_i (\lambda^F_{t,s})_j (\lambda^E_{t,s})_i (\lambda^E_{t,s})_j \\ &+ 2 h_{t,s}^2  e^{\alpha(B_{t}- B_{s})} \left( 1 - \left( \frac{\lambda_i}{\lambda^t_i }\right)^{1/2}\right) (\lambda^F_{t,s})_i (\lambda^E_{t,s})_i
\end{split}
\end{equation}

The sums $\sum_{\ell=0}^{k-1} \gamma \cdot $ in~\eqref{eq:expectation_W2_diff} correspond to left-endpoint rectangle approximation of integrals $\int_0^{T-t_k}$. Using Lemma~\ref{lem:disc}, they can be replaced by integrals at zeroth-order in $\gamma$. Denoting $\mathcal{E}_{T-t_k}^{\text{cov}}(\Delta)$ the term in $\gamma \varepsilon^2$ of~\eqref{eq:expectation_W2_diff}, 
\begin{align*}
\mathcal{E}_{T-t_k}^{\text{cov}}(\Delta) &=  \sum_{i,j=1}^d   \int_0^{T-t_k} \kappa^{(1)}_{T-t_k, T-s} (\lambda_i, \lambda_j) \, \EE_{\Delta}[\langle \Delta_{T-s}, u_iu_j^\top \rangle^2] ds  \\
&+  \sum_{i,j=1}^d  \int_0^{T-t_k} \kappa^{(2)}_{T-t_k, T-s}(\lambda_i,\lambda_j) \, \EE_{\Delta}[\langle \Delta_{T-s}, u_iu_j^\top \rangle \langle \Delta_{T-s}, u_ju_i^\top \rangle ]ds + \mathcal{O}(\gamma) \nonumber
\end{align*} 
Using the change of variable $s \to T-s$, for any $r \in [0,T]$, $\mathcal{E}_{r}^{\text{cov}}(\Delta)$ is then defined by 
\begin{align*}
\mathcal{E}_{r}^{\text{cov}}(\Delta) &=  \sum_{i,j=1}^d   \int_r^{T} \kappa^{(1)}_{r, s} (\lambda_i, \lambda_j) \, \EE_{\Delta}[\langle \Delta_{s}, u_iu_j^\top \rangle^2] ds  \\
&+  \sum_{i,j=1}^d  \int_r^{T} \kappa^{(2)}_{r, s}(\lambda_i,\lambda_j) \, \EE_{\Delta}[\langle \Delta_{s}, u_iu_j^\top \rangle \langle \Delta_{s}, u_ju_i^\top \rangle ]ds  + \mathcal{O}(\gamma)\nonumber
\end{align*} 

\paragraph{Global Wasserstein distance} Grouping the terms from the $L^2$ distance between means~\eqref{eq:L2_means} and the above Bures distance between covariance, the overall $W_2$ distance writes
\begin{equation*}
\begin{split}
    \EE_{\delta, \Delta} \left[W_2^2(p_\mathrm{data},q_k^\varepsilon)\right]  &= \underbrace{ (1- s_{r_k})^2\norm{\mu}^2 + 
    \sum_{i=1}^d \left(\sqrt{\lambda_i} - \sqrt{\lambda^{r_k}_i} \right)^2}_{\mathcal{E}_{T-t_k}^{(0)}} \\
    &+ \gamma \underbrace{\left( \sum_{i=1}^d\Lambda^{D}_{r_k}(\lambda_i) \left( 1 - \sqrt{\frac{\lambda_i}{\lambda^{r_k}_i}} \right)  + \sum_{i=1}^d 2 (s_{r_k}-1) \Lambda^{d}_{r_k}(\lambda_i)(U^\top \mu)^2_i \right)}_{\mathcal{E}_{T-t_k}^{disc}}  \\
    &+ \gamma \varepsilon^2 \biggr( \underbrace{\sum_{i=1}^d \int_{r_k}^{T} h_{r_k,s}^2 \left(\frac{\lambda_i^{r_k}}{\lambda_i^{s}}\right)^{\alpha+1} \EE_{\delta} \left[\langle \cov \delta_{s}, u_i u_i^\top \rangle\right]ds}_{\mathcal{E}_{T-t_k}^{mean}(\delta)} \\
&
\left.
\begin{array}{l} 
\displaystyle + \sum_{i,j=1}^d   \int_{r_k}^{T} \kappa^{(1)}_{r_k, s} (\lambda_i, \lambda_j) \, \EE_{\Delta}[\langle \Delta_{s}, u_iu_j^\top \rangle^2] ds \\
\displaystyle +  \sum_{i,j=1}^d  \int_{r_k}^{T} \kappa^{(2)}_{r_k, s}(\lambda_i,\lambda_j) \, \EE_{\Delta}[\langle \Delta_{s}, u_iu_j^\top \rangle \langle \Delta_{s}, u_ju_i^\top \rangle ]ds \biggr) + \mathcal{O}(\gamma^2, \varepsilon^2)
\end{array} 
\right\}
{\scriptsize \mathcal{E}_{T-t_k}^{cov}(\Delta)} 
\end{split}
\end{equation*}

\paragraph{Simplification of $\kappa^{(1)}_{t, s}$ and $ \kappa^{(1)}_{t, s}$ when $t=0$.}
From their expressions equations~\eqref{eq:kappa1_diff} and~\eqref{eq:kappa2_diff}, we have when $r=0$, using that $h_{0,s} = (1+\alpha) e^{-\alpha B_s} \xi_s$, 
\begin{align*}
    G_{0,s}(\lambda_i, \lambda_j) = (1+\alpha) e^{-2\alpha B_s} \xi_s \frac{\sqrt{\lambda_j}}{ \lambda_i  + \lambda_j }
\end{align*}
and
\begin{align} \label{eq:kappa1_diff_simp}
\kappa^{(1)}_{0,s}(\lambda_i,\lambda_j) &=  2 (1+\alpha)^2 e^{-4 \alpha B_s} \xi_s^2 \frac{\lambda_j^3}{ (\lambda_i  + \lambda_j)^2 } \left(\frac{\lambda_i}{\lambda^s_i}\right)^{\alpha+1}  \left(\frac{\lambda_j}{\lambda^s_j}\right)^{2(\alpha-1)}
\end{align}
\begin{equation} \label{eq:kappa2_diff_simp}
\begin{split}
    \kappa^{(2)}_{0,s}(\lambda_i,\lambda_j) &= 2 (1+\alpha)^2 e^{-4 \alpha B_s} \xi_s^2 \frac{\lambda_i \lambda_j^2}{ (\lambda_i  + \lambda_j)^2 } \left(\frac{\lambda_i}{\lambda^s_i}\right)^{(3\alpha-1)/2} \left(\frac{\lambda_j}{\lambda^s_j}\right)^{(3\alpha-1)/2}  
\end{split}
\end{equation}

\end{proof}

\subsection{Proof of Corollary~\ref{cor:diffusion_final}} \label{app:diffusion_final}
\newtheorem*{repeatcor2}{Corollary 2}
\begin{repeatcor2}[Diffusion sampling error with linear score trained by SGD]
Under Assumption~\ref{ass:gaussian}, consider a linear score $v_t(x, \theta) = - A_t x + b_t$ with parameters \( \theta=(A_t,b_t)\) trained at each $t$ via the SGD algorithm~\eqref{eq:score_matching_SGD_emp}, with \( N \) data samples and constant learning rate \( \tau > 0 \). Then, at step $k$, the discretized diffusion algorithm~\eqref{eq:euler_disc} with score $v_t$ samples a Gaussian distribution $q_k$ that satisfies \vspace{-0.cm}
\begin{equation*} 
\begin{split}
     \mathbb{E}_{\theta}&\left[W_2\left(p, q_k\right)^2\right] = 
 \mathcal{E}^{(0)}_{T-t_k} + \gamma \mathcal{E}^{\text{disc}}_{T-t_k} \\[-0.2cm] &+ \gamma \sum_{i,j = 1}^d \left(\tau k_{T-t_k}^\tau (\lambda_i, \lambda_j)  + \frac{\tau}{N} k_{T-t_k}^{\tau, N}(\lambda_i, \lambda_j) +\frac{1}{N} k_{T-t_k}^N(\lambda_i, \lambda_j)\right)  + o\left( \gamma^2, \tau, \frac{1}{N} \right)
\end{split} 
 \end{equation*}
where, for $r \in [0,T]$, $k_r^\tau, k_r^N, k_r^{\tau, N} : \RR^d \times \RR^d \to \RR$ are kernel functions that capture the interactions between the eigenvalues of $C_\mathrm{data}$. Their expressions, with respect to the kernels $\kappa^{(1)}_{r,s}$ and $\kappa^{(2)}_{r,s}$ defined in Theorem~\ref{thm:distance_general_diff} are:
\begin{align*}
k_r^\tau (\lambda_i, \lambda_j) &=  \int_r^T\frac{1}{2 \sigma_s^2} \left(\frac{\lambda_j}{\lambda^s_j}\kappa^{(1)}_{r,s}(\lambda_i, \lambda_j) +  \mathbf{1}_{i=j} \frac{\lambda_i}{\lambda^s_i}\left( h_{r,s}^2 \left(\frac{\lambda_i^{r}}{\lambda_i^{s}}\right)^{\alpha+1} + \kappa^{(2)}_{r,s}(\lambda_i, \lambda_i) \right) \right) ds\\
k_r^N (\lambda_i, \lambda_j) &=  \int_r^T s_s^2 \Biggl(\left(\kappa^{(1)}_{r,s}(\lambda_i, \lambda_j)  + \kappa^{(2)}_{r,s}(\lambda_i, \lambda_j) \right) \frac{\lambda_i \lambda_j}{(\lambda^s_i \lambda^s_j)^2}\\
&+ \mathbf{1}_{i=j} \left( h_{r,s}^2   \left(\frac{\lambda_i^{r}}{\lambda_i^{s}}\right)^{\alpha+1} \frac{\lambda_i}{(\lambda^s_i)^2} + \left(\kappa^{(1)}_{r,s}(\lambda_i, \lambda_i)  + \kappa^{(2)}_{r,s}(\lambda_i, \lambda_i) \right) \frac{\lambda_i^2}{(\lambda^s_i)^4}\right) \Biggr) ds \nonumber
\\
 k_r^{\tau, N} (\lambda_i, \lambda_j) &= \int_r^T \frac{s_s^2}{\sigma_s^2}\Biggl(
- \left( \frac{\lambda_j}{\lambda^s_j}\right)^2  \kappa^{(1)}_{r,s}(\lambda_i, \lambda_j)   \\ &+ \mathbf{1}_{i=j} \left(\frac{1}{2} h_{r,s}^2 \left(\frac{\lambda_i^{r}}{\lambda_i^{s}}\right)^{\alpha+1} \sum_k \frac{\lambda_k^2}{\lambda^s_k} - \left( \frac{\lambda_i}{\lambda^s_i}\right)^2\left( h_{r,s}^2 \left(\frac{\lambda_i^{r}}{\lambda_i^{s}}\right)^{\alpha+1}  + \kappa^{(2)}_{r,s}(\lambda_i, \lambda_i)\right) \right) \Biggr)ds \nonumber
\end{align*}
\end{repeatcor2}
\begin{remark}
The above kernels are not written in symmetric form but can be symmetrized by the operation $$k_{r} (\lambda_i, \lambda_j)  \gets \frac{1}{2} \left(k_{r} (\lambda_i, \lambda_j)  + k_{r} (\lambda_j, \lambda_i)\right)$$
\end{remark}
without changing the value of the $W_2$ error.
\begin{proof}
We consider the linear score
\begin{equation*}
    v_t(x, \theta) = -A_t x + b_t
\end{equation*}
with parameters \( A_t \) and \( b_t \) trained using the Denoising Score Matching loss~\eqref{eq:DSM_loss} at noise level \( \sigma_t \), with data rescaled by $s_t$ i.e.
\begin{align*}
   F_t(\theta) &\eqdef  \mathbb{E}_{x,w} \left[
    \frac{1}{2}  \| v_{t}\left(s_t(x+\sigma_t w), \theta\right) + \frac{w}{s_t\sigma_t} \|^2 \right]
\end{align*}
We can use Theorem~\ref{thm:SGD_error} and replace $\mu \to s_t \mu = \mu_t$, $C_{\mathrm{data}} \to s_t^2 C_{\mathrm{data}}$, $\sigma \to s_t \sigma_t$, and thus 
$C_\sigma = C_{\mathrm{data}} + \sigma^2 \id  \to s_t^2 C_{\mathrm{data}} +  s_t^2 \sigma_t^2 \id = \Sigma_t$.  Therefore, small enough fixed stepsize $\tau$, the SGD iterates converge to a stationary distribution on the $A_t$ and $\kappa_t$, with means 
\begin{align*}
 \EE[\kappa_t] &= o\left( \frac{1}{N}\right) \quad \text{and} \quad
 \EE[A_t] =  \Sigma_t^{-1} + o\left( \frac{1}{N}\right)
\end{align*}
and covariances
\begin{align*}
\cov(\kappa_t) &=  \tau \frac{1}{2 \sigma_t^2}\Sigma_t^{-1} C_\mathrm{data} + \frac{\tau}{N}\frac{s_t^2}{\sigma_t^2} \left(\frac{1}{2} \langle \Sigma_t^{-1},C_\mathrm{data}^2 \rangle - (\Sigma_t^{-1} C_\mathrm{data})^2\right) \\ &+ \nonumber \frac{1}{N}s_t^2 \Sigma_t^{-2} C_\mathrm{data} + o \left(\tau, \frac{1}{N}\right) \\
\cov(A_t) &=  \frac{\tau}{2\sigma_t^2} \left(\Sigma_t^{-1} C_\mathrm{data} \right)\otimes \id - \frac{\tau}{N} \frac{s_t^2}{\sigma_t^2} (\Sigma^{-1} C_\mathrm{data})^2 \otimes \id \\
&+ \nonumber \frac{1}{N} s_t^2 (\Sigma_t^{-2} C_\mathrm{data} \otimes \Sigma_t^{-2}C_\mathrm{data}) (I + P_{trans}) + o\left(\tau, \frac{1}{N}\right)
\end{align*}
Using this score, we apply Theorem~\ref{thm:distance_general_diff} and obtain an averaged error that depends on the values of the
\[
\langle \cov(\delta_{t}), u_i u_i^\top \rangle, \quad \mathbb{E}[\langle \Delta_{t}, u_i u_j^\top \rangle^2], \quad \text{and} \quad \mathbb{E}[\langle \Delta_{t}, u_i u_i^\top \rangle \langle \Delta_{t}, u_i u_j^\top \rangle].
\]
We can directly use the calculations derived for Langevin in the proof of Corollary~\ref{cor:langevin_final}, with the replacements
$\sigma \to s_t^2 \sigma_t^2$, $\lambda \to s_t^2 \lambda$, $\lambda^\sigma \to \lambda^t$.
\begin{align*}
    \langle \cov(\delta_{t}), u_i u_i^\top \rangle = \frac{\tau}{2\sigma_t^2} \frac{\lambda_i}{\lambda^t_i}
    +  \frac{\tau s_t^2 }{N \sigma_t^2} \left( \frac{1}{2} \sum_k \frac{\lambda_k^2}{\lambda^t_k}- \left(\frac{\lambda_i}{\lambda^t_i}\right)^2 \right) 
    + \frac{1}{N} s_t^2 \frac{\lambda_i}{(\lambda^t_i)^2}
    +  o \left(\tau, \frac{1}{N}\right) 
\end{align*}
\begin{align*}
      \mathbb{E}[\langle \Delta_{t}, u_i u_j^\top \rangle^2]
      &=  \frac{\tau}{2\sigma_t^2} \frac{\lambda_j}{\lambda^t_j} - \frac{\tau s_t^2 }{N \sigma_t^2 } \left( \frac{\lambda_j}{\lambda^t_j}\right)^2 + \frac{s_t^2}{N}\frac{\lambda_i \lambda_j}{(\lambda^t_i \lambda^t_j)^2}( 1  + \mathbf{1}_{i=j})  + o \left(\tau, \frac{1}{N}\right) 
    \end{align*}
\begin{align*}
    \mathbb{E}[\langle \Delta_{t}, u_i u_j^\top \rangle \langle \Delta_{t}, u_j u_i^\top \rangle]
      &=  \frac{\tau}{2\sigma_t^2} \mathbf{1}_{i=j} \frac{\lambda_i}{\lambda^t_i} - \frac{\tau s_t^2}{N \sigma_t^2 } \mathbf{1}_{i=j} \left( \frac{\lambda_i}{\lambda^t_i}\right)^2 + \frac{s_t^2}{N}\frac{\lambda_i \lambda_j}{(\lambda^t_i \lambda^t_j)^2}( 1  + \mathbf{1}_{i=j})  + o \left(\tau, \frac{1}{N}\right) 
    \end{align*}
With the above calculations, the term $ \mathcal{E}_{t}^{\text{mean}}(\delta) + \mathcal{E}_{t}^{\text{cov}}(\Delta)$ in Theorem~\ref{thm:distance_general_diff} can be written as the sum of three kernel norms over the eigenvalues $\lambda_i$ of $C_{\mathrm{data}}$, as defined in the corollary.

\paragraph{Simplifications for $r \to 0$}
Note that the kernel $k^\tau_r$ can be written as $k^\tau_r = \int_r^T \frac{1}{\sigma_s^2} f(r,s)ds$  where $f(r,s)$ has finite limit $f(0,0)$ when $r$ and $s$ tend to $0$. Therefore, when $r\to0$, using the simplifications of $\kappa^{(1)}$ and $\kappa^{(2)}$ calculated in~\eqref{eq:kappa1_diff_simp} and~\eqref{eq:kappa2_diff_simp} 
\begin{align*}
    k_r^\tau (\lambda_i, \lambda_j) &\sim \frac12\left(\kappa^{(1)}_{0,0}(\lambda_i, \lambda_j) +  \mathbf{1}_{i=j} \left( h_{0,0}^2 + \kappa^{(2)}_{0,0}(\lambda_i, \lambda_i) \right) \right) \int_r^T \frac{1}{\sigma_s^2} ds \\
    &\sim  (1+\alpha)^2 \xi_0^2 \left( \frac{\lambda_j^3}{(\lambda_i + \lambda_j)^2} + \mathbf{1}_{i=j}\left(\frac12 + \frac14 \lambda_i\right) \right)\int_r^T \frac{1}{\sigma_s^2} ds
\end{align*}
i.e. $ k_r^\tau (\lambda_i, \lambda_j) \sim C^{\tau}(\lambda_i, \lambda_j) \int_r^T \frac{1}{\sigma_s^2} ds$.
In the same way, for $k^{\tau,N}_r$ and $k^{N}_r$, we can get 
\begin{align*}
    k^{\tau,N}_r \to C^{N}(\lambda_i, \lambda_j)\int_r^T ds = C^{N}(\lambda_i, \lambda_j)(T-r)
\end{align*}
and 
\begin{align*}
    k^{\tau,N}_r \sim C^{\tau,N}(\lambda_i, \lambda_j)\int_r^T \frac{s_s^2}{\sigma_s^2} ds \sim C^{\tau,N}(\lambda_i, \lambda_j) \int_r^T \frac{1}{\sigma_s^2} ds
\end{align*}
with $C^{N}(\lambda_i, \lambda_j)$ and $C^{\tau,N}(\lambda_i, \lambda_j)$ which depend explicitly on $\lambda_i$ and $\lambda_j$ (as well as on $\alpha$ and $\xi_0$).
\end{proof}

\subsection{Non-uniform discretization stepsizes $\gamma_k$}
\label{app:non-uniform-gamma}

When using non-uniform stepsizes $\gamma_k$ in the discretized diffusion process~\eqref{eq:euler_disc}, the theoretical error must be adapted. We consider that $\gamma_k = \gamma_{t_k}$ is the value at timestep $t_k $ of a continuous time-dependent stepsize function $\gamma_t$. Following the same strategy, in Theorem~\ref{thm:distance_general_diff}, we get a similar decomposition but (i) with a different discretization error term $\tilde{\mathcal{E}^{\text{disc}}_{T-t_k}}$ and (ii) with the parameter $\gamma$ replaced by $\gamma_t$ inside the integrals defining $\mathcal{E}_{r}^{\text{mean}}(\delta)$ and $\mathcal{E}_{r}^{\text{cov}}(\Delta)$. More specifically, Theorem~\ref{thm:distance_general_diff} becomes, denoting $\gamma \eqdef \max_k \gamma_{t_k}$:
\begin{equation*}
\EE_{\delta, \Delta}  \left[ W_2^2(p_\mathrm{data},q_k^\varepsilon) \right] =  \mathcal{E}^{(0)}_{T-t_k} +  \tilde{\mathcal{E}^{\text{disc}}_{T-t_k}} + \varepsilon^2 \left( \mathcal{E}_{T-t_k}^{\text{mean}}(\delta) + \mathcal{E}_{T-t_k}^{\text{cov}}(\Delta) \right) + o(\gamma, \varepsilon^2)
\end{equation*}
where, for $r = T-t$,
\begin{align*}
\mathcal{E}_{r}^{\text{mean}}(\delta)  &= \sum_{i=1}^d \int_{r}^{T} \gamma_r h_{r,s}^2 \left(\frac{\lambda_i^{r}}{\lambda_i^{s}}\right)^{\alpha+1} \EE_{\delta} \left[\langle \cov \delta_{s}, u_i u_i^\top \rangle\right]ds \\
\mathcal{E}_{r}^{\text{cov}}(\Delta)  &=  \sum_{i,j=1}^d  \int_{r}^{T} \gamma_r \! \!\left( \kappa^{(1)}_{r,s}(\lambda_i, \lambda_j)  \EE_{\Delta}\!\left[\langle \Delta_s, u_iu_j^\top \rangle^2\right]  + \!\kappa^{(2)}_{r,s}(\lambda_i, \lambda_j) \, \EE_{\Delta}\!\left[\langle \Delta_s, u_iu_j^\top \rangle \langle \Delta_s, u_ju_i^\top \rangle \right] \right) ds
\end{align*}
In practice, we observe numerically that the discretization error is negligible compared to other sources of error when $\gamma_k$ is sufficiently small. This is confirmed in Figure~\ref{fig:diffusion}, where the theoretical curve \emph{Theory decreasing $\gamma_k$}, obtained with $\tilde{\mathcal{E}^{\text{disc}}_{T-t_k}} = 0$, closely matches the empirical one.



\subsection{Trade-off in the choice of the final timestep $t_k$}
\label{app:analysis_diffusion}
Similar to the analysis conducted for Langevin in Appendix~\ref{app:analysis_Langevin} for the choice of the noise parameter~$\sigma^*$, we analyze here the choice of the ultimate iteration~$k$ (or equivalently of the ultimate timestep~$t_k^*$). 

Recall the form of the $W_2$ error (from Corollary~\ref{cor:diffusion_final}) of the diffusion process at iteration $k$: 
\begin{equation} \label{eq:W2_diff_exp}
\begin{split} 
     \mathbb{E}_{\theta}&\left[W_2\left(p, q_k\right)^2\right] = 
 \mathcal{E}^{(0)}_{T-t_k} + \gamma \mathcal{E}^{\text{disc}}_{T-t_k} \\[-0.2cm] &+ \gamma \sum_{i,j = 1}^d \left(\tau k_{T-t_k}^\tau (\lambda_i, \lambda_j)  + \frac{\tau}{N} k_{T-t_k}^{\tau, N}(\lambda_i, \lambda_j) +\frac{1}{N} k_{T-t_k}^N(\lambda_i, \lambda_j)\right)  + o\left( \gamma^2, \tau, \frac{1}{N} \right)
\end{split} 
 \end{equation}
with, for $r = T-t$,
\begin{align*}
\mathcal{E}^{(0)}_{r} &=  \sum_{i=1}^d \left(\sqrt{\lambda_i} - \sqrt{\lambda^r_i}\right)^2 +  (1- s_{r})^2\norm{\mu}^2. 
\end{align*}
As shown in Figure~\ref{fig:langevin}, the evolution of the $W_2$ error with respect to $T-t_k$ shows that there is an optimal choice of $t_k$, called $t_k^*$, that minimizes the error.

\paragraph{Error analysis for small $T-t_k$}  In order to explain this, we analyze theoretically the behavior of the $W_2$ error~\eqref{eq:W2_diff_exp} for small $r_k = T-t_k$. We start with the analysis of the first-order term~$\mathcal{E}^{(0)}_{r}$. Using that ${\lambda^r_i = s_r^2(\lambda^r_i + \sigma_r^2 \id)}$, 
\begin{align*}
\mathcal{E}^{(0)}_{r} &= \sum_{i=1}^d \lambda_i \left(1 - s_r\sqrt{1 + \frac{\sigma_r^2}{\lambda_i}}\right)^2 + (1- s_{r})^2\norm{\mu}^2
\end{align*}
Now expanding the first term (using that $s_r \to 1$ and $\sigma_r \to 0$)
\begin{align*}
\mathcal{E}^{(0)}_{r} &=  (1- s_{r})^2 \sum_{i=1}^d \lambda_i - (1-s_r)\sigma_r^2 d + \sigma_r^4 \sum_{i=1}^d \frac{1}{4 \lambda_i} +   (1- s_{r})^2\norm{\mu}^2 + o((1-s_r)^2, \sigma_r^4) \\
&= \mathcal{O}((1- s_{r})^2) + \mathcal{O}((1- s_{r})\sigma_r^2) + \mathcal{O}(\sigma_r^4)
\end{align*}
 In the Variance Preserving (VP) setting presented in Section~\ref{sec:background}, we have $\sigma_r^2 = \frac{1}{s_r^2} (1 - s_r^2) =  \mathcal{O}(1- s_{r})$ and thus $\mathcal{E}^{(0)}_{r} = \mathcal{O}(\sigma_r^4)$. In the Variance Exploding (VE) setting, we have $s_r = 1$ and thus again ${\mathcal{E}^{(0)}_{r} = \mathcal{O}(\sigma_r^4)}$. 
 
 On the other hand, as seen at the end of the proof of Corollary~\ref{cor:diffusion_final}, the terms $k_{r}^\tau$ and $k_{r}^{\tau,N}$ diverge at rate $\mathcal{O}\left(\int_{r}^{T} \frac{1}{\sigma_s^2}ds\right)$ when $r \to 0$, and finally $k^N_{r}$ has finite limit. For small $\tau, \gamma, \frac{1}{N}$ and small $r_k = T-t_k$, there are some constants $C_0$ and $C_1$ which depend on the data spectrum, such that, we have approximately
\begin{align*}
  \mathbb{E}_{\theta}&\left[W_2\left(p, q_k\right)^2\right] \approx C_0 \sigma_{r_k}^4 + \tau  C_1 \int_{r_k}^{T} \frac{1}{\sigma_s^2}ds 
\end{align*}
The behavior of the optimal $r_k^* = T - t_k^*$ then depends on the choice of the evolution of the $\sigma_t$. In the classical VP setting of Figure~\ref{fig:diffusion}, with the linear evolution of the $\beta_t$, $B_r = \int_0^r \beta_s ds = \mathcal{O}(r)$ and $\sigma_r^2 = e^{2 B_r} - 1 = \mathcal{O}(r)$, which leads to an expression of the form 
\begin{align*}
  \mathbb{E}_{\theta}&\left[W_2\left(p, q_k\right)^2\right] \approx C_0 r_k^4 - \gamma \tau C_1 \log(r_k)
\end{align*}
The $r_k^* = T - t_k^*$ that minimizes the above expression is
\begin{align*}
r_k^* = \left(\frac{ C_1}{C_0} \right)^{1/4} \tau^{1/4}
\end{align*}

\begin{figure}[h]
    \centering
    \begin{subfigure}[b]{0.45\textwidth}
    \centering
     \includegraphics[width=0.9\linewidth]{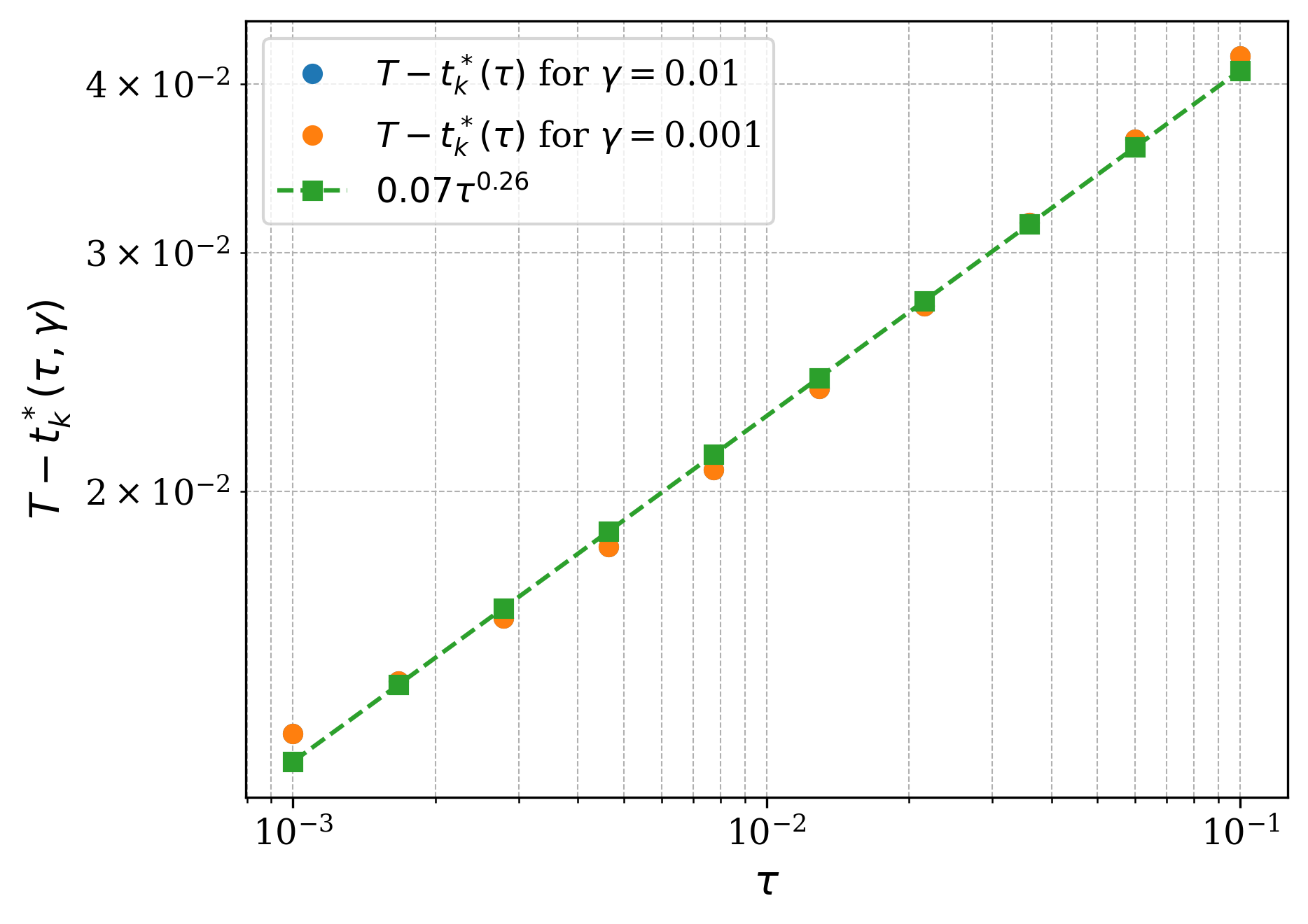}
     \caption{$r_k^*(\tau)$}
     \label{fig:sigma_opt_tau_siff}
    \end{subfigure}
    \begin{subfigure}[b]{0.45\textwidth}
    \centering
    \includegraphics[width=0.9\linewidth]{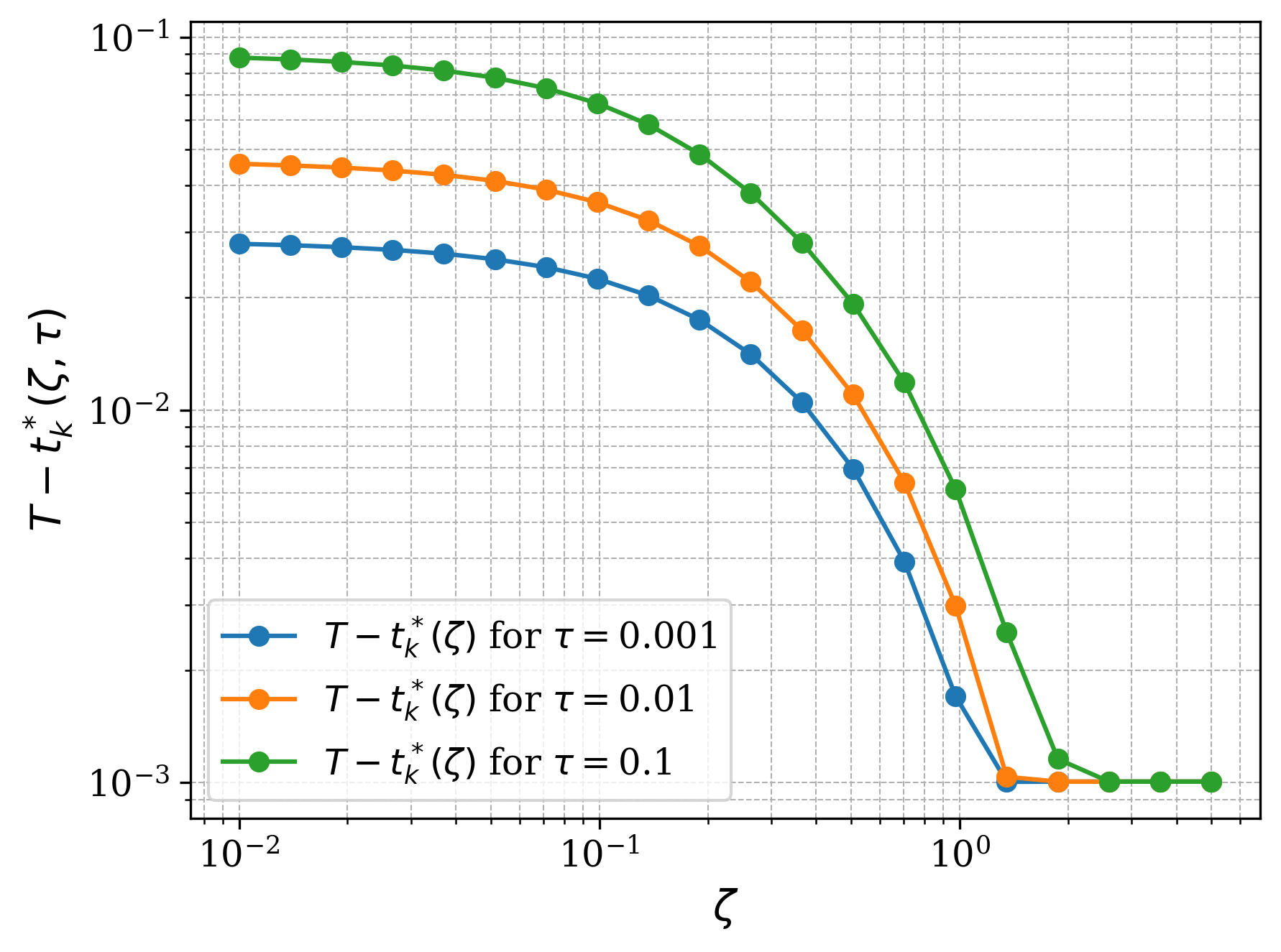}
     \caption{$r_k^*(\zeta)$ for various $\tau$.}
      \label{fig:sigma_opt_zeta_siff}
    \end{subfigure}
    \caption{Evolution of the optimal stopping time $r_k^* = T-t_k^*$ with respect to the SGD learning rate~$\tau$ and the data power law coefficient $\zeta$.}
    \label{fig:analysis_diffusion}
\end{figure}

\paragraph{Numerical calculation of $t_k^*$} Similar to Appendix~\ref{app:analysis_Langevin} we calculate numerically the value of the $t_k^*$ that minimizes the theoretical $W_2$ error~\eqref{eq:W2_diff_exp}.  We plot in Figure~\ref{fig:analysis_diffusion} the evolution of the optimal $r_k^* = T-t_k^*$, with respect to~$\tau$ (Figure~\ref{fig:sigma_opt_tau_siff}) and with respect to the data power spectrum coefficient~$\zeta$ (introduced in Appendix~\ref{app:analysis_Langevin}) (Figure~\ref{fig:sigma_opt_zeta_siff}).

First, in Figure~\ref{fig:sigma_opt_tau_siff}, we numerically confirm that $r_k^*(\tau)$ scales as $\tau^{1/4}$. Second, Figure~\ref{fig:sigma_opt_zeta_siff} illustrates that the value of $t_k^*$ is highly sensitive to the shape of the data spectrum. When $\zeta$ decreases, meaning the eigenvalues decay more rapidly, the diffusion process needs to be stopped earlier. This observation aligns with the theoretical analysis in Appendix~\ref{app:analysis_Langevin}, which shows that for smaller $\zeta$, the optimal final noise-scale parameter (i.e., $\sigma$ in Langevin dynamics or $T - t_k^*$ in diffusion) becomes larger.

\end{document}